\documentclass[12pt]{article}
\usepackage{amsmath,amssymb}
\usepackage{mathtools}
\usepackage{graphicx}
\usepackage{enumerate}
\usepackage{natbib}
\usepackage{url} 
\usepackage{caption}
\usepackage{subcaption}
\usepackage{amsthm}
\usepackage{threeparttable}
\usepackage{multirow}
\usepackage{hyperref}
\hypersetup{
    colorlinks,
    linkcolor=red,
    citecolor=blue,
    urlcolor=red
}
\usepackage{xcolor}
\usepackage{subfiles}

\newcommand{\blind}{1} 
\newtheorem{theorem}{Theorem}
\newtheorem{lemma}{Lemma}
\newtheorem{proposition}{Proposition}
\newtheorem{remark}{Remark}

\numberwithin{equation}{section}

\begin{document}

\def\spacingset#1{\renewcommand{\baselinestretch}%
{#1}\small\normalsize} \spacingset{1}


\if1\blind
{
  \title{\large \bf \textcolor{black}{Mini-batch Estimation for Deep Cox Models: \\Statistical Foundations and Practical Guidance}}
  \author{Lang Zeng \\
    Department of Biostatistics and Health Data Science, University of Pittsburgh\\
    \\
    Weijing Tang \\
    Department of Statistics and Data Science, Carnegie Mellon University\\
    \\
    Zhao Ren \\
    Department of Statistics, University of Pittsburgh\\
    \\
    Ying Ding\thanks{Corresponding Author Email: yingding@pitt.edu}\hspace{.2cm}\\
    Department of Biostatistics and Health Data Science, University of Pittsburgh}
  \date{\vspace{-5ex}}
  \maketitle
} \fi

\if0\blind
{
  \bigskip
  \bigskip
  \bigskip
  \begin{center}
  {\large \bf \textcolor{black}{Mini-batch Estimation for Deep Cox Models: \\Statistical Foundations and Practical Guidance}}
\end{center}
  \medskip
} \fi

\bigskip
\begin{abstract}
The stochastic gradient descent (SGD) algorithm has been widely used to optimize deep Cox neural network (Cox-NN) by updating model parameters using mini-batches of data. 
We show that SGD aims to optimize the average of mini-batch partial-likelihood, which is different from the standard partial-likelihood. This distinction requires developing new statistical properties for the global optimizer, namely, the mini-batch maximum partial-likelihood estimator 
(mb-MPLE). We establish that mb-MPLE for Cox-NN is consistent and achieves the optimal minimax convergence rate up to a polylogarithmic factor. For Cox regression with linear covariate effects, we further show that mb-MPLE is $\sqrt{n}$-consistent and asymptotically normal with asymptotic variance approaching the information lower bound as batch size increases, \textcolor{black}{which is confirmed by simulation studies.}
Additionally, we offer practical guidance on using SGD, \textcolor{black}{supported by theoretical analysis and numerical evidence.} 
For Cox-NN, we demonstrate that the ratio of the learning rate to the batch size is critical in SGD dynamics, offering insight into hyperparameter tuning. For Cox regression, we characterize the iterative convergence of SGD, ensuring that the global optimizer, mb-MPLE, can be approximated with sufficiently many iterations. Finally, we demonstrate the effectiveness of mb-MPLE in a large-scale real-world application where the standard MPLE is intractable.
\end{abstract}

\noindent%
\vfill

\newpage
\spacingset{1.9} 

\section{Introduction\label{Introduction}}

Cox proportional hazards regression \citep{cox1972regression} is one of the most commonly used approaches in survival analysis, where the outcome of interest is the time to a certain event. It assumes that the covariates have linear effects on the log-hazard function. With the development of deep learning, Cox deep neural networks (Cox-NN, or deep Cox model) have been proposed to capture the potential nonlinear relationship between covariates and survival outcomes to improve survival prediction accuracy \citep{faraggi1995neural,katzman2018deepsurv,ching2018cox}. Despite the success of the Cox model in survival analysis, it faces a significant optimization challenge when applied to large-scale data. In particular, the Cox model is typically trained by maximizing the partial likelihood \citep{cox1975partial}. The maximum partial likelihood estimator (MPLE) is obtained through the gradient descent (GD) algorithm, which requires the entire dataset to compute the gradient. This approach is computationally demanding and memory-intensive, especially with large datasets. For example, in our motivating application, where images are used to predict survival outcomes (as detailed in Section \ref{sec: Real Data Analysis}), performing GD in Cox-NN with high-dimensional predictors and a large sample size is infeasible due to hardware memory constraints. Even traditional Cox regression faces optimization challenges for large-scale data. \citet{tarkhan2024online} reported that the GD algorithm for Cox regression is prone to round-off errors when dealing with large sample sizes. Therefore, the scalability of the Cox model is substantially limited by the inefficiencies of GD optimization when applied to large-scale data.

The stochastic gradient descent algorithm (SGD) is a scalable solution for optimization with large-scale data and has been widely used for training NN \citep{amari1993backpropagation,bottou2012stochastic}. SGD alleviates the computational and memory burdens by using a randomly selected (small) subset of the dataset, known as a mini-batch, to compute the gradient and update parameters in each iteration. Moreover, in NN optimization, \citet{xie2020diffusion} showed that, compared to GD, SGD favors flat minima, which often leads to better generalization. This explains why SGD is preferred for training an NN.
However, SGD cannot directly optimize the partial likelihood of all samples through mini-batches because evaluating the partial likelihood of an event subject requires access to data for all subjects who survived longer. 

Several attempts have been made to enable parameter updates for Cox models using mini-batch data.
\citet{kvamme2019time} followed the idea of nested case-control Cox regression from \citet{goldstein1992asymptotic}, proposing an approximation of the gradient of partial likelihood using the case-control pairs instead of using all at-risk samples. \citet{sun2020genome} fitted a Cox-NN to establish a prediction model for an eye disease progression through SGD, where the iteration was based on the partial likelihood of a random subset of data. \citet{tarkhan2024online} studied SGD for Cox regression in an online learning setting and demonstrated that the gradient of the expected mini-batch partial likelihood at the true parameter is zero. Despite the successful application of SGD to Cox models, the statistical foundations of the mini-batch maximum partial likelihood estimator (mb-MPLE), of which the SGD seeks to estimate, remain unexplored. Notably, the average mini-batch partial likelihood depends on the batch size and differs from the full partial likelihood, as highlighted in (\ref{eq:sgd neq gd}). Such a distinction in the objective function differentiates mb-MPLE from MPLE, requiring the development of new statistical properties for the mb-MPLE.

Motivated by the existing knowledge gap, this work investigated the statistical properties of mb-MPLE for Cox models and provided practical guidance for the SGD application to find the mb-MPLE. Specifically, our contributions come from the following three folds: 

\begin{itemize}

    \item Firstly, for Cox-NN, where the SGD algorithm is commonly used, we establish the consistency and convergence rate of mb-MPLE. Unlike MPLE, which targets the partial likelihood of all samples \citep{cox1975partial}, mb-MPLE minimizes a different objective function (see (\ref{eq:SGD NN minimizer}) for details). Consequently, the statistical properties developed for  MPLE by \citet{zhong2022deep} cannot be applied directly. We demonstrate that the mb-MPLE remains consistent and achieves the minimax optimal convergence rate up to a polylogarithmic factor. \textcolor{black}{The consistency is also supported by numerical evidence.} These results provide a statistical foundation for mini-batch estimation in Cox-NN.
    
    \item Secondly, we provide practical guidance for training Cox-NN via SGD to search for the mb-MPLE. In practice, both the learning rate and batch size used in SGD are important hyperparameters, as they significantly influence the SGD dynamics. In optimizations where the objective function is independent of batch size, such as empirical risk minimization, \textit{the ratio of the learning rate to the batch size} has been identified as a key factor in SGD dynamics during NN training  \citep{goyal2017accurate,jastrzkebski2017three,xie2020diffusion}. This observation guides a hyperparameter tuning strategy, where the learning rate or the batch size is fixed while tuning the other. However, for Cox-NN, we note that the function targeted by SGD is batch-size dependent, so it remains unclear whether these results can be applied. To this end, we investigate how the local convexity of the objective changes with the batch size and extend the applicability of the tuning strategy to Cox-NN optimization. \textcolor{black}{We provide both theoretical insights and numerical evidence} that \textit{the ratio of the learning rate to the batch size} remains a crucial factor in SGD dynamics during Cox-NN training. This offers insight into Cox-NN hyperparameter tuning.

    \item Lastly, we develop the statistical properties of mb-MPLE for Cox regression with linear covariate effects. We show that mb-MPLE is $\sqrt{n}$-consistent and asymptotically normal, with asymptotic variance depending on the batch size. \textcolor{black}{Furthermore, based on theoretical analysis and numerical experiments on the impact of batch size on local convexity,} we demonstrate that doubling the batch size improves the statistical efficiency of mb-MPLE. This phenomenon is not observed in other SGD optimizations, such as empirical risk minimization, where the optimizer's statistical efficiency is independent of the SGD batch size. Moreover, for Cox regression, we study the numerical convergence of the SGD algorithm to the mb-MPLE. We note that the objective function optimized by SGD is not global strong convex, and an additional projection step is necessary (namely, the \textit{projected SGD}). We follow the non-asymptotic analysis of the projected SGD from \citet{moulines2011non} to demonstrate that the algorithm approximates the mb-MPLE given sufficiently many iterations.
    
\end{itemize}

The rest of the paper is organized as follows. In Section \ref{sec:Background and Problem Setup}, we review the Cox model and the SGD algorithm, with an introduction to the average mini-batch log-partial likelihood and the mb-MPLE for the Cox model. In Section \ref{sec:SGD Estimator for Cox Neural Network}, we establish the statistical properties of mb-MPLE for Cox-NN and investigate the impact of SGD batch size on training Cox-NN to find the mb-MPLE. The statistical properties of the mb-MPLE for Cox regression, as well as the convergence of SGD algorithm to the mb-MPLE over iterations, are studied in Section \ref{sec:SGD for Cox regression}. Section \ref{sec:Numerical Implementation and Results} presents simulation studies, and Section \ref{sec: Real Data Analysis} presents a real-world data analysis. Finally, we conclude and discuss further investigation directions in Section \ref{sec:discussion}.

\section{Background and Problem Setup\label{sec:Background and Problem Setup}}
Let $D(n) = \{D_i\}_{i=1}^n$ denote $n$ independently and identically distributed (i.i.d.) observations. Survival analysis is to analyze the association between the covariate $X\in\mathbb{R}^{p}$ and the time-to-event outcome $T^*$ (e.g., time-to-death). Let $C^*$ denote the censoring time. Due to the right censoring, the observed time-to-event data is the triplet set $D_i= (X_i, T_i,\Delta_i)$, where $T_i=\min(T^*_i, C^*_i)$ is the observed time and $\Delta_i=I(T^*_i \leq C^*_i)$ is the event indicator.

\subsection{Cox Model and Deep Neural Network\label{sec:Cox Model and Deep Neural Network}}
The Cox model assumes a multiplicative effect of covariates on the hazard function, i.e., 
\begin{equation}\label{eq:cox model}
\lambda(t|X=x) = \lambda_0(t)\exp\{f_0(x)\},
\end{equation}
where $\lambda_0(t)$ is the baseline hazard function and $f_0(x)$ is the relative risk given covariates~$x$. When assuming the effect of $X$ is linear, i.e., $f_0(X) = X^T\theta_0$, the model (\ref{eq:cox model}) reduces to the Cox regression model. When the function $f_0(\cdot)$ is unspecified and is modeled through a neural network (NN) $f_\theta$ with parameter~$\theta$, it is referred to as Cox-NN. 

We briefly review the structure of a NN below. 
Let $K$ be a positive integer and $\mathbf{p} = \{p_0,\dots,p_K,p_{K+1}\}$ be a positive integer sequence. A $(K+1)$-layer NN with layer-width $\mathbf{p}$, i.e., the number of neurons, is a composite function $f:\mathbb{R}^{p_0}\to\mathbb{R}^{p_{K+1}}$ recursively defined as
\begin{equation}\label{eq:NN}
\begin{split}
f(x)&=W_Kf_{K}(x)+v_K, \\
f_k(x)&= \sigma(W_{k-1}f_{k-1}(x)+v_{k-1}) \text{ for }k=2,\dots,K, \\
f_1(x)&= \sigma(W_{0}x+v_{0}),
\end{split}
\end{equation}
where the matrices $W_{k}\in \mathbb{R}^{p_{k+1}}\times\mathbb{R}^{p_{k}}$ and vectors $v_k\in \mathbb{R}^{p_{k+1}}$ (for $k=0,\dots,K$) are the parameters of the NN. The pre-specified activation function $\sigma(\cdot)$ is a nonlinear transformation that operates component-wise on a vector, that is, $\sigma((x_1,\dots,x_{p_k})^T) = (\sigma(x_1),\dots,\sigma(x_{p_k}))^T$, for $k=1,\dots,K$. 
For NN in (\ref{eq:NN}), $K$ denotes the depth of the network. The sequence $\mathbf{p}$ lists the width of each layer with the first element $p_0$ being the dimension of the input variable, $p_1,\dots,p_K$ are the dimensions of the $K$ hidden layers, and $p_{K+1}$ is the dimension of the output layer. The matrix entries $(W_k)_{i,j}$ are the weight linking the $j$th neuron in layer $k$ to the $i$th neuron in layer $k+1$, and the vector entries $(v_k)_i$ represent a shift term associated with the $i$th neuron in layer $k+1$. In Cox-NN, the output dimension $p_{K+1} = 1$ since the output $f_\theta(X)$ is a real value. 
We consider the commonly used activation function named Rectified Linear Units (ReLU) in \cite{nair2010rectified}, i.e., $\sigma(x) = \max\{x,0\}$,
which has been similarly considered in \cite{schmidt2020nonparametric} and \cite{zhong2022deep}. 

The standard approach to optimizing the Cox model is through minimizing the negative log-partial likelihood function, which is defined as
\begin{equation}
L^{(n)}_{Cox}(\theta) :=  -\frac{1}{n}\sum_{i=1}^n \Delta_i \log \frac{\exp\{f_\theta(X_i)\}}{\sum_{j=1}^n I(T_j\geq T_i)\exp\{f_\theta(X_j)\}}.  \label{eq:coxph_lik}
\end{equation}
We refer to the minimizer of (\ref{eq:coxph_lik}) as the MPLE. In the partial likelihood, sample $i$ who has experienced the event ($\Delta_i = 1$) is compared to all samples that have not yet experienced the event up to time $T_i$, known as the \textit{at-risk} set. Minimizing (\ref{eq:coxph_lik}) is typically solved using gradient-descent-based algorithms, where the gradient $\nabla_\theta L^{(n)}_{Cox}(\theta)$ is calculated using all $n$ samples at each iteration \citep{therneau2015package,katzman2018deepsurv,zhong2022deep}. The consistency and asymptotic normality of the MPLE have been well-studied by \citet{andersen1982cox} for Cox regression. For Cox-NN, the consistency of the MPLE and its convergence rate have been developed by \cite{zhong2022deep}. However, when optimizing the Cox model through the SGD algorithm using a subset of samples, the algorithm does not directly minimize (\ref{eq:coxph_lik}), which will be further explained in the next section.

\subsection{Stochastic Gradient Descent for Cox Model}\label{sec:SGD}
The loss function of the Cox model distinguishes its optimization from the typical loss function (e.g., mean squared error (MSE)) when applying the SGD algorithm. In this section, we first provide a brief overview of the SGD algorithm, followed by a discussion of how SGD for the Cox model differs from its standard applications, such as minimizing the MSE, along with the introduction of the mini-batch loss function for Cox model.

Let $L^{(n)}(D(n);\theta)$ be an empirical loss function over $n$ samples to be minimized. In each iteration of the SGD algorithm, the gradient is computed based on a subset of data $D(s)\subset D(n)$, referred to as a mini-batch, where $s$ denotes the batch size $\ (1\leq s\leq n)$. Throughout this paper, we assume the batch size $s$ is fixed and is independent of the sample size $n$. At the $(t+1)$-th iteration step of SGD, the parameter is updated through
\begin{equation}
\hat\theta_{t+1} = \hat\theta_{t} - \gamma_t {\nabla_\theta}L^{(s)}(D_t(s);\hat \theta_t),    \label{eq:SGDupdate}
\end{equation}
where $\gamma_t$ is a pre-scheduled learning rate for the $t$-th iteration. The gradient ${\nabla_\theta}L^{(s)}(D_t(s);\hat \theta_t)$ is calculated from the mini-batch data $D_t(s)$ at $t$-th iterations. Suppose $D_t(s)$ is randomly selected from $D(n)$ with ${\nabla_\theta}L^{(s)}(D_t(s);\hat \theta_t)$ uniformly bounded for all $t$, when $t\to\infty$, it has been shown that SGD (\ref{eq:SGDupdate}) with an appropriate learning rate will find the minimizer of $\mathbb{E}_{D_t(s)}[L^{(s)}(D_t(s);\hat \theta _{t})|D(n)]$ if it is strongly convex, where the expectation is over the randomness of generating $D_t(s)$ from $D(n)$ \citep{moulines2011non,toulis2017asymptotic}.
Throughout the paper, we assume $D_t(s)$ is randomly sampled from $D(n)$ at each iteration, so its distribution is independent of~$t$.  To simplify the notation, for any~$\theta$, we omit the input data and drop the subscript~$t$ and let $L^{(s)}(\theta):=L^{(s)}(D(s);\theta)$, $L^{(n)}(\theta):=L^{(n)}(D(n);\theta)$, and $\mathbb{E}[L^{(s)}(\theta)|D(n)]:=\mathbb{E}_{D(s)}[L^{(s)}(D(s);\theta)|D(n)]$.

Take optimizing the MSE for linear regression as an example, the loss function $L^{(n)}_{MSE}(\theta) =\frac{1}{n}\sum_{i=1}^nL_i(\theta)$ is the average of independent individual loss $L_i(\theta):=(Y_i- X_i^T\theta)^2$ which is the discrepancy between the prediction and the observed for subject~$i$. Suppose $D(s)$ is generated by randomly sampling $s$ subjects without replacement from $D(n)$. When optimizing MSE by SGD with the gradient calculated by ${\nabla_\theta}L^{(s)}_{MSE}(\theta)=\frac{1}{s}\sum_{i: D_i \in D(s)}{\nabla_\theta}L_i(\theta)$, SGD optimizes the loss function
\begin{equation}
\begin{split}
\mathbb{E}[L^{(s)}_{MSE}(\theta)&|D(n)] = \frac{1}{\binom{n}{s}}\sum_{D(s)\subset D(n)}\left\{\frac{1}{s}\sum_{i: D_i \in D(s)}L_i(\theta)\right \} \\
 &= \frac{\binom{n-1}{s-1}}{s\binom{n}{s}}\sum_{i: D_i \in D(n)}L_i(\theta)= \frac{1}{n}\sum_{i: D_i \in D(n)}L_i(\theta) = L^{(n)}_{MSE}(\theta).
\end{split}\label{eq:expectation-of-decomposable-loss}
\end{equation}
This suggests that the loss function targeted by SGD coincides with the MSE, so that minimizers of $\mathbb{E}[L^{(s)}_{MSE}(\theta)|D(n)]$ and $L^{(n)}_{MSE}(\theta)$  exhibit the same statistical properties.

For the Cox model, the SGD implementation \citep{sun2020genome,tarkhan2024online} requires at least two samples to calculate the gradient, which restricts $s\geq 2$. The parameters are updated through 
\begin{equation}
\hat\theta_{t+1} = \hat\theta_{t} - \gamma_t {\nabla_\theta}L^{(s)}_{Cox}(\hat\theta_t),    \label{eq:SGDupdate for Cox}
\end{equation}
where the gradient of mini-batch partial likelihood from $D(s)$ is calculated by
\begin{equation}
\label{eq:coxph_batch_lik}
{\nabla_\theta}L^{(s)}_{Cox}(\theta) :=  {\nabla_\theta}\left \{-\frac{1}{s}\sum_{i:D_i \in D(s)} \Delta_i \log \frac{\exp\{f_\theta(X_i)\}}{\sum_{j: D_j \in D(s)} I(T_j\geq T_i)\exp\{f_\theta(X_j)\}}\right \}. 
\end{equation}
and the at-risk set is constructed on $D(s)$. Since the partial likelihood of each individual depends on other at-risk individuals, the average mini-batch partial likelihood does not equal the partial likelihood of the entire dataset. Suppose $D(s)$ is generated by randomly sampling $s$ subjects without replacement from $D(n)$, the SGD recursion (\ref{eq:SGDupdate for Cox}) is to optimize 
\begin{equation}\label{eq:sgd neq gd}
\mathbb{E}\left[L^{(s)}_{Cox}(\theta)|D(n)\right]  = \frac{1}{\binom{n}{s}}\sum_{D(s)\subset D(n)}L^{(s)}_{Cox}(\theta)\neq L^{(n)}_{Cox}(\theta).
\end{equation}
 As a result, the loss function targeted by SGD differs from the negative log-partial likelihood $L^{(n)}_{Cox}(\theta)$ in (\ref{eq:coxph_lik}). Therefore, the theoretical results for the MPLE, which minimizes $L^{(n)}_{Cox}(\theta)$, cannot be directly applied to the mb-MPLE, which minimizes $\mathbb{E}\left [L^{(s)}_{Cox}(\theta)|D(n)\right ]$. This motivates us to investigate the statistical properties of the mb-MPLE.
 

\section{mb-MPLE for Cox-NN\label{sec:SGD Estimator for Cox Neural Network}}

We first establish the consistency and the convergence rate of the mb-MPLE for Cox-NN in Section \ref{sec:consistency and convergence rate of SGD estimator}. In Section \ref{sec:impact of Mini-batch Size in Cox-NN Optimization}, we examine the impact of batch size on Cox-NN training when using SGD to search for the mb-MPLE in practice.

\subsection{Statistical Properties of the mb-MPLE \label{sec:consistency and convergence rate of SGD estimator}}

\citet{zhong2022deep} studied the asymptotic properties of the MPLE which minimizes (\ref{eq:coxph_lik}). They showed that the MPLE achieves the minimax optimal convergence rate up to a polylogarithmic factor. While our work is inspired by the theoretical developments in \citet{zhong2022deep}, it has significant differences. First, the at-risk term $\frac{1}{n}\sum_{j=1}^n I(T_j\geq t)\exp\{f_\theta(X_j)\}$ converges to a fixed function $\mathbb{E}[I(Y>t)\exp\{f_\theta (X)\}]$. However, in the average mini-batch log-partial likelihood, the at-risk term in $L^{(s)}_{Cox}(\theta)$ always consists of $s$ sub-samples. Thus, both the empirical loss $\mathbb{E}[L^{(s)}_{Cox}(\theta)\lvert D(n)]$ and the population loss $\mathbb{E}[L^{(s)}_{Cox}(\theta)]$ depend on $s$ and are different from \citet{zhong2022deep}. Second, when $D(s)$ is generated by randomly sampling $s$ subjects without replacement from $D(n)$ in each iteration, $\mathbb{E}[L^{(s)}_{Cox}(\theta)\lvert D(n)]$ is an average of $\binom{n}{s}$ combinations of mini-batches, where different batches may share the same samples. Hence, the correlation between batches needs to be handled when deriving the properties of the minimizer of $\mathbb{E}[L^{(s)}_{Cox}(\theta)\lvert D(n)]$.

We impose the following assumptions for the time-to-event data with right censoring: 
\begin{itemize}
    \item [(A1)] The failure time $T_i^*$ and censoring time $C_i^*$ are independent given the covariates $X_i$.
    \item [(A2)] There is a truncation time $\tau < \infty $ such that, for some constant $\delta>0$, $\mathbb{P}(T^* >  \tau|X)\geq \delta$ and $\mathbb{P}(\Delta=1|X)\geq \delta$ almost surely with respect to the probability measure of $X$. The stochastic integrals computed from here on will be truncated at this value $\tau$.
    \item [(A3)] $X$ takes value in a compact subset of $\mathbb{R}^{p}$ with probability density function bounded away from zero. Without loss of generality, we assume that the domain of $X$ is $[0,1]^{p}$.
\end{itemize}

(A1)-(A3) are common regularity assumptions in survival analysis \citep{huang1999efficient,zhong2022deep}. (A1) ensures that the censoring mechanism is noninformative. 
(A2) is a technical assumption that prevents the partial likelihood and score functions from becoming unbounded at the endpoint of the observed event time support. It also ensures the probability of being uncensored is positive regardless of the covariate value.

Following \cite{schmidt2020nonparametric} and \cite{zhong2022deep}, we consider a class of NN with sparsity constraints defined as
\begin{equation}\label{eq:NN family}
\begin{split}
\mathcal{F}(K,\varsigma,\mathbf{p},\mathcal{D})=\{& f:f\text{ is a DNN with ($K$+1) layers and width vector $\mathbf{p}$ such that} \\ 
& \max\{\lVert W_k\rVert_\infty,\lVert v_k\rVert_\infty\}\leq 1\}\text{ , for all $k=0,\dots,K$}, \\
& \sum_{k=1}^K\lVert W_k\rVert_0+\lVert v_k\rVert_0\leq\varsigma,\lVert f\rVert_\infty \leq \mathcal{D}\},
\end{split}
\end{equation}
where $\lVert W_k \rVert_\infty$ and $\lVert v_k \rVert_\infty$ denote the sup-norm of a matrix or vector, $\lVert \cdot \rVert_0$ is the number of nonzero entries of a matrix or vector, $\lVert f\rVert_\infty$ is the sup-norm of the function $f$. The constant $\mathcal{D}>0$ and the sparsity constraint $\varsigma$ is a positive integer. The sparsity assumption is employed due to the widespread application of techniques such as weight pruning \citep{srinivas2017training}, dropout \citep{srivastava2014dropout}, or $L_1$ regularization. These methods effectively reduce the total number of nonzero parameters, preventing neural networks from overfitting, which results in a sparsely connected NN. We approximate $f_0$ in (\ref{eq:cox model}) using a NN $f_\theta\in\mathcal{F}(K,\varsigma,\mathbf{p},\mathcal{D})$. More precisely, $f_0$ is estimated by minimizing $\mathbb{E}\left [L^{(s)}_{Cox}(\theta)|D(n)\right ]$ through the SGD procedure in (\ref{eq:SGDupdate for Cox}), where $D(s)$ is generated by randomly sampling $s$ subjects without replacement from $D(n)$. To simplify the notation, we omit $\theta$ and denote the estimator by
\begin{equation}
\Tilde{f}^{(s)}_n = \displaystyle\arg \min_{f\in \mathcal{F}(K,\varsigma,\mathbf{p},\mathcal{D})} \frac{1}{\binom{n}{s}}\displaystyle\sum_{D(s)\subset D(n)}L_{Cox}^{(s)}(f),
\label{eq:SGD NN minimizer}
\end{equation}
where $L_{Cox}^{(s)}(f) :=\frac{1}{s}\sum_{j:D_j\in D(s)}\Delta_j\left [-f(X_j)+\log \sum_{k:D_k\in D(s)}I(T_k\geq T_j)\exp\{f(X_k)\}\right]$. We use the superscript $(s)$ to indicate its dependency on the batch size $s$.

We assume $f_0$ belongs to a composite smoothness function class, which is broad and also assumed by \cite{zhong2022deep}. Specifically, let $\mathcal{H}^{\alpha}_{r}(\mathbb{D},M)$ be a H\"older class of smooth functions with parameters $\alpha,M>0$ and domain $\mathbb{D}\subseteq\mathbb{R}^r$ defined by
$$\mathcal{H}^{\alpha}_{r}(\mathbb{D},M) = \left\{h:\mathbb{D}\to\mathbb{R}:\sum_{\beta:\lvert\beta\rvert<\alpha}\lVert\partial^\beta h\rVert_\infty+\sum_{\beta:\lvert\beta\rvert=\lfloor \alpha\rfloor}\sup_{x,y\in\mathbb{D},x\neq y}\frac{\lvert \partial^\beta h(x) - \partial^\beta h(y) \rvert}{\lVert x-y \rVert_\infty^{\alpha-\lfloor \alpha\rfloor}}\leq M \right\},$$
where $\lfloor \alpha\rfloor$ is the largest integer strictly smaller than $\alpha$, $\partial^\beta:=\partial^{\beta_1}\dots \partial^{\beta_r}$ with $\beta = (\beta_1,\dots,\beta_r)$, and $\lvert\beta\rvert = \sum_{k=1}^r \beta_k$. 

Let $q\in \mathbb{N}$,  $\vec{\alpha} = (\alpha_0,\dots,\alpha_q)\in \mathbb{R}_+^{q+1}$ and $\textbf{d}=(d_0,\dots,d_{q+1})\in \mathbb{N}_+^{q+2}$, $\mathbf{\Tilde{d}}=(\Tilde{d}_0,\dots,\Tilde{d}_{q})\in \mathbb{N}_+^{q+1}$ with $\Tilde{d}_j\leq d_j,j=0,\dots q$, where $\mathbb{R}_+$ is the set of all positive real numbers. The composite smoothness function class is
\begin{equation}\label{eq:holder class}
\begin{split}
\mathcal{H}(q,\vec{\alpha},\mathbf{d},\mathbf{\Tilde{d}},M):=\{& h = h_q\circ \dots \circ h_0: h_i = (h_{i1},\dots,h_{id_{i+1}})^T \text{ and }\\
& h_{ij}\in\mathcal{H}^{\alpha_i}_{\Tilde{d}_i}([a_i,b_i]^{\Tilde{d}_i},M),\text{ for some }\lvert a_i \rvert,\lvert b_i \rvert \leq M\},
\end{split}
\end{equation}
where $\mathbf{\Tilde{d}}$ is the intrinsic dimension of the function. For example, if
$$f(x) = f_{21}(f_{11}(f_{01}(x_1,x_2),f_{02}(x_3,x_4)),f_{12}(f_{03}(x_5,x_6),f_{04}(x_7,x_8)))\quad x\in[0,1]^8,$$
and $f_{ij}$ are twice continuously differentiable, then the smoothness parameter $\alpha = (2,2,2)$, the dimension vectors $\mathbf{d} = (8,4,2,1)$ and $\mathbf{\Tilde{d}} = (2,2,2)$. We assume that

\begin{itemize}
    \item[(N1)] The unknown function $f_0$ is an element of $\mathcal{H}_0=\{f \in \mathcal{H}(q,\vec{\alpha},\mathbf{d},\mathbf{\Tilde{d}},M):\mathbb{E}[f(X)] = 0\}.$
\end{itemize}
The mean zero constraint in (N1) is for the identifiability of $f_0$ since the presence of two unknown functions in (\ref{eq:cox model}). 

We denote $\Tilde{\alpha}_i=\alpha_i\prod_{k=i+1}^q(\alpha_k \wedge 1)$ and $\Upsilon_n=\max_{i=0,\dots,q} n^{-\Tilde{\alpha}_i/(2\Tilde{\alpha}_i+\Tilde{d}_i)}$ with notation $a\wedge b :=\min\{a,b\}$. Furthermore, we denote $a_n\lesssim b_n$ as $a_n\leq cb_n$ for some constant $c>0$ and any $n$. $a_n \asymp b_n$ means $a_n\lesssim b_n$ and $b_n\lesssim a_n$. We assume the following NN structure:
\begin{itemize}
    \item[(N2)] $K=O(\log n)$, $\varsigma=O(n\Upsilon_n^2\log n)$, $\mathcal{D}>2M$, and $n\Upsilon_n\lesssim \displaystyle\min_{k=1,\dots,K} (p_k)\leq \displaystyle\max_{k=1,\dots,K} (p_k) \lesssim n$.
\end{itemize}
A large neural network leads to a smaller approximation error and a larger estimation error. (N2) defines the structure of the NN family in (\ref{eq:NN family}) and was also adopted in \citet{zhong2022deep} to balance the trade-off between approximation and estimation errors. 

We first present a lemma that is critical to studying the asymptotic properties of $\Tilde{f}^{(s)}_n$:
\begin{lemma}\label{lemma: minimizer does not depend on s}
Let $L^{(s)}_{0}(f) := \mathbb{E}[L^{(s)}_{Cox}(f)] = \mathbb{E}_{D(n)}\left[\mathbb{E}\left [L^{(s)}_{Cox}(\theta)|D(n)\right ]\right] $. Under the Cox model, with assumptions (A1)-(A3) and (N1), for any integer $s \geq 2$ and constant $c>0$, we have
$$L^{(s)}_{0}(f)-L^{(s)}_{0}(f_0)\asymp  d^2(f,f_0)$$
for all $f\in \{f:\lVert f\rVert_\infty\leq c, \mathbb{E}[f(X)]=0\}$, where $d(f,f_0)=[\mathbb{E}\{f(X)-f_0(X)\}^2]^{\frac{1}{2}}$.
\end{lemma}

\begin{remark}\label{remark: minimizer does not depend on s}
Suppose $\Tilde{f}^{(s)}=\displaystyle\arg\min_{f:\lVert f\rVert_\infty\leq c, \mathbb{E}[f(X)]=0}L^{(s)}_{0}(f)$ and by definition $L^{(s)}_{0}(\Tilde{f}^{(s)})\leq L^{(s)}_{0}(f_0)$. On the other hand, we have $L^{(s)}_{0}(\Tilde{f}^{(s)})\geq L^{(s)}_{0}(f_0)$ from Lemma \ref{lemma: minimizer does not depend on s}. Hence, $L^{(s)}_{0}(\Tilde{f}^{(s)})-L^{(s)}_{0}(f_0)=0$ and it implies $\Tilde{f}^{(s)}=f_0$. That is, for any integer $s\geq 2$, the minimizer of  $L^{(s)}_{0}(f)$ on a neighborhood of $f_0$ is $f_0$ and does not depend on $s$. Our result can be viewed as a generalization of the result in \citet{tarkhan2024online}, where they consider a parametric function $f_{\theta}$ with the truth $f_0 = f_{\theta_0}$ and demonstrate that $\nabla_\theta L^{(s)}_{0}(f_\theta)|_{\theta=\theta_0}=0$. 
\end{remark}

Next, we establish the consistency of the mb-MPLE $\Tilde{f}^{(s)}_n$ defined in (\ref{eq:SGD NN minimizer}) for Cox-NN. Note that the expectation of the estimator $\Tilde{f}^{(s)}_n$ in (\ref{eq:SGD NN minimizer}) is not necessarily zero. For the identifiability condition $\mathbb{E}[f_0] = 0$ in (N1), we apply a mean shift by subtracting the empirical average $\Bar{\Tilde{f}}^{(s)}_n:= \frac{1}{n}\sum_{i=1}^n\Tilde{f}^{(s)}_n(X_i)$. The following theorem gives the convergence rate of the mean-shifted mb-MPLE. This is different from \citet{zhong2022deep}, where the convergence rate was established for the MPLE after shifting by the true mean.
\begin{theorem}\label{thm:convergence rate}
Under the Cox model, with assumptions (A1)-(A3) and (N1)-(N2), for any integer $s\geq 2$, 
we have
$$\lVert (\Tilde{f}^{(s)}_n-\Bar{\Tilde{f}}^{(s)}_n)-f_0 \rVert_{L^2([0,1]^{p})}=O_p(\Upsilon_n\log^2 n).$$
\end{theorem}

\begin{remark}\label{remark:curse of dimensionality}
For any integer $s\geq 2$, the convergence rate of $\Tilde{f}^{(s)}_n$ is the same as the MPLE in \cite{zhong2022deep}. The rate is determined by the smoothness and the intrinsic dimension $\mathbf{\Tilde{d}}$ of the function $f_0$, rather than the dimension $\mathbf{d}$. Therefore, the estimator $\Tilde{f}^{(s)}_n$ can circumvent the curse of dimensionality \citep{bauer2019deep} and has a fast convergence rate when the intrinsic dimension is low. The batch size $s$ only implicitly influences the constant of the rate. 
Later, as we consider the parametric Cox model in Section~\ref{sec:SGD for Cox regression}, the impact of $s$ on the asymptotic variance of the estimator becomes more apparent. 
\end{remark}

\begin{remark}\label{remark:minmax convergence rate}
The minimax lower bound for estimating $f_0$ can be derived following the proof of Theorem 3.2 in \citet{zhong2022deep} as their Lemma 4 and derivations still hold without the parametric linear term. Specifically, let $\Omega_0 = \{\lambda_0 (t):\int_0^\infty \lambda_0(s)ds <\infty \text{ and }\lambda_0 (t)\geq 0 \text{ for } t\geq 0\}$. Under the Cox model with assumptions (A1)-(A3) and (N1), there exists a constant $c>0$, such that
$\inf_{\hat{f}}\sup_{(\lambda_0,f_0)\in \Omega_0\times \mathcal{H}_0}\mathbb{E}\{\hat{f}(X)-f_0(X)\}^2\geq c\gamma^2_n$, where the infimum is taken over all possible estimators $\hat{f}$ based on the observed data. Thus, the NN-estimator is rate-optimal as it attains the minimax lower bound up to a polylogarithm factor.
\end{remark}

\begin{remark}
We note that Theorem \ref{thm:convergence rate} established the properties for the global optimal $\Tilde{f}^{(s)}_n$. In practice, however, NN training involves solving a challenging non-convex optimization problem, and the SGD does not generally guarantee convergence to the global minimum. As a result, there may be an optimization error between the SGD output $\hat{f}^{(s)}_n$ and the mb-MPLE $\Tilde{f}^{(s)}_n$. In the case of Cox regression, where the optimization problem is convex, we showed in Theorem \ref{thm:non-asymp-bound} that SGD converges to the global optimum given sufficient iterations. For the general non-convex setting in Cox-NN, the optimization error depends on factors, such as the learning rate and initialization (see \citet{choromanska2015loss,kleinberg2018alternative,schmidt2020nonparametric} for more discussion). We leave the analysis of the optimization error in Cox-NN for future work.
\end{remark}

We conducted simulations to numerically verify the Theorem \ref{thm:convergence rate} by evaluating the root mean squared error (RMSE) and prediction performance of the estimator $\hat{f}^{(s)}-\Bar{\hat{f}}^{(s)}$, given by SGD, under different training sample sizes $n$ with different choices of $s$. Across various choices of $s$, as the training sample size increases, the RMSE decreases, and the prediction accuracy is improved on holdout test data (see details in the Supplementary Material).

\subsection{Impact of Batch Size on Cox-NN Training\label{sec:impact of Mini-batch Size in Cox-NN Optimization}}

The previous section focuses on the statistical properties of the mb-MPLE $\Tilde{f}^{(s)}_n$, the global minimizer of (\ref{eq:SGD NN minimizer}). In practice, searching for the mb-MPLE through SGD is challenging since the NN training is a highly non-convex optimization problem and can be affected by many factors \citep{li2018visualizing}. It has been noticed that the ratio of the learning rate to the batch size $\gamma/s$ is a key factor that determines SGD dynamics \citep{goyal2017accurate,jastrzkebski2017three}. 
\citet{goyal2017accurate} demonstrates that keeping $\gamma/s$ constant makes the SGD training process almost unchanged on a broad range of batch sizes, which is known as the \textit{linear scaling rule}. This rule guides hyperparameter tuning in NN such that we can fix either $\gamma$ or $s$ and only tune the other parameter to optimize the behavior of SGD. 

It remains unclear whether the linear scaling rule can be applied to Cox-NN training, as the population loss $\mathbb{E}[ L^{(s)}_{Cox}(\theta)]$ in Cox-NN depends on the batch size. When applying SGD in training a neural network $f_\theta$ with a specified structure, we treat the parameter $\theta$ as finite-dimensional. Existing theoretical work focused on the population loss $\mathbb{E}[L(\theta)]$ which is invariant to different batch sizes and assumed that the Hessian equals the covariance of gradient at the truth, that is $\nabla^2_\theta\mathbb{E}[L(\theta)]|_{\theta=\theta_0} = \mathbb{V}[\nabla_\theta L(\theta)]|_{\theta=\theta_0}$ \citep{jastrzkebski2017three,xie2020diffusion}.
These key properties of the objective function are violated for the population loss $\mathbb{E}[ L^{(s)}_{Cox}]$ under the Cox model, as shown in the following theorem.

\begin{theorem}
\label{thm: batch size and Hessian}
Under the Cox model with $f_0 = f_{\theta_0}$ parameterized by $\theta_0$ of finite dimension, with assumptions (A1)-(A3) and suppose $\nabla_\theta f_\theta$, $\nabla^2_\theta f_\theta$ exist and $f_\theta$,$\nabla_\theta f_\theta$, $\nabla^2_\theta f_\theta$ are element-wise bounded for all $X$ on a neighborhood of $\theta_0$ with $ \nabla^2_\theta\mathbb{E}[L_{Cox}^{(s)}(\theta)]|_{\theta=\theta_{0}}$ being positive definite, then for any integer $s\geq 2$ we have
\begin{equation}
    \nabla^2_\theta\mathbb{E}[L_{Cox}^{(s)}(\theta)]|_{\theta=\theta_{0}} = s\mathbb{V}[\nabla_\theta L_{Cox}^{(s)}(\theta)]|_{\theta=\theta_{0}}, \label{eq: Hessian and covariance}
\end{equation}
\begin{equation}
    \nabla^2_\theta\mathbb{E}[L_{Cox}^{(2s)}(\theta)]|_{\theta=\theta_{0}} \succeq \nabla^2_\theta\mathbb{E}[L_{Cox}^{(s)}(\theta)]|_{\theta=\theta_{0}}, \label{eq: batch size and Hessian}
\end{equation}
\textcolor{black}{where $\mathbb{V}$ denotes the variance}, $\nabla^2_\theta$ is second-order derivative operator with respect to $\theta$ and $A \succeq B$ denotes $A-B$ is positive semi-definite.
\end{theorem}

\begin{remark}\label{remark:convexity change of parametric Cox model}
For $i$th individual loss $L_i$, the property $\nabla^2_\theta\mathbb{E}[L_i(\theta)]|_{\theta=\theta_0} = \mathbb{V}[\nabla_\theta L_i(\theta)]|_{\theta=\theta_0}$ immediately implies $\nabla^2_\theta\mathbb{E}[\frac{1}{s}\sum_{i=1}^s L_i(\theta)]|_{\theta=\theta_{0}} = s\mathbb{V}[\nabla_\theta \frac{1}{s}\sum_{i=1}^s L_i(\theta)]|_{\theta=\theta_{0}}$ for any mini-batch of $s$ i.i.d. samples. The equality (\ref{eq: Hessian and covariance}) indicates this batch-level relation still holds for the negative-log-partial likelihood, which is useful to study the SGD dynamic \citep{jastrzkebski2017three,xie2020diffusion}. Moreover, the inequality (\ref{eq: batch size and Hessian}) and Remark \ref{remark: minimizer does not depend on s} together depict the properties of $\mathbb{E}[ L^{(s)}_{Cox}(\theta)]$. The global minimizer of $\mathbb{E}[ L^{(s)}_{Cox}(\theta)]$ is always $\theta_0$, which does not depend on the batch size $s$, while the local convexity of $\mathbb{E}[ L^{(s)}_{Cox}(\theta)]$ increases when $s$ doubles. Figure \ref{fig:double_batch_size} presents an illustrative picture of $\mathbb{E}[ L^{(s)}_{Cox}(\theta)]$ showing these two properties.
\end{remark}

The relation (\ref{eq: batch size and Hessian}) is only established at the true $\theta_0$, but it gives us the intuition that the convexity of $\mathbb{E}[ L^{(s)}_{Cox}(\theta)]$ increases when $s$ increases. Nevertheless, we show that the convexity change is negligible when $s$ is large, as shown in the following proposition:

\begin{proposition}\label{prop:trace and batch size}
Let $H_s = \nabla^2_\theta\mathbb{E}[L_{Cox}^{(s)}(\theta)]|_{\theta=\theta_{0}}$ and let Tr($\cdot$) denote the trace. Under the Cox model with the assumptions in Theorem \ref{thm: batch size and Hessian} hold, then $\text{Tr}(H_{2s})-\text{Tr}(H_{s}) \lesssim 1/s.$
\end{proposition}

\textcolor{black}{The Proposition \ref{prop:trace and batch size} suggests that, when $s$ is sufficiently large, we can treat $ \nabla^2_\theta\mathbb{E}[ L^{(s)}_{Cox}(\theta)]$ as approximately invariant to $s$, allowing existing insights on SGD to extend to Cox-NN.} Following \cite{jastrzkebski2017three}, we approximate the SGD by a stochastic differential equation and show that the linear scaling rule for Cox-NN training remains approximately valid when the batch size is large (see details in the Supplementary Material). This is numerically verified in Section \ref{sec:Numerical Implementation and Results}. As a hyperparameter tuning strategy, we can fix either $\gamma$ or $s$ and only tune the other to optimize the behavior of SGD. This strategy reduces one hyperparameter to be tuned in Cox-NN training.

\section{mb-MPLE for Cox Regression}\label{sec:SGD for Cox regression}
In this section, we further establish the asymptotic normality of the mb-MPLE for Cox regression, where the linear effect of covariates is assumed. Specifically, we study the impact of $s$ on the asymptotic variance of the linear coefficient estimator and the convergence of the SGD algorithm over iterations.
We consider the Cox regression model with $f(X)=\theta_0^TX$ and estimate $\theta_0$ through the SGD procedure (\ref{eq:SGDupdate for Cox}) with 
\begin{equation}
{\nabla_\theta}L^{(s)}_{Cox}(\theta) :=  -\frac{1}{s}\sum_{i:D_i \in D(s)} \Delta_i \left [X_i- \frac{\sum_{j: D_j \in D(s)} I(T_j\geq T_i)\exp\{\theta^TX_j\}X_j}{\sum_{j: D_j \in D(s)} I(T_j\geq T_i)\exp\{\theta^TX_j\}}\right ]. 
\end{equation}
We assume the following assumptions for the regression setting:
\begin{itemize}
    \item[(R1)] The true parameter $\theta_0$ associated with the relative risk $f_0(X) = \theta_0^TX$ is an interior point of $\mathbb{R}_M^{p} :=\{\theta \in \mathbb{R}^{p}:\lVert\theta\rVert_2\leq M\}$.
    \item[(R2)] There exist constants $0<c_1<c_2<\infty$ such that the subdensities $p(x,t,\Delta = 1)$ of $(X,T,\Delta=1)$ satisfies $c_1<p(x,t,\Delta = 1)<c_2$ for all $(x,t)\in [0,1]^{p}\times[0,\tau]$.
\end{itemize}
The subdensity $p(x,t,\Delta = \delta)$ is defined as
$$p(x,t,\Delta = \delta)=\frac{\partial^2\mathbb{P}(X\leq x,T\leq t,\Delta=\delta)}{\partial t\partial x}.$$
Assumptions (R1) and (R2) are sufficient to demonstrate  the strong convexity of $\mathbb{E} [L^{(s)}_{Cox}(\theta)]$ on any compact subspace of $\mathbb{R}^{p}$, so that $\mathbb{E}[\nabla^2_\theta L_{Cox}^{(s)}(\theta)]|_{\theta=\theta_0}\succ 0$ and $\theta_0$ is identifiable. See Lemma \ref{lemma: strongly convex} for the details.

Next, depending on the nature of data collection, we study the mb-MPLE for Cox regression under the offline scenario (unstreaming data) and the online scenario (streaming data). In the first scenario, the entire dataset $D(n)$ of $n$ i.i.d. samples has been collected. In the second scenario, the observations arrive in a continual stream of strata, and the model is continuously updated as new data arrives without storing the entire dataset. 

\subsection{Offline Cox Regression\label{sec:SGD for Offline Cox regression}}
To study the statistical properties of $\Tilde{\theta}_n^{(s)}:=\displaystyle\arg\min_{\theta}\mathbb{E} [L^{(s)}_{Cox}(\theta)|D(n) ]$, we consider two sampling strategies of $D(s)$ from $D(n)$ which determines the form of $\mathbb{E} [L^{(s)}_{Cox}(\theta)|D(n)]$. The first strategy has been considered in Theorem \ref{thm:convergence rate} where $D(s)$ is generated by randomly sampling $s$ subjects from $D(n)$ without replacement. We refer to it as a stochastic batch (SB) with the estimator
\begin{equation}
\tilde{\theta}^{SB(s)}_{n} = \arg\min_{\theta\in\mathbb{R}_M^{p}}\frac{1}{\binom{n}{s}}\sum_{D(s)\subset D(n)} L_{Cox}^{(s)}(\theta).\label{eq:SBGD estimator}
\end{equation}

Another popular strategy in SGD applications is to randomly split the whole sample into $m = n/s$ non-overlapping batches. Once the mini-batches are established, they are fixed and then repeatedly used throughout the rest of the algorithm \citep{qi2023statistical}. We refer to it as a fixed batch (FB) with the estimator \begin{equation}
\tilde{\theta}^{FB(s)}_{n} = \arg\min_{\theta\in\mathbb{R}_M^{p}} \frac{1}{m}\sum_{D(s)\in D(n|s)} L_{Cox}^{(s)}(\theta),
\label{eq:FBGD estimator}
\end{equation} 
where we use $D(n|s)$ to denote that $D(n)$ has been partitioned into $m = n/s$ fixed disjoint batches. The element of $D(n|s)$ is a mini-batch containing $s$ i.i.d. samples and $D(s)$ is generated by randomly picking one element from $D(n|s)$. The impact of batch size on the asymptotic variance of $\tilde{\theta}^{FB(s)}_{n}$ can be well-established.

We present the asymptotic properties of $\tilde{\theta}^{SB(s)}_{n}$ and $\tilde{\theta}^{FB(s)}_{n}$ in the following theorem:

\begin{theorem}\label{thm:asymp-dist-SBFB}
Under the Cox model, with assumptions (A1)-(A3) and (R1)-(R2), for any integer $s\geq 2$, we have
\begin{align}
\sqrt{n}(\tilde{\theta}_{n}^{SB(s)}-\theta_0) & \to^d N(0,s^2H^{-1}_s\Sigma_{(s|1)}(H_s^{-1})^T), \label{eq:asymp-dist-SBGD} \\
\sqrt{n}(\tilde{\theta}_{n}^{FB(s)}-\theta_0) & \to^d N(0,sH^{-1}_s\Sigma_s(H_s^{-1})^T),
\label{eq:asymp-dist-FBGD}
\end{align}
when $n\to \infty$, where $H_s = \mathbb{E}[\nabla^2_\theta L_{Cox}^{(s)}(\theta)]|_{\theta=\theta_0}$, $\Sigma_s = \mathbb{V}[\nabla_\theta L_{Cox}^{(s)}(\theta)]|_{\theta=\theta_0}$, and
$$ \left . \Sigma_{(s|1)} = \mathbb{V}\left \{ \nabla_\theta L_{Cox}^{(s)}(D_{i_1},D_{i_2},\dots,D_{i_s}|\theta),\nabla_\theta L_{Cox}^{(s)}(D_{i_1},\Tilde D_{{i}_2},\dots,\Tilde D_{{i}_s}|\theta)\right \}\right \rvert_{\theta=\theta_0}, $$
which is the covariance of $\nabla_\theta L_{Cox}^{(s)}$ on two mini-batches $D(s)$ sharing the same sample $D_{i_1}$ but different rest $s-1$ samples (denoted by $\Tilde D$).
\end{theorem}

\begin{remark}\label{remark: asymptotic distributing of FBSB estimators}
Theorem \ref{thm:asymp-dist-SBFB} states that, for any choice of batch size $s$, both $\tilde{\theta}_{n}^{SB(s)}$ and $\tilde{\theta}_{n}^{FB(s)}$ are $\sqrt{n}$-consistent, asymptotically normal with a sandwich-type variance while the middle part of the asymptotic variance differs. By the theory of the U-statistics \citep{hoeffding1992class}, we have $s^2\Sigma_{(s|1)} \preceq s\Sigma_s$ and the equality holds only if $\nabla_\theta L^{(s)}(\theta)$ can be written as the average of $s$ i.i.d. gradients, such as MSE in (\ref{eq:expectation-of-decomposable-loss}). Because this condition does not hold for $\nabla_\theta L_{Cox}^{(s)}(\theta)$ presented in (\ref{eq:coxph_batch_lik}), it implies that $s^2H^{-1}_s\Sigma_{(s|1)}(H_s^{-1})^T \prec sH^{-1}_s\Sigma_s(H_s^{-1})^T$ and, thus,  $\tilde{\theta}_{n}^{SB(s)}$ is asymptotically more efficient than $\tilde{\theta}_{n}^{FB(s)}$. The FB strategy ignores the ranking between samples from different non-overlapping batches. This loss of information consequently reduces the
efficiency. Note that this phenomenon is not typically observed in SGD optimizations. Take the MSE in (\ref{eq:expectation-of-decomposable-loss}) for instance, one can verify that $s^2H^{-1}_s\Sigma_{(s|1)}(H_s^{-1})^T = sH^{-1}_s\Sigma_s(H_s^{-1})^T$ and $\tilde{\theta}_{n}^{FB(s)}$ is therefore as efficient as $\tilde{\theta}_{n}^{SB(s)}$.
\end{remark}

\begin{remark}\label{remark: FB}
The convergence rate established in Theorem \ref{thm:convergence rate} still holds if we consider the FB strategy for NN training and replace the empirical loss in (\ref{eq:SGD NN minimizer}) by $\frac{1}{m}\sum_{D(s)\in D(n|s)} L_{Cox}^{(s)}(f)$. The proof is simpler because there are no correlations between the batches, and the empirical loss is the average of $m$ i.i.d. tuples. Hence, the convergence rate is $O_p(\gamma_m\log^2 m)$, which is equivalent to $O_p(\Upsilon_n\log^2 n)$ as $m = n/s$ with $s$ being a fixed constant.
\end{remark}

\begin{remark}
We extend our results for the mb-MPLE to the partially linear Cox model considered by \citet{zhong2022deep}, where $\lambda(t|X, Z) = \lambda_0(t)\exp\{\theta_0^TZ+f_0(X)\}$ including both the linear component $\theta_0^TZ$ and the nonlinear component $f_0(X)$, and we find similar statistical properties of mb-MPLE as shown in this work. Specifically, the NN estimator of the nonlinear component achieves the minimax optimal rate of convergence while the finite-dimensional estimator for the linear covariate effect is $\sqrt{n}$-consistent and asymptotically normal with variance depending on the batch size (see details in Supplementary Material). This result integrates Theorem \ref{thm:convergence rate} for the nonparametric Cox model and Theorem \ref{thm:asymp-dist-SBFB} for the parametric Cox model.
\end{remark}

\begin{remark}\label{remark: impact of s in FB}
By Theorem \ref{thm: batch size and Hessian}, we have
\begin{equation}
H_s = \mathbb{E}[\nabla^2_\theta L_{Cox}^{(s)}(\theta)]|_{\theta=\theta_0}= s\mathbb{V}[\nabla_\theta L_{Cox}^{(s)}(\theta)]|_{\theta=\theta_0} = s\Sigma_s,\label{eq:finite-property-pl}
\end{equation}
\begin{equation}\label{eq:A2s>As}
H_{2s} = \mathbb{E}[\nabla^2_\theta L_{Cox}^{(2s)}(\theta)]|_{\theta=\theta_0} \succeq \mathbb{E}[\nabla^2_\theta L_{Cox}^{(s)}(\theta)]|_{\theta=\theta_0} = H_s.
\end{equation} Equality (\ref{eq:finite-property-pl}) indicates that the asymptotic variance of $\tilde{\theta}_{n}^{FB(s)}$ is $sH^{-1}_s\Sigma_s(H_s^{-1})^T = H^{-1}_s$. Then by (\ref{eq:A2s>As}), we have $H^{-1}_{2s} \preceq H^{-1}_s$, i.e., the asymptotic efficiency of $\tilde{\theta}_{n}^{FB(s)}$ improves when the $s$ doubles. The asymptotic variance of $\tilde{\theta}_{n}^{SB(s)}$ cannot be further simplified to directly evaluate the impact of batch size. Nevertheless, the decrease of its upper bound $sH^{-1}_s\Sigma_s(H_s^{-1})^T$ implies the efficiency improvement when batch size doubles.  Moreover, such an impact of batch size on the asymptotic variance is, again, not typically observed in SGD optimizations. Taking the MSE in (\ref{eq:expectation-of-decomposable-loss}) as an example, one can verify that $sH^{-1}_s\Sigma_s(H_s^{-1})^T$ degenerates to $H^{-1}_1\Sigma_1(H_1^{-1})^T$, which does not depend on $s$. The efficiency improvement of the mb-MPLE is because the objective function $\mathbb{E}[\nabla_\theta L_{Cox}^{(s)}(\theta)]$ gets closer to the efficient score function \citep{tsiatis2006semiparametric} for Cox regression when $s$ increases.
\end{remark}

\begin{remark}\label{remark: finite-property-pl}
One can verify that
\begin{equation*}
\mathbb{E}[{\nabla^2_\theta}L_{Cox}^{(s)}(\theta)]|_{\theta=\theta_0}\to I(\theta_0)\text{ and } s\mathbb{V}[{\nabla_\theta}L_{Cox}^{(s)}(\theta)]|_{\theta=\theta_0} \to I(\theta_0) \text{ when $s\to\infty,$ }
\end{equation*}
 where $I(\theta_0)$ is the information matrix of $\theta_0$. This implies that $H_s^{-1}$ decreases towards its lower bound $I(\theta_0)^{-1}$ as $s$ continues to double, suggesting that the mb-MPLE is less efficient than the MPLE, whose asymptotic variance is $I(\theta_0)^{-1}$. On the other hand, when the batch size is large and $H_s^{-1}$ is approaching $I(\theta_0)^{-1}$, the efficiency gain from doubling the batch size would diminish, which has been empirically reported in \citet{tarkhan2024online}. 
\end{remark}

\subsection{Online Cox Regression\label{sec:SGD for Online Cox regression}}

In contrast to the offline Cox regression, which minimizes the empirical loss $\mathbb{E} [L^{(s)}_{Cox}(\theta)|D(n) ]$ with given $D(n)$, the online Cox regression minimizes the population loss $\mathbb{E} [L^{(s)}_{Cox}(\theta)]$ by directly sampling the mini-batch $D(s)$ from the population. For online learning, it is of interest to study the convergence of $\hat\theta_t$ to $\theta_0$ when the iteration step $t\to\infty$, where $\hat\theta_t$ is the estimator at the $t$-th iteration in SGD. This is an optimization problem of whether an algorithm can reach the global minimizer over the iterations. It differs from our previous investigation of an estimator's asymptotic properties when the sample size goes to infinity.

\textcolor{black}{The strong convexity of the objective function has been widely used to establish the fast convergence of SGD algorithm to the global minimizer \citep{ruppert1988efficient,polyak1992acceleration,moulines2011non,toulis2017asymptotic}.} It requires the function to grow as fast as a quadratic function. Specifically, a function $h(x)$ is strongly convex over a domain $\mathcal{X}$ if there exists a constant $\mu>0$ such that its Hessian satisfies $\nu^T\nabla^2_x h(x)\nu\geq \mu>0$ for any $x\in \mathcal{X}$ and vector $ \nu\in \{\nu:\lVert \nu \rVert_2 = 1\}$, where $\lVert \cdot \rVert_2$ denotes the Euclidean norm. In our case, the population loss of online Cox regression
$\mathbb{E} [L^{(s)}_{Cox}(\theta)] = \mathbb{E}\left[-\frac{1}{s}\sum_{i:D_i\in D(s)} \Delta_i \log \frac{\exp(\theta^TX_i)}{\sum_{j:D_j\in D(s)}I(T_j\geq T_i)\exp(\theta^TX_j)}\right]$
\textcolor{black}{is not strongly convex globally over~$\mathbb{R}^p$}. For example, when $p=1$, it is straightforward to verify that $\mathbb{E} [L^{(s)}_{Cox}(\theta)]=O(\theta)$ and the Hessian vanishes when $\theta\to \infty$, hence the \textcolor{black}{global} strong convexity does not hold. 
\begin{color}{black}
However, we show that $\mathbb{E} [L^{(s)}_{Cox}(\theta)]$ is strongly convex within any ball centered at the origin that contains the true parameter $\theta_0$. This local strong convexity facilitates the application of existing SGD convergence results to the Cox regression setting.
\end{color}

\begin{lemma}\label{lemma: strongly convex}
Suppose the integer $s\geq 2$, the constant $B>0$, and $\theta_0\in \mathbb{R}^p_B:=\{\theta \in \mathbb{R}^{p}:\lVert\theta\rVert_2\leq B\}$. Under Cox regression, with assumptions (A1)-(A3), (R1), (R2), there exist a constant $\mu>0$ such that for any $\theta\in \mathbb{R}^p_B$
$$\nu^T\mathbb{E}[\nabla^2_\theta L^{(s)}_{Cox}(\theta)]\nu \geq \mu >0, \quad  \forall \nu\in \{\nu:\lVert \nu \rVert_2 = 1\}. $$
\end{lemma}

The radius $B>0$ can be arbitrarily large so that $\mathbb{R}^p_B$ covers the true parameter $\theta_0$ to guarantee $\mathbb{E}[L^{(s)}_{Cox}(\theta)]$ is strongly convex on $\mathbb{R}^p_B$. Therefore, a modification of (\ref{eq:SGDupdate for Cox}) could be applied to restrict the domain of $\theta$ and, hence, to establish the fast convergence of the SGD algorithm for Cox regression, which is called projected SGD. That is,
\begin{equation}
\hat\theta_{t+1} = \Pi_{\mathbb{R}^p_B}[ \hat\theta_t - \gamma_t \nabla_\theta L_{Cox}^{(s)}(\hat\theta_t)],    \label{eq:projected-SGDupdate for Cox}
\end{equation}
where $\Pi_{\mathbb{R}^p_B}$ is the orthogonal projection operator on the ball $\mathbb{R}^p_B:=\{\theta \in \mathbb{R}^{p}:\lVert\theta\rVert_2\leq B\}$ \citep{moulines2011non}. The projection step keeps iterates within the area where \textcolor{black}{the local strong convexity of $\mathbb{E} [L^{(s)}_{Cox}(\theta)]$ holds}. \textcolor{black}{By applying Theorem 2 of \citet{moulines2011non}, we obtain the following non-asymptotic result with respect to the iteration step $t$.}
\begin{theorem}\label{thm:non-asymp-bound}
Consider the SGD procedure (\ref{eq:projected-SGDupdate for Cox}) with learning rate $\gamma_t = \frac{C}{t^\alpha}$ where the constant $C>0$ and $\alpha\in[0,1]$. Under the Cox regression, with assumptions (A1)-(A3), (R1)-(R2), and assume $\lVert \theta_0\rVert \leq B$, then for any integer $s\geq 2$, we have 
\begin{equation}\label{eq:non-asymp-bound}
\mathbb{E}\lVert \hat\theta_t-\theta_0\rVert^2\leq  
\begin{cases}
    \{\delta_0^2+D^2C^2\varphi_{1-2\alpha}(t)\}\exp(-\frac{\mu C}{2}t^{1-\alpha})+\frac{2D^2C^2}{\mu t^\alpha}       & \quad \text{if } \alpha\in [0,1);\\
    \delta_0^2 t^{-\mu C}+2D^2C^2t^{-\mu C}\varphi_{\mu C-1}(t)  & \quad \text{if } \alpha=1,
\end{cases}
\end{equation}
where $\delta_0$ is the distance between the initialization of SGD and $\theta_0$, $D=\displaystyle\max_{\theta\in \mathbb{R}^p_B,D(s)} \lVert \nabla_\theta L_{Cox}^{(s)}(\theta) \rVert$, and $\mu$ is the strong-convexity constant in Lemma \ref{lemma: strongly convex}. The function $\varphi_{\beta}(t):\mathbb{R}^+ \setminus \{0\}\to\mathbb{R}$ is given by
$$\varphi_{\beta}(t) = 
\begin{cases}
    \frac{t^\beta-1}{\beta}    & \quad \text{if } \beta\neq 0,\\
    \log t  & \quad \text{if } \beta=0.
\end{cases}
$$
\end{theorem}

\begin{remark}\label{remark:convergence rate} This result ensures that for Cox regression, the global minimizer of $\mathbb{E}[L^{(s)}_{Cox}(\theta)]$ can be approximated by the projected SGD algorithm with a large enough number of iterations. Specifically, the upper bound in (\ref{eq:non-asymp-bound}) goes to 0 when $\alpha$ is not 0. When $\alpha\in (0,1)$, the convergence is at rate $O(\frac{1}{t^{\alpha}})$. When $\alpha=1$, the convergence rate is $O(\frac{1}{t})$ if $\mu C>1$, the convergence rate is $O(\frac{\log t}{t})$ if $\mu C=1$, and is $O(\frac{1}{t^{\mu C}})$ if $\mu C<1$. The learning rate can be set as $\gamma_t = \frac{C}{t}$ with a large constant $C$ to achieve the optimal convergence rate.

\begin{color}{black}
\begin{remark}
The high-probability error bound for the projected SGD estimator is also derived and presented in the online Supplementary Materials. Additionally, we show that the running average SGD (ASGD) estimator $\Bar{\hat{\theta}}_t = \frac{1}{t}\sum_{k=1}^t\hat\theta_{k}$ (no projection), achieves the optimal $O(t^{-1})$ rate and $\sqrt{t}(\bar{\hat{\theta}}_t - \theta_0)$ is asymptotically normal for $\gamma_t = \frac{C}{t^\alpha}$ with $\alpha \in (0.5, 1)$ and $C > 0$. This is useful for uncertainty quantification in online Cox regression (see Supplementary Materials).
\end{remark}
\end{color}

\end{remark}

\begin{remark}
The convergence of SGD for offline Cox regression can be similarly established. If $D(s)$ is sampled from the $D(n)$ using either method considered in Section \ref{sec:SGD for Offline Cox regression}, and if $\Tilde{\theta}_n := \displaystyle\arg\min_\theta \mathbb{E}[L^{(s)}_{Cox}(\theta)|D(n)]$ is finite, then the expected distance between $\hat{\theta}_t$ and $\Tilde{\theta}_n$ converges over the iterations (\ref{eq:projected-SGDupdate for Cox}). This justifies our previous investigation of statistical properties of $\Tilde{\theta}_n$, as this is the limit converged by $\hat{\theta}_t$ through the SGD algorithm. 
\end{remark}

\section{Simulation Studies\label{sec:Numerical Implementation and Results}}

\subsection{Impact of Batch Size on the Local Convexity}
We conduct simulation studies to evaluate how the local convexity of $\mathbb{E} [ \nabla^2_\theta L_{Cox}^{(s)}(\theta)]|_{\theta = \theta_0}$ changes with $s$ in Cox-regression with $\theta\in\mathbb{R}^1$. We set $\theta_0=1$ and generated $X$ from a uniform distribution $[0,10]$. For each $s$, we estimate $\hat{\mathbb{E}} [\nabla_\theta  L_{Cox}^{(s)}(\theta)]$ on a neighborhood of $\theta_0$ from the average of 20,000 realizations of $\nabla_\theta  L_{Cox}^{(s)}(\theta)$. Each realization consists of $s$ i.i.d. time-to-event data $(X_i,T_i,\Delta_i)$ where $T_i^*$is from a Cox model with $f_0(X_i) = \theta_0X_i$ and $C_i^*$ is from an independent exponential distribution with censoring rate 45\%. 

\begin{figure}[!ht]
\centering
\begin{subfigure}{0.9\textwidth}
\centering
\includegraphics[width=0.4\textwidth]{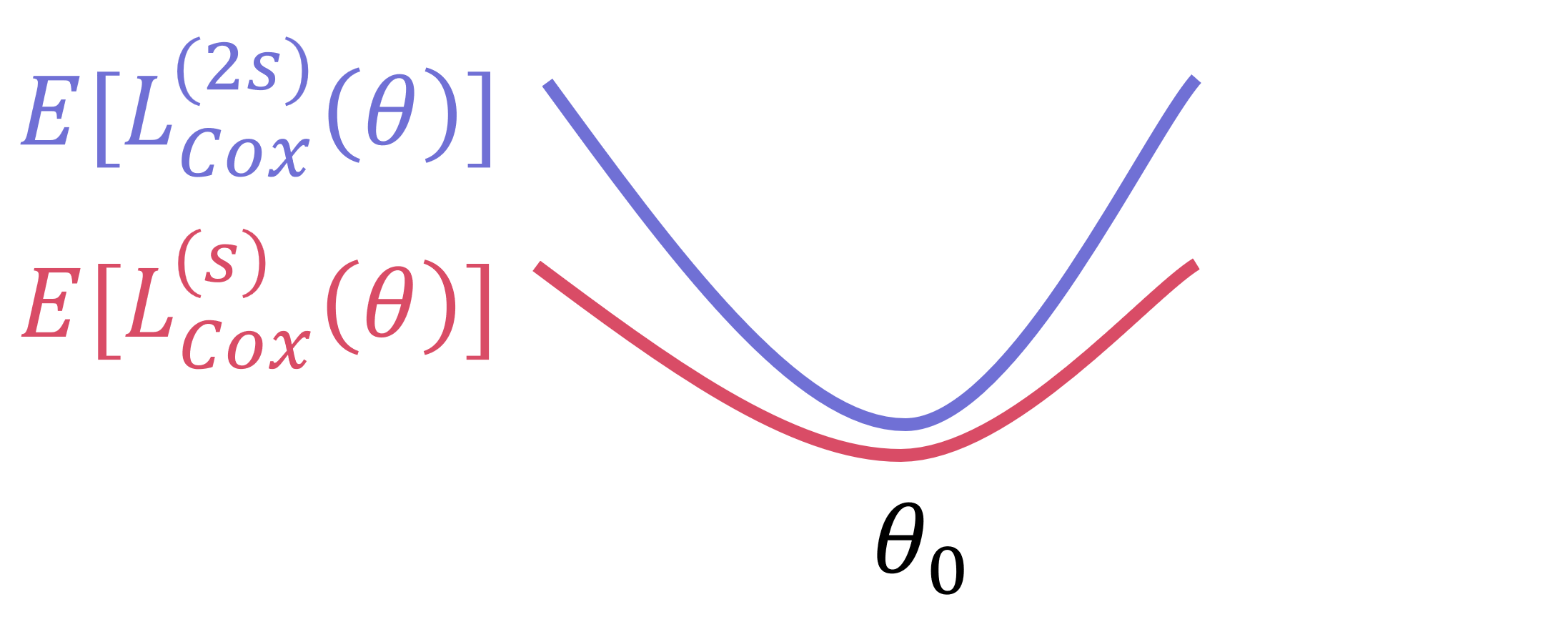}
\caption{}
\label{fig:double_batch_size}
\end{subfigure}
\begin{subfigure}{0.9\textwidth}
\centering
\includegraphics[width=1\textwidth]{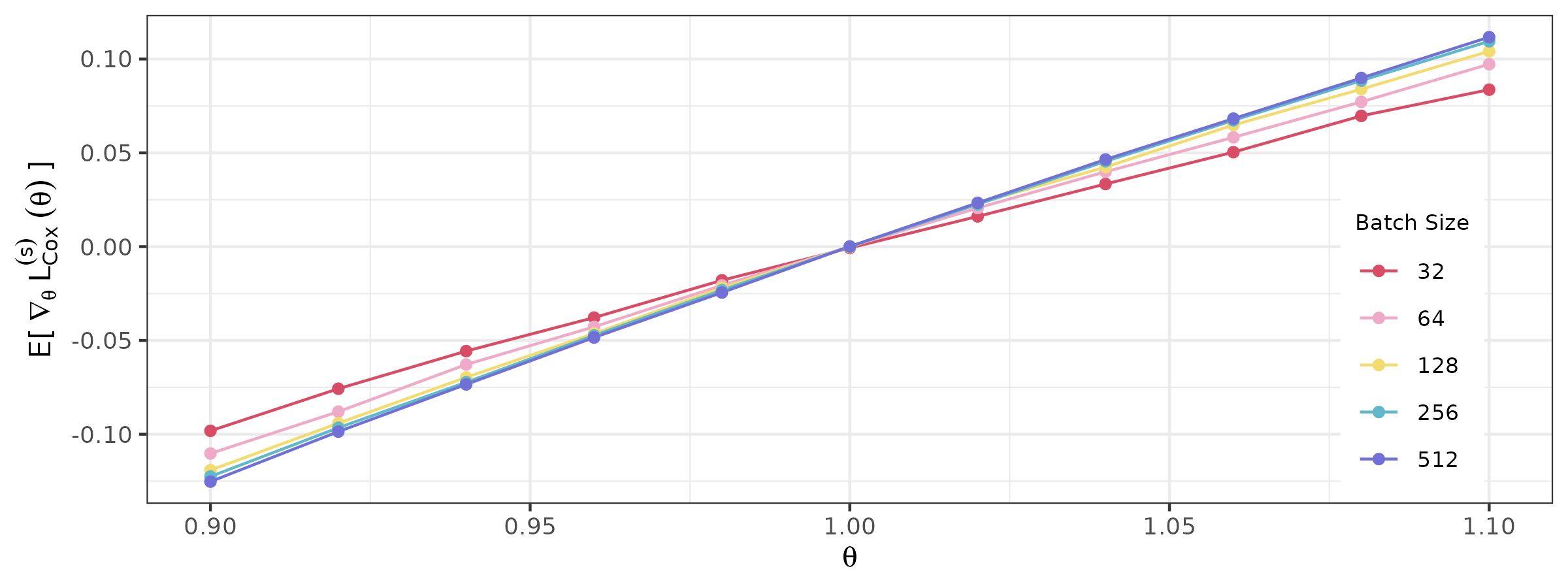}
\caption{}
\label{fig:simulation_population_gradient}
\end{subfigure}
\caption{\textbf{(a)} An illustrative picture showing the properties of ${\mathbb{E}} [L_{Cox}^{(s)}(\theta)]$ in Cox-regression: ${\mathbb{E}} [L_{Cox}^{(s)}(\theta)]$ reaches the minimum at $\theta_0$ regardless the choice of $s$ while its local convexity at $\theta_0$ increases when $s$ doubles. \textbf{(b)} Estimated ${\mathbb{E}} [\nabla_\theta  L_{Cox}^{(s)}(\theta)]$ at a neighborhood of $\theta_0=1$ with different batch sizes $s$. Each estimation is based on 20,000 realizations of the mini-batch data consisting of $s$ i.i.d. samples generated from a Cox model with $f_0(X) = \theta_0X$. }
\end{figure}

 Figure (\ref{fig:simulation_population_gradient}) presents ${\mathbb{E}} [\nabla_\theta  L_{Cox}^{(s)}(\theta)]$ in a neighborhood of $\theta_0=1$ (i.e., $\theta \in [0.90, 1.10]$) with different batch sizes $s$. As shown in Figure (\ref{fig:simulation_population_gradient}), $\theta_0 = 1$ is always the root of ${\mathbb{E}} [\nabla_\theta  L_{Cox}^{(s)}(\theta)]$ regardless the choice of $s$. Moreover, it verifies $\mathbb{E} [\nabla^2_\theta  L_{Cox}^{(2s)}(\theta)]|_{\theta = \theta_0}\geq \mathbb{E} [ \nabla^2_\theta L_{Cox}^{(s)}(\theta)]|_{\theta = \theta_0}$, given the increase in the slope of ${\mathbb{E}} [\nabla_\theta  L_{Cox}^{(s)}(\theta)]$ at $\theta_0=1$ when the batch size doubles. The increment becomes negligible for large $s$, as discussed in Remark \ref{remark: finite-property-pl}.

\subsection{Impact of Batch Size in Cox Regression\label{sec: Impact of Batch-size in Offline Cox Regression}}

We carried out simulations with 200 runs to empirically assess the impact of batch size in Cox regression. The true event time $T^*_i$ is generated from $\lambda(t|X_i) = \lambda_0(t)\exp(X_i^T\theta_0)$ where $\lambda_0(t) = 1$, $\theta_0 = \textbf{1}_{10\times 1}$, and $X_{ip}\displaystyle\operatorname*{\sim}^{i.i.d.}$ Uniform(0,1) for $p \in \{1,2,\dots,10\}$. The true censoring time $C^*_i$ is generated from an independent exponential distribution with a censoring rate $30\%$. We performed projected SGD (\ref{eq:projected-SGDupdate for Cox}) to estimate $\theta_0$ based on $n = 2,048$ samples. The SGD batch size is $2^k$ where $k=2,\dots,9$. The total number of epochs (train the model with all the training data for one cycle) is $200$ and the learning rate is set as $\gamma_E = \frac{2^{k-5}}{E+1}$ at epoch $E$, which is proportional to the batch size and decreases after each epoch.  Besides SGD-SB and SGD-FB, we fit a stratified Cox model (CoxPH-strata) by treating the fixed batches from SGD-FB as strata. Note that CoxPH-strata directly solves Eq. ($\ref{eq:FBGD estimator}$) using the GD algorithm. The convergence of the SGD algorithm can be evaluated by comparing the estimators from SGD-FB and CoxPH-strata. The MPLE (CoxPH) was also fitted.

\begin{figure}[ht]
\centering
\includegraphics[width=0.8\textwidth]{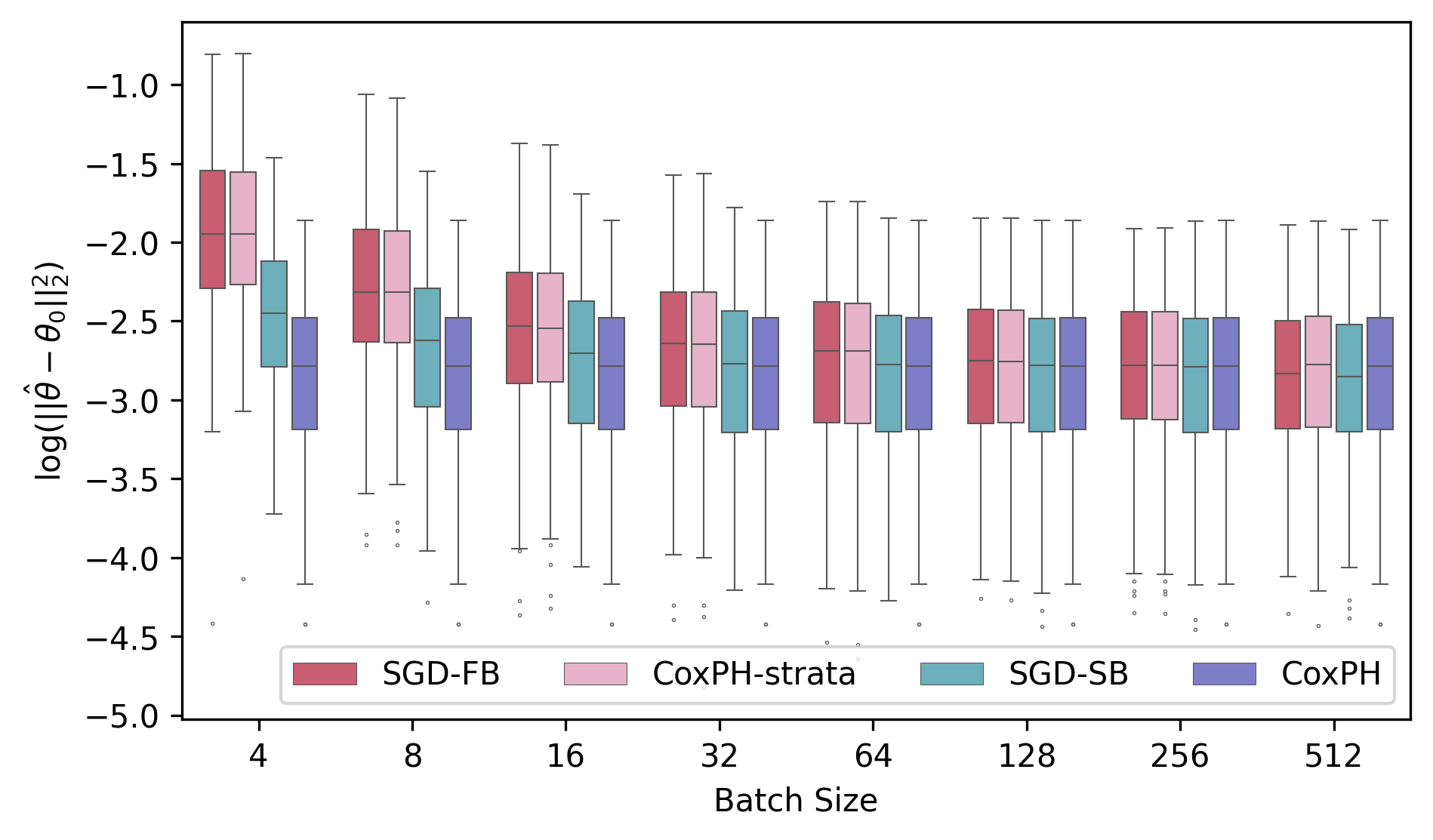}
\caption{Boxplots of $\log(\lVert \hat\theta-\theta_0\rVert_2^2)$ over $200$ runs with sample size $n=2,048$ where $\hat\theta$ is solved by four different methods. SGD with two different batch sampling strategies is considered, either with a fixed batch sampling strategy (FB) or with a stochastic batch sampling strategy (SB). CoxPH-strata is a stratified Cox model treating the fixed batches from SGD-FB as strata and serves as the global minimizer for SGD-FB.}
\label{fig:PH-reg-simulation}
\end{figure}

The simulation results are presented in Figure \ref{fig:PH-reg-simulation}. The convergence of SGD is validated by the negligible difference between the mb-MPLE $\hat\theta_t^{FB(s)}$ and the estimator from CoxPH-strata $\hat\theta^{strata}$ where $\log(\lVert \hat\theta^{FB(s)}-\hat\theta^{strata} \rVert_2^2)<-7.5$ for all $s$ throughout the simulations. Since the bias is small, the level of $\lVert \hat\theta-\theta_0\rVert_2^2$ primarily reflects the variance of the estimator (see Supplementary Material for the bias, empirical variance, and asymptotic variance of the estimators). Figure \ref{fig:PH-reg-simulation} shows that SGD-SB is more efficient than SGD-FB, especially when the batch size is small. There is an efficiency loss for both SGD-SB and SGD-FB compared to CoxPH. The efficiency loss becomes negligible when the batch size is large.

\subsection{Linear Scaling Rule for Cox Neural Network\label{sec: Learning Rate Scaling Rule for Cox Neural Network}}
\begin{figure}[!ht]
\centering
\includegraphics[width=0.8\linewidth]{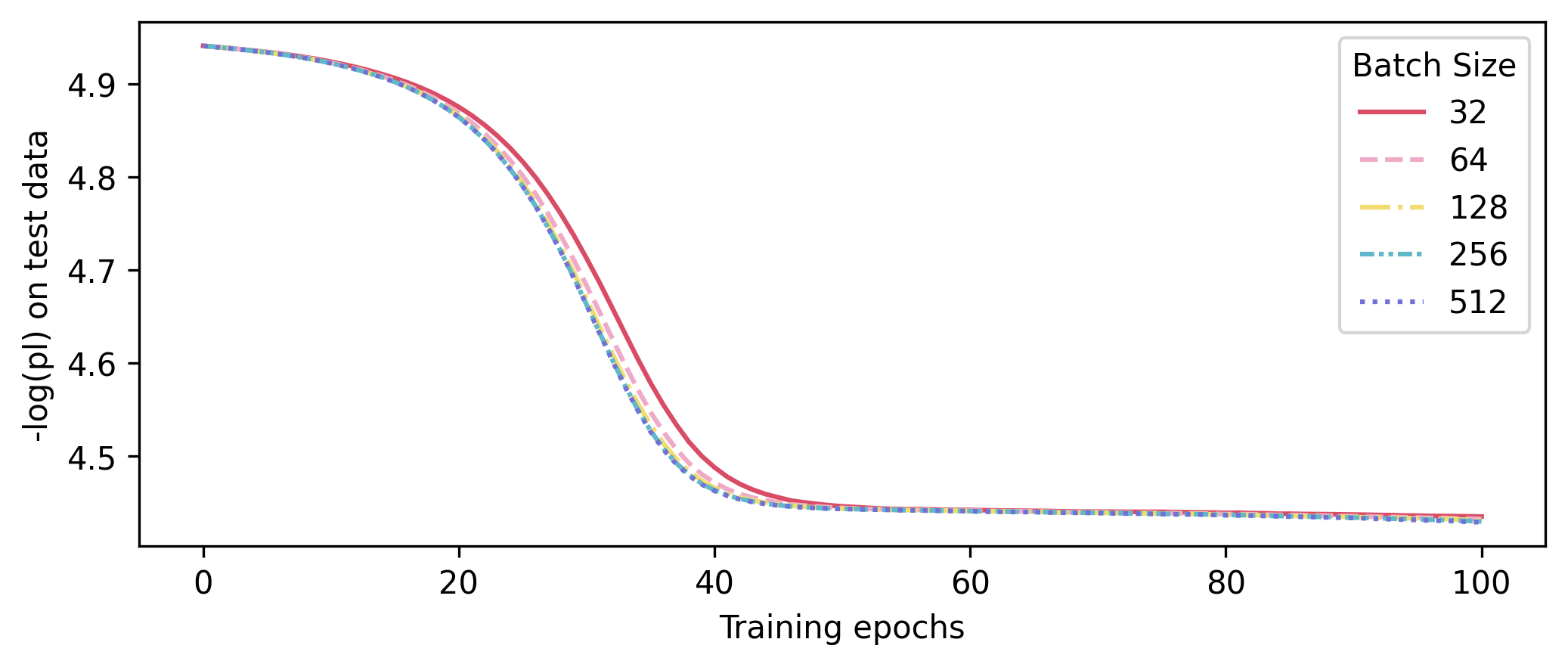}
\caption{The negative log-partial likelihood $L_{Cox}^{(N_{test})}(\theta)$ evaluated on a test data ($N_{test}= 2,048$) over the training epochs. The learning rate $\gamma$ is 0.1/16 when the batch size is 32. $\gamma$ is doubled when doubling the batch size. All the other hyperparameters are kept the same.}
\label{fig:PH-NN-LS}
\end{figure}

We further evaluated whether keeping the ratio of the learning rate to the batch size constant makes the SGD training process unchanged in Cox-NN. The data generating mechanism is the same as in Section \ref{sec: Impact of Batch-size in Offline Cox Regression} except that true event time $T^*_i$ is generated from a Cox model with
\begin{equation*}
f_0(X) = X_1^2X_2^3+\log(X_3+1)+\sqrt{X_4X_5+1}+\exp(X_5/2)-8.6,\ X_p\sim U(0,1), \ p=1,...,5 
\end{equation*} where $-8.6$ is used to center $\mathbb{E}[f_0(X)]$ towards zero. A Cox-NN is fitted under different SGD learning rates and batch sizes. The full negative log-partial likelihood $L_{Cox}^{(N_{test})}$ was calculated on the same test data to reflect the training history. Identical to the spirit of \citet{goyal2017accurate}, our goal is to match the test errors across batch sizes by only adjusting the learning rate. Figure \ref{fig:PH-NN-LS} shows that the linear scaling rule still holds in Cox-NN, especially when batch size is large (the three curves of batch size $s=128,\ 256,\ 512$ are overlapped with each other). The differences between the curves (e.g., batch sizes 32 and 64) are due to the convexity change of the loss function when performing SGD with different batch sizes. 

\section{Real-world Data Analysis}\label{sec: Real Data Analysis}

\begin{figure}[!ht]
\centering
\includegraphics[width=0.8\textwidth]{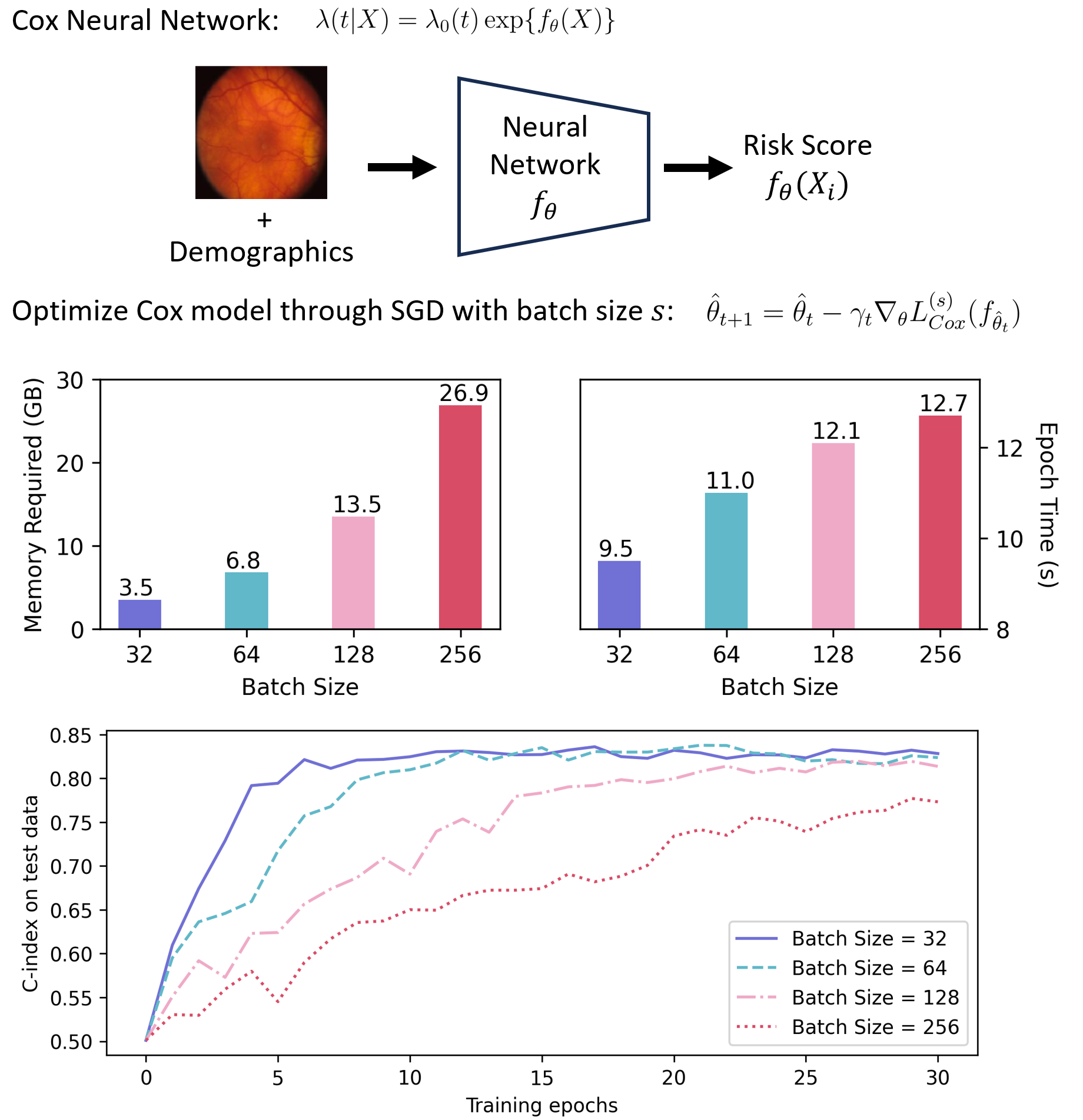}
\caption{First panel: the structure of the Cox-NN model optimized by SGD to predict the time-to-AMD progression based on the fundus image and demographics; second panel: the required memory and running time of SGD over different batch sizes; third panel: the C-index (on the test data) over training epochs under the choice of different batch sizes.}
\label{fig:AREDS application}
\end{figure}

We applied the Cox-NN on the Age-Related Eye Disease Study (AREDS) data \citep{age1999age} and built a prediction model for a progressive eye disease, Age-related Macular Degeneration (AMD). The analysis data set includes 7,865 eyes of 4,335 subjects, with the outcome of interest being time from enrollment (i.e., baseline) to progression to late AMD. The main predictor is the colored fundus image taken at the baseline. Other predictors include demographic variables (age at enrollment, educational level, and smoking status). There have been several existing works using fundus images to predict AMD progression. For example, \citet{peng2020predicting} implemented a two-step approach, where the first step is to fit a convolutional neural network on images with a binary outcome (progress vs not progress) to obtain a lower-dimensional predictor, and then the second step is to perform a Cox regression with this lower-dimensional image-based predictor from the first stage (together with other predictors) to predict time-to-progression. Our goal is to build a one-step prediction model by implementing Cox-NN directly on fundus images. The size of the raw fundus image is $3\times 2300\times 3400$. We cropped out the middle part of the fundus image and resized it to $3\times 224\times 224$. In this application, the Cox-NN employed the ResNet50 structure \citep{he2016deep} to take the fundus image as input (shown in the first panel in Figure \ref{fig:AREDS application}), which was optimized through SGD on a training set (7,087 samples) using a Nvidia L40s GPU with 48 GB memory and then evaluated the concordance index (C-index) by \citet{harrell1982evaluating} in a separate test set (778 samples) to measure the predictive performance. 

We first investigate the performance of SGD when optimizing a given Cox-NN (details in the Supplementary Material) under different batch sizes with a fixed learning rate ($\gamma=0.002$). The second panel of Figure \ref{fig:AREDS application} presents the memory required to perform SGD with different choices of batch size. The memory required for SGD increases approximately linearly as the batch size grows, and is already 26.9 GB for a batch size of 256. The GD is equivalent to setting the batch size to 7,087 and is therefore not feasible in this application. Moreover, SGD with a smaller batch size leads to a shorter time to run an epoch and a faster learning process (as presented in the second and third panels of Figure~\ref{fig:AREDS application}).

\begin{figure}[!ht]
\centering
\includegraphics[width=0.8\textwidth]{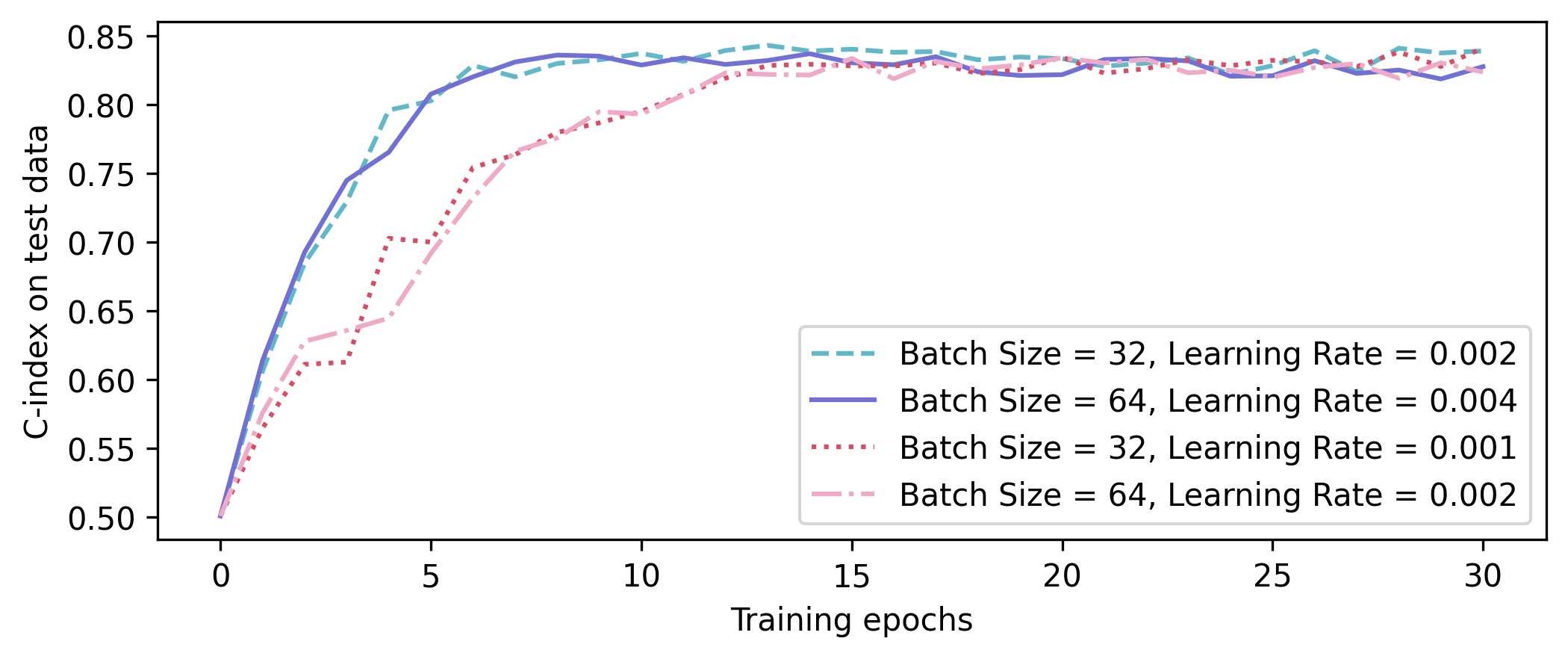}
\caption{The C-index on test data of Cox-NN over training epochs in the AREDS application. The Cox-NN is optimized by SGD with different choices of batch size and learning rate. All the other hyperparameters are fixed.}
\label{fig:AREDS_CoxSNN}
\end{figure}

Next, we verified the linear scaling rule by adjusting both the batch size and the learning rate of the SGD algorithm. From the trajectories of the C-index over the training epochs in Figure \ref{fig:AREDS_CoxSNN}, the SGD training history would be similar when $\gamma / s$ is the same. Moreover, reducing the batch size by half is equivalent to doubling the learning rate, which leads to faster convergence. For example, decreasing the batch size from 64 to 32 with a learning rate of 0.002 (dash-dotted line to dashed) or increasing the learning rate from 0.002 to 0.004 with a batch size of 64 (dash-dotted line to solid line) generates similar training trajectories. This justifies our discussion in Section \ref{sec:impact of Mini-batch Size in Cox-NN Optimization} that the ratio of learning rate to batch size $\gamma /s$ determines the dynamics of SGD in Cox-NN training.

Lastly, we built a prediction model by fine-tuning the hyperparameters, such as the Cox-NN structures and SGD parameters, through a grid search. Guided by the linear scaling rule, we fixed the batch size to 32 and only tuned the learning rate. Detailed configurations of the hyperparameters are displayed in the Supplementary Material. To tune the hyperparameters, we held out a validation set (20\%) from the 7,087 training data and evaluated the C-index on the validation set after training with different configurations. We chose the hyperparameters that maximized the C-index on the validation set for our final model, which achieved a C-index of 0.85 on the test data.

\section{Discussion\label{sec:discussion}}

This paper studies the statistical properties of the mb-MPLE for deep Cox models, establishing its consistency and optimal convergence rate. We show that the SGD seeks to optimize a function that depends on batch size, which differs from the all-sample partial likelihood. The paper investigates the properties of this batch-size-dependent function and presents the impact of SGD batch size in Cox-NN and regression. \textcolor{black}{While we analyzed the optimization error of SGD in Cox regression, extending this analysis to Cox-NN is beyond the scope of the current work and remains an important direction for future research.}

The partial likelihood in the Cox model essentially models the rank of event times through a Plackett-Luce (PL) model \citep{plackett1975analysis,luce1959individual} with censored outcome data. PL model with neural networks has been widely used in various tasks, such as learning-to-rank (LTR) \citep{cao2007learning} and contrastive learning \citep{chen2020simple}. These applications are equivalent to applying Cox-NN to time-to-event data without right-censoring. Thus, our results and discussions can be extended to these tasks as a potential direction for future research.




\bibliographystyle{agsm}
\bibliography{reference}

\end{document}


\def\spacingset#1{\renewcommand{\baselinestretch}%
{#1}\small\normalsize} \spacingset{1}


\if1\blind
{
  \title{\large \bf Mini-batch Estimation for Deep Cox Models: \\Statistical Foundations and Practical Guidance \\ Supplementary Material}
  \author{Lang Zeng \\
    Department of Biostatistics and Health Data Science, University of Pittsburgh\\
    \\
    Weijing Tang \\
    Department of Statistics and Data Science, Carnegie Mellon University\\
    \\
    Zhao Ren \\
    Department of Statistics, University of Pittsburgh\\
    \\
    Ying Ding\thanks{Corresponding Author Email: yingding@pitt.edu}\hspace{.2cm}\\
    Department of Biostatistics and Health Data Science, University of Pittsburgh}
  \date{\vspace{-5ex}}
  \maketitle
} \fi

\if0\blind
{
  \bigskip
  \bigskip
  \bigskip
  \begin{center}
  {\large \bf Mini-batch Estimation for Deep Cox Models: \\Statistical Foundations and Practical Guidance} \\Supplementary Material
\end{center}
  \medskip
} \fi

\spacingset{1.9} 

{
\let\clearpage\relax
\tableofcontents
}

\section{Notation and Assumptions}

In this section, we introduce the notation and assumptions used for the proofs. Separate notations and assumptions are introduced in Section \ref{sec:PLNN} for the partially linear Cox model.

\subsection{Notation}
Let $D(n) = \{D_i\}_{i=1}^n$ denote $n$ observed independent and identical distributed (i.i.d.) samples. Survival analysis is to understand the effect of covariate $X\in\mathbb{R}^{p}$ on the time-to-event variable $T^*$ (e.g., time-to-death). Let $C^*_i$ denote the true censoring time. Due to the right censoring, the observed time-to-event data is the triplet set $D_i= (X_i, T_i,\Delta_i)$, where $T_i=\min(T^*_i, C^*_i)$ is the observed time and $\Delta_i=I(T^*_i \leq C^*_i)$ is the event indicator. $Y_i(u)$ is the at-risk process representing whether patient $i$ is at
risk at time $u$, $Y_i(u) = I(T_i \geq u)$ and $Y^{(s)}(u) = \sum_{i=1}^s Y_i(u)$. $N_i(u) = \Delta_i I(T_i\leq u)$ is the counting process for subject $i$ and $N^{(s)}(u) = \sum_{i=1}^s N_i(u)$.

We use $\otimes$ to denote the outer product such that $X_1^{\otimes 2} = X_1\otimes X_1=X_1X_1^T$ and $X_1\otimes X_2 = X_1X_2^T$. When $X_1$ and $X_2$ are real values, it is the real number multiplication such that $X_1^{\otimes 2} = X_1^2$ and $X_1\otimes X_2 = X_1X_2$.

For square matrices $A$ and $B$, we use $A \succeq B$ to denote $A-B$ is positive semi-definite and use $A\succ B$ to denote $A-B$ is positive definite.

$\mathcal{F}_u^{(s)}$ is the filtration that includes all information up to time $u$ such that
$$\mathcal{F}_u^{(s)} = \sigma(\{(N_i(t),Y_i(t),X_i),i\in(1,\dots,s),\forall t<u\}).$$

For any vector $v=(v_1,\dots,v_p)^T\in\mathbb{R}^p$, $\lVert v\rVert=(\sum_{i=1}^pv_i^2)^{1/2}$ and $\lVert v\rVert_\infty = \max_i\lvert v_i\rvert$, and for any matrix $W=(w_{ij})\in\mathbb{R}^{m\times n}$, $\lVert W\rVert_\infty=\max_{i,j}\lvert W_{ij}\rvert$. For any function $h$, $\lVert h\rVert_\infty$ and $\lVert h\rVert_{L_2}$ are the sup-norm and $L^2$-norm of $h$, respectively. For any $f_1,f_2$, define $d(f_1,f_2):=\lVert f_1-f_2\rVert_{L_2} = [\mathbb{E}_X\{f_1(X)-f_2(X)\}^2]^{1/2}$. Some empirical process notation will be used, such as $\mathbb{P}f := \mathbb{E}_X{[f]} := \int_{X\in\mathcal{X}} f(x) dF_X(x)$ and $\mathbb{P}_n f = \frac{1}{n}\sum_{i=1}^n f(X_i)$. For any vector function $\mathbf{h}=(h_1.\dots,h_p)^T$, $\lVert \mathbf{h}\rVert_\infty=\max_i \lVert h_i\rVert_\infty$. Denote $a_n\lesssim b_n$ as $a_n\leq cb_n$ for some $c>0$ and any $n$. And $a_n \asymp b_n$ means $a_n\lesssim b_n$ and $b_n\lesssim a_n$.

The partial likelihood of $s$ samples is
\begin{align*}
L_{Cox}^{(s)}(\theta)  = & -\frac{1}{s}\sum_{i=1}^s \Delta_i \log \frac{\exp\{f_\theta(X_i)\}}{\sum_{T_j \geq T_i} \exp\{f_\theta(X_j)\}} \\
= &\int_0^\tau \frac{1}{s}\sum_{i=1}^s \left\{ -f_{\theta}(X_i)+\log \left(\sum_{j=1}^s Y_j(u)\exp\{f_{\theta}(X_j)\}\right) \right\} \mathrm{d}N_i(u)    
\end{align*}

Define
\begin{equation}
\begin{split}
L_{Cox}^{(s)}(f)& :=L_{Cox}^{(s)}(D(s);f) \\
& := \frac{1}{s}\sum_{j:D_j\in D(s)}\Delta_j\left [-f(X_j)+\log \sum_{k:D_k\in D(s)}Y_k(T_j)\exp(f(X_k))\right]
\end{split}
\label{eq:define L_Cox_f}
\end{equation}
and
\begin{equation*}
\begin{split}
L^{(s)}_{n}(f):= &\frac{1}{\binom{n}{s}}\sum_{D(s)\subset D(n)} L_{Cox}^{(s)}(f)\\
L^{(s)}_0(f):=& \mathbb{E}\left[L_{Cox}^{(s)}(f)\right ].
\end{split}
\end{equation*}
Let 
\begin{align*}
\mathcal{F}(K,\varsigma,\mathbf{p},\mathcal{D})=\{& f:f\text{ is a DNN with (K+1) layers and width vector $\mathbf{p}$ such that} \\ 
& \max\{\lVert W_k\rVert_\infty,\lVert v_k\rVert_\infty\}\leq 1\}\text{ , for all $k=0,\dots,K$}, \\
& \sum_{k=1}^K\lVert W_k\rVert_0+\lVert v_k\rVert_0\leq\varsigma,\lVert f\rVert_\infty < \mathcal{D}\},
\end{align*}
where $\lVert \cdot \rVert_0$ is the number of nonzero entries of a matrix or vector.

Consider the NN estimator
\begin{equation}
\Tilde{f}^{(s)}_n = \displaystyle\arg \min_{f\in \mathcal{F}(K,\varsigma,\mathbf{p},\mathcal{D})} \frac{1}{\binom{n}{s}}\displaystyle\sum_{D(s)\subset D(n)}L_{Cox}^{(s)}(f) = \displaystyle\arg \min_{f\in \mathcal{F}(K,\varsigma,\mathbf{p},\infty)} L^{(s)}_{n}(f).
\label{eq:SGD NN minimizer}
\end{equation}

Let $\mathcal{H}^{\alpha}_{r}(\mathbb{D},M)$ be a H\"older class of smooth functions with parameters $\alpha,M>0$ and domain $\mathbb{D}\subseteq\mathbb{R}^r$ defined by
$$\mathcal{H}^{\alpha}_{r}(\mathbb{D},M) = \left\{h:\mathbb{D}\to\mathbb{R}:\sum_{\beta:\lvert\beta\rvert<\alpha}\lVert\partial^\beta h\rVert_\infty+\sum_{\beta:\lvert\beta\rvert=\lfloor \alpha\rfloor}\sup_{x,y,\in\mathbb{D},x\neq y}\frac{\lvert \partial^\beta h(x) - \partial^\beta h(y) \rvert}{\lVert x-y \rVert_\infty^{\alpha-\lfloor \alpha\rfloor}}\leq M \right\},$$
where $\lfloor \alpha\rfloor$ is the largest integer strictly smaller than $\alpha$, $\partial^\beta:=\partial^{\beta_1}\dots \partial^{\beta_r}$ with $\beta = (\beta_1,\dots,\beta_r)$, and $\lvert\beta\rvert = \sum_{k=1}^r \beta_k$. 

Let $q\in \mathbb{N}$, $M>0$, $\vec{\alpha} = (\alpha_0,\dots,\alpha_q)\in \mathbb{R}_+^{q+1}$ and $\textbf{d}=(d_0,\dots,d_{q+1})\in \mathbb{N}_+^{q+2}$, $\mathbf{\Tilde{d}}=(\Tilde{d}_0,\dots,\Tilde{d}_{q})\in \mathbb{N}_+^{q+1}$ with $\Tilde{d}_j\leq d_j,j=0,\dots q$, where $\mathbb{R}_+$ is the set of all positive real numbers. We consider the unknown true function to belong to a composite smoothness function class:
\begin{align*}
\mathcal{H}(q,\vec{\alpha},\mathbf{d},\mathbf{\Tilde{d}},M):=\{& h = h_q\circ \dots \circ h_0: h_i = (h_{i1},\dots,h_{id_{i+1}})^T \text{ and }\\
& h_{ij}\in\mathcal{H}^{\alpha_i}_{\Tilde{d}_i}([a_i,b_i]^{\Tilde{d}_i},M),\text{ for some }\lvert a_i \rvert,\lvert b_i \rvert \leq M\}.
\end{align*}

Furthermore, we denote $\Tilde{\alpha}_i=\alpha_i\prod_{k=i+1}^q(\alpha_k \wedge 1)$ and $\Upsilon_n=\max_{i=0,\dots,q} n^{-\Tilde{\alpha}_i/(2\Tilde{\alpha}_i+\Tilde{d}_i)}$ with notation $a\wedge b :=\min\{a,b\}$.

\subsection{Assumptions}
\begin{itemize}
    \item [(A1)] The failure time $T_i^*$ and censoring time $C_i^*$ are independent given the covariates $X_i$.
    \item [(A2)] There is a truncation time $\tau < \infty $ such that, for some constant $\delta>0$, $\mathbb{P}(T^* >  \tau|X)\geq \delta$ and $\mathbb{P}(\Delta=1|X)\geq \delta$ almost surely with respect to the probability measure of $X$. The stochastic integrals computed from here on will be truncated at this value $\tau$.
    \item [(A3)] $X$ takes value in a compact subset of $\mathbb{R}^{p}$ with probability density function bounded away from zero. Without loss of generality, we assume that the domain of $X$ is taken to be $[0,1]^{p}$.
    \item[(N1)] The unknown function $f_0$ is an element of $\mathcal{H}_0=\{f \in \mathcal{H}(q,\vec{\alpha},\mathbf{d},\mathbf{\Tilde{d}},M):\mathbb{E}[f(X)] = 0\}.$
    \item[(N2)] $K=O(\log n)$, $\varsigma=O(n\Upsilon_n^2\log n)$, $\mathcal{D}>2M$, and $n\Upsilon_n\lesssim \displaystyle\min_{k=1,\dots,K} (p_k)\leq \displaystyle\max_{k=1,\dots,K} (p_k) \lesssim n$.
    \item[(R1)] The true relative risk $f_0(X) = \theta_0^TX$ with parameter $\theta_0$ is an interior point of $\mathbb{R}_M^{p} :=\{\theta \in \mathbb{R}^{p}:\lVert\theta\rVert\leq M\}$.
    \item[(R2)] There exist constants $0<c_1<c_2<\infty$ such that the subdensities $p(x,t,\Delta = 1)$ of $(X,T,\Delta=1)$ satisfies $c_1<p(x,t,\Delta = 1)<c_2$ for all $(x,t)\in [0,1]^{p}\times[0,\tau]$.
\end{itemize}

(A1) ensures that the censoring mechanism is noninformative. (A2) is a technical assumption to avoid the unboundedness of the partial likelihood and the partial score functions at the end point of the support of the observed event time \citep{huang1999efficient}. It also ensures the probability of being uncensored is positive regardless of the covariate value. (A1)-(A3) are the standard assumptions adopted in survival analysis \citep{fleming2013counting,zhong2022deep} and are required in all theorems in this work. (N1) and (N2) are used to develop the statistical properties for the neural network estimator in Cox-NN. (R1) and (R2) are used in Cox regression and are sufficient to demonstrate the strong convexity of $\mathbb{E} [L^{(s)}_{Cox}(\theta)]$.

\section{Proof of Theorems}

\subsection{Proof of Theorem 1}
\begin{theorem}\label{thm:convergence rate}
Under the Cox model, with assumptions (A1)-(A3), (N1)-(N2), for any integer $s\geq 2$, let $\Tilde{f}^{(s)}_n$ be defined in (\ref{eq:SGD NN minimizer}) and empirical average $\Bar{\Tilde{f}}^{(s)}_n:= \frac{1}{n}\sum_{i=1}^n\Tilde{f}^{(s)}_n(X_i)$, then
$$\lVert (\Tilde{f}^{(s)}_n-\Bar{\Tilde{f}}^{(s)}_n)-f_0 \rVert_{L^2([0,1]^{p})}=O_p(\Upsilon_n\log^2 n).$$
\end{theorem}

\begin{proof}
To show
$$\lVert \Tilde{f}^{(s)}_n-\frac{1}{n}\sum_{i=1}^n\Tilde{f}^{(s)}_n(X_i)-f_0 \rVert_{L^2([0,1]^{p})}=O_p(\Upsilon_n\log^2 n),$$
we will first show that 
$d(\Tilde{f}^{(s)}_n-\mathbb{E}_X[\Tilde{f}^{(s)}_n],f_0)=O_p(\Upsilon_n\log^2n),$
where $\mathbb{E}_X[f]:=\int_{[0,1]^p}f(x)dP(x)$. Some commonly used empirical process notations will be used, such as $\mathbb{P}f:=\mathbb{E}_X[f]$ and $\mathbb{P}_n f:=\frac{1}{n}\sum_{i=1}^nf(X_i).$

Consider $$\mathcal{G} := \mathcal{G}_{D(s)} = \left\{g(D(s);f) = L^{(s)}_{Cox}(f): f\in \mathcal{F}\right\},
$$ and we have $\mathbb{E}[g(D(s);f)] = L^{(s)}_0(f)$.

Let $U^{(s)}_{n}(g) := \frac{1}{\binom{n}{s}}\sum_{D(s)\subset D(n)} g(D(s);f)=L^{(s)}_{n}(f).$
For any $f_1, f_2\in \mathcal{F}$
\begin{equation*}
\label{eq: L1 relation}
\begin{split}
& \sup_{D(s)} \lvert g (D(s);f_1) - g(D(s);f_2)\rvert \\ 
\leq & \sup_{D(s)}\left [ \frac{1}{s} \sum_{j:D_j\in D(s)} \Delta_j \lvert f_1 (X_j) - f_2(X_j)\rvert\right ]+\\
& \sup_{D(s)}\left [ \frac{1}{s} \sum_{j:D_j\in D(s)}\Delta_j\left \lvert \log \sum_{k:D_k\in D(s)}Y_k(T_j)\exp(f_1(X_k))-\log \sum_{k:D_k\in D(s)}Y_k(T_j)\exp(f_2(X_k)) \right \rvert \right ]\\
(1)\leq & \sup_{X} \lvert f_1 (X) - f_2(X)\rvert+\\
& \sup_{D(s),j:D_j\in D(s)} \left \lvert \log \sum_{k:D_k\in D(s)}Y_k(T_j)\exp(f_1(X_k))-\log \sum_{k:D_k\in D(s)}Y_k(T_j)\exp(f_2(X_k)) \right \rvert \\
(2) \lesssim & \sup_{X} \lvert f_1 (X) - f_2(X)\rvert+\\
& \sup_{D(s),j:D_j\in D(s)} \left \lvert \sum_{k:D_k\in D(s)}Y_k(T_j)\exp(f_1(X_k))- \sum_{k:D_k\in D(s)}Y_k(T_j)\exp(f_2(X_k)) \right \rvert \\
(3) \leq & \sup_{X}\lvert f_1 (X) - f_2(X)\rvert+\sup_{D(s)}\sum_{k:D_k\in D(s)}\left \lvert \exp(f_1(X_k))-\exp(f_2(X_k))\right \rvert \\
(4) \lesssim & \sup_{X}\lvert f_1 (X) - f_2(X)\rvert+\sup_{D(s)}\sum_{k:D_k\in D(s)}\left \lvert f_1(X_k)-f_2(X_k)\right \rvert \\
 \lesssim & \sup_{X}\lvert f_1 (X) - f_2(X)\rvert.
\end{split}
\end{equation*}
Inequality (1) follows from the triangle inequality, the fact that the samples are i.i.d., and that $\Delta$ takes values only in $\{0,1\}$. Inequality (2) follows from the boundedness of $\sum_{k:D_k\in D(s)} Y_k(T_j)\exp(f(X_k))$ and the fact that the logarithm function is Lipschitz on a bounded interval. Inequality (3) follows from the triangle inequality and the fact that $Y$ takes values only in $\{0,1\}$. Inequality (4) follows from the boundedness of $f$ and the fact that the exponential function is Lipschitz on a bounded interval.

This implies that
$\mathcal{N}(\epsilon,\mathcal{G},\lVert\cdot\rVert_\infty)\lesssim \mathcal{N}(\epsilon,\mathcal{F},\lVert\cdot\rVert_\infty)$, where $\mathcal{N}(\epsilon,\mathcal{G},\lVert\cdot\rVert_\infty)$ is the $\epsilon$-covering number of $\mathcal{G}$ under sup-norm. From the proof of Lemma 6 in \cite{zhong2022deep}, we have the $\epsilon$-bracket number $\mathcal{N}_{[]}(\epsilon,\mathcal{F},\lVert\cdot\rVert_\infty)\lesssim (\frac{U}{\epsilon})^{\varsigma+1}$ so that $\mathcal{N}(\epsilon,\mathcal{G},L_{1}(P^s_n)) \leq \mathcal{N}(\epsilon,\mathcal{G},\lVert\cdot\rVert_\infty)\leq \mathcal{N}(\epsilon,\mathcal{F},\lVert\cdot\rVert_\infty) \leq \mathcal{N}_{[]}(\epsilon,\mathcal{F},\lVert\cdot\rVert_\infty) \lesssim (\frac{U}{\epsilon})^{\varsigma+1}$, where $U = K\prod_{k=0}^K(p_k+1)\sum_{k=0}^Kp_kp_{k+1}$. Therefore, the conditions of Corollary 3.2 in \cite{arcones1993limit} follows as $g\in\mathcal{G}$ is bounded and $\log\mathcal{N}(\epsilon,\mathcal{G},L_{1}(P^s_n))/n=O_p(\frac{\varsigma}{n}\log U)=O_p({\Upsilon_n^2\log n (\log\log n+\log^2 n)})=O_p(\Upsilon_n^2\log^{3} n)\overset{p}{\to}0$, and this implies that
$$\sup_{g\in\mathcal{G}}\lvert U^{(s)}_{n}(g)-\mathbb{E}[g]\rvert \overset{p}{\to} 0.$$
Thus,
\begin{equation*}
\label{eq:universal convergence0}
\sup_{f\in\mathcal{F}}\lvert L^{(s)}_{n}(f)-L^{(s)}_{0}(f)\rvert \overset{p}{\to} 0.
\end{equation*}
Because $L^{(s)}_{n}(f-\mathbb{E}_X[f])=L^{(s)}_{n}(f)$ for all $f\in\mathcal{F}$, therefore
\begin{equation}\label{eq:universal convergence}
\sup_{f\in\mathcal{F}}\lvert L^{(s)}_{n}(f-\mathbb{E}_X[f])-L^{(s)}_{0}(f)\rvert \overset{p}{\to} 0.
\end{equation}

Define
\begin{equation}
\label{ap-eq: f check}
\check{f} = \arg \min_{f\in \mathcal{F}(K,\varsigma,\mathbf{p},\mathcal{D})} \lVert f-f_0\rVert_{L^2}.
\end{equation}
By the proof of Theorem 1 in \cite{schmidt2020nonparametric}, we have $\lVert \check{f}-f_0\rVert_{L^2}=O(\Upsilon_n\log^2n)$. Moreover,
\begin{equation}\label{eq:Schmidt's result}
\begin{split}
\lVert \check{f}-\mathbb{E}_X[\check{f}]-f_0\rVert_{L^2}&=\lVert \check{f}-f_0-\mathbb{E}_X\{\check{f}-f_0\}\rVert_{L^2}\\
& \leq \lVert \check{f}-f_0\rVert_{L^2}+\lvert\mathbb{E}_X\{\check{f}-f_0\}\rvert\\
& \leq \lVert \check{f}-f_0\rVert_{L^2}+\lVert \check{f}-f_0\rVert_{L^1}\\
&\lesssim \lVert \check{f}-f_0\rVert_{L^2}=O(\Upsilon_n\log^2n).
\end{split}
\end{equation}
Notice that $L^{(s)}_{0}(f) = L^{(s)}_{0}(f+c)$ for any constant $c\in \mathbb{R}$, therefore $L^{(s)}_{0}(\check{f}) = L^{(s)}_{0}(\check{f}-\mathbb{E}_X[\check{f}])$. Then, by (\ref{eq:universal convergence}), Lemma \ref{lemma: minimizer does not depend on s}, (\ref{eq:Schmidt's result}) and law of large numbers, we have
\begin{align*}
\lvert L^{(s)}_{n}(\check{f})-L^{(s)}_{n}(f_0)\rvert & \leq \lvert L^{(s)}_{n}(\check{f})-L^{(s)}_{0}(\check{f})\rvert+\lvert L^{(s)}_{0}(\check{f})-L^{(s)}_{0}(f_0)\rvert+\lvert L^{(s)}_{0}(f_0)-L^{(s)}_{n}(f_0)\rvert\\
& = \lvert L^{(s)}_{n}(\check{f})-L^{(s)}_{0}(\check{f})\rvert+\lvert L^{(s)}_{0}(\check{f}-\mathbb{E}_X[\check{f}])-L^{(s)}_{0}(f_0)\rvert+\lvert L^{(s)}_{0}(f_0)-L^{(s)}_{n}(f_0)\rvert\\
& =o_p(1).
\end{align*}
Since $\Tilde{f}^{(s)}_n$ is the minimizer of (\ref{eq:SGD NN minimizer}), we have
\begin{equation}\label{eq: empirical loss at minimizer is smaller than at the truth}
L^{(s)}_{n}(\Tilde{f}^{(s)}_n-\mathbb{E}_X[\Tilde{f}^{(s)}_n])=L^{(s)}_{n}(\Tilde{f}^{(s)}_n)\leq L^{(s)}_{n}(\check{f}) = L^{(s)}_{n}(f_0)+o_p(1).
\end{equation}

Moreover, Lemma \ref{lemma: minimizer does not depend on s} implies that for any small $\epsilon>0$,
\begin{equation}\label{eq: loss reaches minimum at the f_0}
 L^{(s)}_{0}(f_0)<\inf_{\substack{d(f,f_0)\geq \epsilon,\\\mathbb{E}_X[f]=0}} L^{(s)}_{0}(f).
\end{equation}

Therefore, with $\mathbb{E}_X[\Tilde{f}^{(s)}_n-\mathbb{E}_X[\Tilde{f}^{(s)}_n]] = 0$, the conditions of Theorem 5.7 in \citet{van2000asymptotic} follow from (\ref{eq:universal convergence}), (\ref{eq: empirical loss at minimizer is smaller than at the truth}) and (\ref{eq: loss reaches minimum at the f_0}), and this implies that $d(\Tilde{f}^{(s)}_n-\mathbb{E}_X[\Tilde{f}^{(s)}_n],f_0)\to 0$ as $n\to\infty$.

Next, we show the convergence rates $d(\Tilde{f}^{(s)}_n-\mathbb{E}_X[\Tilde{f}^{(s)}_n],f_0)=O_p(\tau_n)$ with $\tau_n = \Upsilon_n\log^2n$. Our strategy is to apply Theorem 3.2.5 in \cite{van2023weak} by verifying the following four conditions:
\begin{itemize}
    \item Let $\mathcal{B}_\delta=\{f-\mathbb{E}_X[f]:f\in\mathcal{F}, d(f_0,f-\mathbb{E}_X[f])\leq \delta\}$. For every $\delta>0$,
\begin{equation}\label{eq: outer-measure}
\mathbb{E}^*\sup_{f\in\mathcal{B}_\delta}\sqrt{n}\left\lvert (L^{(s)}_n- L^{(s)}_0)(f)-(L^{(s)}_n- L^{(s)}_0)(f_0)\right \rvert\lesssim \phi_n(\delta),
\end{equation}
where $\mathbb{E}^*$ is the outer measure and $\phi_n(\delta) = \delta\sqrt{\varsigma\log \frac{U}{\delta}}$ with $U = K\prod_{k=0}^K(p_k+1)\sum_{k=0}^Kp_kp_{k+1}$.

\item For every $f$ in a neighborhood of $f_0$,
\begin{equation} \label{ap-eq: convergence rate 2}
L_0^{(s)}(f)-L_0^{(s)}(f_0)\gtrsim  d^2(f,f_0). 
\end{equation}

\item The function $\delta \mapsto \phi_n(\delta)/\delta$  is decreasing and \begin{equation}\label{ap-eq: convergence rate 3}
\tau_n^{-2}\phi(\tau_n)\leq \sqrt{n}.
\end{equation}

\item The sequence $\Tilde{f}^{(s)}_n-\mathbb{E}_X[\Tilde{f}^{(s)}_n]$ converges in outer probability to $f_0$ and satisfies \begin{equation}\label{ap-eq: convergence rate 4}
L_n^{(s)}(\Tilde{f}^{(s)}_n-\mathbb{E}_X[\Tilde{f}^{(s)}_n])\leq L_n^{(s)}(f_0)+O_p(\tau_n^2).
\end{equation}
\end{itemize}

Recall that $L^{(s)}_n(f) = U^{(s)}_{n}(g)$ and $L^{(s)}_0(f) = \mathbb{E}[g]$ where 
\begin{align*}
U^{(s)}_{n}(g) =& \frac{1}{\binom{n}{s}}\sum_{D(s)\subset D(n)} g(D(s)_1,D(s)_2,\dots,D(s)_s;f) \\
=& \frac{1}{\binom{n}{s}}\sum_{D(s)\subset D(n)}\left\{\frac{1}{s}\sum_{j:D_j\in D(s)}\Delta_j\left [-f(X_j)+\log \sum_{k:D_k\in D(s)}Y_k(T_j)\exp(f(X_k))\right]\right \}.
\end{align*}
By Hoeffding decomposition for U-statistics,
$$(L^{(s)}_n- L^{(s)}_0)(f) = U^{(s)}_n(g)-\mathbb{E}[g]
= \sum_{c=1}^s\binom{s}{c} \Tilde{U}^{(c)}_n(f),$$
where
$$\Tilde{U}^{(c)}_n(f) = \binom{n}{c}^{-1}\sum_{1\leq l_1<\dots<l_c\leq n}\xi^{(c)}(D_{l_1},\dots,D_{l_c};f)$$
and 
\begin{align*}
\xi^{(c)}(D_{l_1},\dots,D_{l_c};f) =\sum_{k=0}^c(-1)^{c-k}\sum_{1\leq l_1<\dots<l_k\leq c}\mathbb{E}[g(D_{l_1},\dots,D_{l_k},D_{k+1},\dots,D_s;f)\lvert D_{l_1},\dots,D_{l_k}].
\end{align*}


Therefore,
\begin{equation}\label{ap-eq:hoeffding decomposition}
\begin{split}
(L^{(s)}_n- L^{(s)}_0)(f) 
& = \binom{s}{1}\Tilde{U}^{(1)}_n(f)+\sum_{c=2}^s\binom{s}{c} \Tilde{U}^{(c)}_n(f) \\
& = s(\mathbb{P}_n-\mathbb{P})\Tilde{g}^{(1)}(f)+\sum_{c=2}^s\binom{s}{c} \Tilde{U}^{(c)}_n(f),
\end{split}
\end{equation}
where $\Tilde{g}^{(1)}(D_1;f) = \mathbb{E}[g(D_1,D_2,\dots,D_s;f)\lvert D_1]$ and $\mathbb{E}[\Tilde{g}^{(1)}] = \mathbb{E}[g]$.

With (\ref{ap-eq:hoeffding decomposition}), by inspection of the proof of Theorem 3.2.5 in \citet{van2023weak}, to show $d(\Tilde{f}^{(s)}_n-\mathbb{E}_X[\Tilde{f}^{(s)}_n],f_0)=O_p(\tau_n)$, it suffices to change the condition (\ref{eq: outer-measure}) of the required four conditions to
\begin{equation}
\label{eq:leading term}
\mathbb{E}^*\sup_{f\in\mathcal{B}_\delta}\left[\sqrt{n}(\mathbb{P}_n-\mathbb{P})(\Tilde{g}^{(1)}(.;f)-\Tilde{g}^{(1)}(.;f_0))\right]^+ \lesssim \phi_n(\delta).
\end{equation}
and
\begin{equation}
\label{eq:remainder term}
\sup_{f\in\mathcal{F}}\left\lvert\sum_{c=2}^s\binom{s}{c} \Tilde{U}^{(c)}_n(f)\right\rvert = o_p(\tau^2_n).
\end{equation}
This refinement of condition (\ref{eq: outer-measure}) is appropriate because the proof of Theorem 3.2.5 in \citet{van2023weak} allows shifting the modulus downwards by a negligible quadratic term (compared to $\tau^2_n$), see the 3.4.3 Addendum in \citet{van2023weak}.

Condition (\ref{eq:leading term}) is given by Lemma \ref{lemma: convergence rate}. Condition (\ref{eq:remainder term}) is needed to demonstrate that the remainder terms are negligible compared to the rate $\tau^2_n$, which is guaranteed by the standard
uniform law of the large number for a degenerate U-process. Specifically, through the symmetrization and the convergence rate of the Rademacher chaos process of degree 2 (see Corollary 5.1.8 in \citet{de2012decoupling}), along with the metric entropy and bracketing number $\mathcal{N}_{[]}(\epsilon,\mathcal{F}, L_\infty(P))\lesssim (\frac{U}{\epsilon})^{\varsigma+1}$ from the proof of Lemma 6 in \cite{zhong2022deep}, one can show that
\begin{align*}
\sup_{f\in\mathcal{F}}\lvert\Tilde{U}^{(2)}_n(f)\rvert &= O_p\left(\frac{1}{n}\int_0^\mathcal{D}\log\mathcal{N}_{[]}(\epsilon,\mathcal{F},L_\infty(P))\mathrm{d}\epsilon\right) \\
& = O_p\left(\frac{1}{n}\int_0^\mathcal{D}\log\left(\frac{U}{\epsilon}\right)^{\varsigma+1}\mathrm{d}\epsilon\right) \\
& = O_p\left(\frac{\varsigma\log U}{n}\right)  =O_p(\Upsilon_n^2\log^{3} n) = o_p(\tau^2_n),
\end{align*}
and similarly we have $\sup_{f\in\mathcal{F}}\lvert\Tilde{U}^{(c)}_n(f)\rvert =  o_p(\tau^2_n )$ for $c=3,\dots,s$. Because $s$ is finite, we have condition (\ref{eq:remainder term}) verified.

Next, we verify the other three conditions to apply Theorem 3.2.5 in \cite{van2023weak}. For (\ref{ap-eq: convergence rate 2}), it is guaranteed by Lemma \ref{lemma: minimizer does not depend on s}. 
For (\ref{ap-eq: convergence rate 3}), $\phi_n(\delta)/\delta$ is a decreasing function of $\delta$. Furthermore, by (N1), it is clear that $\tau_n^{-2}\phi(\tau_n)\leq \sqrt{n}$.

Lastly, we verify the condition (\ref{ap-eq: convergence rate 4}). For $\check{f}$ defined in (\ref{ap-eq: f check}), by (\ref{eq: outer-measure}), along with law of the large number,
\begin{equation}
    \label{ap-eq: convergence rate 4.1}
\begin{split}
    & \lvert L^{(s)}_{n}(\check{f})-L^{(s)}_{0}(\check{f})-(L^{(s)}_{n}(f_0)-L^{(s)}_{0}(f_0))\lvert \\
    = & \lvert L^{(s)}_{n}(\check{f}-\mathbb{E}_X[\check{f}])-L^{(s)}_{0}(\check{f}-\mathbb{E}_X[\check{f}])-(L^{(s)}_{n}(f_0)-L^{(s)}_{0}(f_0))\lvert \\
    \lesssim & O_p(n^{-1/2}\phi(\tau_n))
\end{split}
\end{equation}

The rate in (\ref{ap-eq: convergence rate 4.1}), together with Lemma \ref{lemma: minimizer does not depend on s} and (\ref{eq:Schmidt's result}), indicates
\begin{align*}
\lvert L^{(s)}_{n}(\check{f})-L^{(s)}_{n}(f_0)\rvert & = \lvert L^{(s)}_{n}(\check{f})-L^{(s)}_{0}(\check{f})+ L^{(s)}_{0}(\check{f})-L^{(s)}_{0}(f_0)+ L^{(s)}_{0}(f_0)-L^{(s)}_{n}(f_0)\rvert\\
& \leq \lvert L^{(s)}_{n}(\check{f})-L^{(s)}_{0}(\check{f})-(L^{(s)}_{n}(f_0)-L^{(s)}_{0}(f_0))\lvert+\rvert L^{(s)}_{0}(\check{f})-L^{(s)}_{0}(f_0)\rvert\\
& = \lvert L^{(s)}_{n}(\check{f})-L^{(s)}_{0}(\check{f})-(L^{(s)}_{n}(f_0)-L^{(s)}_{0}(f_0))\lvert+\rvert L^{(s)}_{0}(\check{f}-\mathbb{E}_X[\check{f}])-L^{(s)}_{0}(f_0)\rvert\\
& \lesssim O_p(n^{-1/2}\phi(\tau_n))+\lVert \check{f}-\mathbb{E}_X[\check{f}] - f_0\rVert^2_{L^2} \\
& = O_p(\tau_n^2).
\end{align*}

Thus, by the definition of $\Tilde{f}^{(s)}_n$ in (\ref{eq:SGD NN minimizer}), we have
\begin{equation}
L_n^{(s)}(\Tilde{f}^{(s)}_n-\mathbb{E}_X[\Tilde{f}^{(s)}_n])=L_n^{(s)}(\Tilde{f}^{(s)}_n)\leq L_n^{(s)}(\check{f})\leq L_n^{(s)}(f_0)+O_p(\tau_n^2).
\end{equation}

Therefore, by (\ref{eq:leading term}), (\ref{eq:remainder term}), (\ref{ap-eq: convergence rate 2}), (\ref{ap-eq: convergence rate 3}), and (\ref{ap-eq: convergence rate 4}) we obtain that
$$d(\Tilde{f}^{(s)}_n-\mathbb{E}_X[\Tilde{f}^{(s)}_n],f_0) = O_p(\tau_n)$$
by applying Theorem 3.2.5 in \cite{van2023weak}.

Now, we have shown that
\begin{equation}
\label{eq: Pf shift}
\lVert \Tilde{f}^{(s)}_n-\mathbb{E}_X[\Tilde{f}^{(s)}_n]-f_0 \rVert_{L^2([0,1]^{p})}=O_p(\Upsilon_n\log^2 n),
\end{equation}
and what remains to be shown is
$$\lVert \Tilde{f}^{(s)}_n-\frac{1}{n}\sum_{i=1}^n\Tilde{f}^{(s)}_n(X_i)-f_0 \rVert_{L^2([0,1]^{p})}=O_p(\Upsilon_n\log^2 n).$$
Notice that
\begin{equation}
\label{eq: empirical average shift}
\begin{split}
&\lVert \Tilde{f}^{(s)}_n-\frac{1}{n}\sum_{i=1}^n\Tilde{f}^{(s)}_n(X_i)-f_0 \rVert_{L^2([0,1]^{p})}^2\\
& =\lVert \Tilde{f}^{(s)}_n-\mathbb{E}_X[\Tilde{f}^{(s)}_n]+\mathbb{E}_X[\Tilde{f}^{(s)}_n]-\frac{1}{n}\sum_{i=1}^n\Tilde{f}^{(s)}_n(X_i)-f_0 \rVert_{L^2([0,1]^{p})}^2\\
& =\int_{[0,1]^p} \left\{\Tilde{f}^{(s)}_n(x)-\mathbb{E}_X[\Tilde{f}^{(s)}_n]+\mathbb{E}_X[\Tilde{f}^{(s)}_n]-\frac{1}{n}\sum_{i=1}^n\Tilde{f}^{(s)}_n(X_i)-f_0(x)\right\}^2 dF(x)\\
& =\int_{[0,1]^p} \left\{\Tilde{f}^{(s)}_n(x)-\mathbb{E}_X[\Tilde{f}^{(s)}_n]-f_0(x)\right\}^2 dF(x)\\
&\quad +2\int_{[0,1]^p} \left\{\Tilde{f}^{(s)}_n(x)-\mathbb{E}_X[\Tilde{f}^{(s)}_n]-f_0(x)\right\}\left\{\mathbb{E}_X[\Tilde{f}^{(s)}_n]-\frac{1}{n}\sum_{i=1}^n\Tilde{f}^{(s)}_n(X_i)\right\} dF(x)\\
&\quad +\int_{[0,1]^p} \left\{\mathbb{E}_X[\Tilde{f}^{(s)}_n]-\frac{1}{n}\sum_{i=1}^n\Tilde{f}^{(s)}_n(X_i)\right\}^2 dF(x)\\
&= \lVert \Tilde{f}^{(s)}_n-\mathbb{E}_X[\Tilde{f}^{(s)}_n]-f_0 \rVert_{L^2([0,1]^{p})}^2+0+\left\lvert\mathbb{E}_X[\Tilde{f}^{(s)}_n]-\frac{1}{n}\sum_{i=1}^n\Tilde{f}^{(s)}_n(X_i)\right\rvert^2\\
&= \lVert \Tilde{f}^{(s)}_n-\mathbb{E}_X[\Tilde{f}^{(s)}_n]-f_0 \rVert_{L^2([0,1]^{p})}^2+\left\lvert(\mathbb{P}_n-\mathbb{P})\Tilde{f}^{(s)}_n\right\rvert^2.
\end{split}
\end{equation}

By Lemma \ref{lemma: convergence rate}, we obtain
$$\sup_{f\in \mathcal{F}(K,\varsigma,\mathbf{p},\mathcal{D})}\left\lvert\sqrt{n}(\mathbb{P}_n-\mathbb{P})f\right\rvert = O_p(\mathcal{D}\sqrt{\varsigma\log \frac{U}{\mathcal{D}}}+\frac{\varsigma \mathcal{D}}{\sqrt{n}}\log \frac{U}{\mathcal{D}}).$$

Therefore,
\begin{equation}
\label{eq: convergence rate of fn}
\left\lvert(\mathbb{P}_n-\mathbb{P})\Tilde{f}^{(s)}_n\right\rvert = O_p(\sqrt{\frac{\varsigma}{n}\log U})=O_p(\Upsilon_n\log^{3/2} n) 
\end{equation}

By (\ref{eq: Pf shift}), (\ref{eq: empirical average shift}), and (\ref{eq: convergence rate of fn}), we have
$$\lVert \Tilde{f}^{(s)}_n-\frac{1}{n}\sum_{i=1}^n\Tilde{f}^{(s)}_n(X_i)-f_0 \rVert_{L^2([0,1]^{p})}=O_p(\Upsilon_n\log^2 n).$$
\end{proof}

\subsection{Proof of Theorem 2}
\begin{theorem}
\label{thm: batch size and Hessian}
Under the Cox model with $f_0 = f_{\theta_0}$ parameterized by $\theta_0$ of finite dimension, with assumptions (A1)-(A3) and suppose $\nabla_\theta f_\theta$, $\nabla^2_\theta f_\theta$ exist and $f_\theta$,$\nabla_\theta f_\theta$, $\nabla^2_\theta f_\theta$ are element-wise bounded for all $X$ on a neighborhood of $\theta_0$ with $ \nabla^2_\theta\mathbb{E}[L_{Cox}^{(s)}(\theta)]|_{\theta=\theta_{0}}$ is positive definite, then for any integer $s\geq 2$ we have
\begin{equation}
    \nabla^2_\theta\mathbb{E}[L_{Cox}^{(s)}(\theta)]|_{\theta=\theta_{0}} = s\mathbb{V}[\nabla_\theta L_{Cox}^{(s)}(\theta)]|_{\theta=\theta_{0}}, \label{eq: Hessian and covariance}
\end{equation}
and
\begin{equation}
    \nabla^2_\theta\mathbb{E}[L_{Cox}^{(2s)}(\theta)]|_{\theta=\theta_{0}} \succeq \nabla^2_\theta\mathbb{E}[L_{Cox}^{(s)}(\theta)]|_{\theta=\theta_{0}}, \label{eq: batch size and Hessian}
\end{equation}
where $\nabla^2_\theta$ is second-order derivative operator with respect to $\theta$ and $A \succeq B$ denotes $A-B$ is positive semi-definite.
\end{theorem}

\subsubsection*{Proof of (\ref{eq: Hessian and covariance})}
\begin{proof}

The boundedness of the $X_i$ and the functions from the assumptions are sufficient to justify the interchange of
differentiation and expectation through the dominated convergence theorem. We first rewrite ${\nabla_\theta}\mathbb{E}[L_{Cox}^{(s)}(\theta)]$:
\begin{align*}
{\nabla_\theta}\mathbb{E}[L_{Cox}^{(s)}(\theta)]
= & {\nabla_\theta}\mathbb{E}\left[\int_0^\tau \frac{1}{s}\sum_{i=1}^s \left\{ -f_{\theta}(X_i)+\log \left(\sum_{j=1}^s Y_j(u)\exp\{f_{\theta}(X_j)\}\right) \right\} \mathrm{d}N_i(u)\right ] \\
= &\int_0^\tau \mathbb{E}\left[\frac{1}{s}\sum_{i=1}^s \left\{ -{\nabla_\theta}f_{\theta}(X_i)+\frac{S^{f(1)}_s (u,\theta)}{S^{f(0)}_s (u,\theta)} \right\} \mathrm{d}N_i(u)\right]\\
= &\int_0^\tau \mathbb{E}_{\mathcal{F}^{(s)}_u}\left[\frac{1}{s}\sum_{i=1}^s \left\{ -{\nabla_\theta}f_{\theta}(X_i)+\frac{S^{f(1)}_s (u,\theta)}{S^{f(0)}_s (u,\theta)} \right\} \mathbb{E}[\mathrm{d}N_i(u)|\mathcal{F}^{(s)}_u]\right],
\end{align*}
where
\begin{align*}
S^{f(0)}_s (u,\theta) = &\sum_{j=1}^s Y_j(u)\exp\{f_{\theta}(X_j)\} \\
S^{f(1)}_s (u,\theta) = & \sum_{j=1}^s Y_j(u)\exp\{f_{\theta}(X_j)\}{\nabla_\theta} [ f_{\theta}(X_j)]
\end{align*}
and are predictable with respect to $\mathcal{F}^{(s)}_u$.
Under Cox model $\lambda(t|X) = \lambda_0(u)\exp\{f_{\theta_0}(X)\}$, and we have
\begin{equation}
\mathbb{E}[\mathrm{d}N_i(u)|\mathcal{F}^{(s)}_u] =Y_i(u)\lambda(t|X_i) \mathrm{d}u=  Y_i(u)\exp(f_{\theta_0}(X_i))\lambda_0(u)\mathrm{d}u. \label{ap-eq:compensator}
\end{equation}
from the martingale theory.
 Therefore,
\begin{equation}
{\nabla_\theta}\mathbb{E}[L_{Cox}^{(s)}(\theta)]
= \int_0^\tau \mathbb{E}_{\mathcal{F}^{(s)}_u}\left[-\frac{1}{s}S^{\nabla f}_s (u,\theta;\theta_0)+\frac{1}{s}\frac{S^{f(1)}_s (u,\theta)}{S^{f(0)}_s (u,\theta)}S^{f(0)}_s (u,\theta_0) \right]\lambda_0(u)\mathrm{d}u, \label{ap-eq: E of score}
\end{equation}
where
$$S^{\nabla f}_s (u,\theta;\theta_0) = \sum_{j=1}^s Y_j(u)\exp(f_{\theta_0}(X_j)){\nabla_\theta} [ f_{\theta}(X_j)]$$
and we omit the subscript ${\mathcal{F}^{(s)}_u}$ for simplicity.

Notice that $S^{\nabla f}_s (u,\theta;\theta_0)|_{\theta = \theta_0} = S^{f(1)}_s (u,\theta_0)$, we immediately have $\mathbb{E}[{\nabla_\theta}L_{Cox}^{(s)}(\theta)]|_{\theta = \theta_0} = 0$ for any fixed $s\geq 2$. This demonstrates that $\theta_0$ is the root of $\mathbb{E}[{\nabla_\theta}L_{Cox}^{(s)}(\theta)]|_{\theta = \theta_0}$ regardless the choice of $s$, which was first presented by \citep{tarkhan2024online}.

To prove (\ref{eq: Hessian and covariance}), we first derive the expression for the left-hand side of (\ref{eq: Hessian and covariance}):
\begin{align*}
& \nabla^2_\theta\mathbb{E}[L_{Cox}^{(s)}(\theta)] \\
=&  \int_0^\tau \mathbb{E}\left[-\nabla_\theta S^{\nabla f}_s (u,\theta;\theta_0)+S^{f(0)}_s (u,\theta_0)\nabla_\theta\left\{ \frac{S^{f(1)}_s (u,\theta)}{S^{f(0)}_s (u,\theta)}\right\} \right]\frac{\lambda_0(u)}{s}\mathrm{d}u \\
=&  \int_0^\tau \mathbb{E}\left[-S^{\nabla^2 f}_s (u,\theta;\theta_0)+S^{f(0)}_s (u,\theta_0)\left\{ \frac{S^{f(2)}_s (u,\theta)}{S^{f(0)}_s (u,\theta)}-\frac{S^{f(1)}_s (u,\theta)^{\otimes 2}}{S^{f(0)}_s (u,\theta)^{\otimes 2}}  \right\} \right]\frac{\lambda_0(u)}{s}\mathrm{d}u,
\end{align*}
where
\begin{align*}
 S^{\nabla^2 f}_s (u,\theta;\theta_0) &= \nabla_\theta S^{\nabla f}_s (u,\theta;\theta_0) = \sum_{j=1}^s Y_j(u)\exp(f_{\theta_0}(X_j))\nabla^2_\theta [ f_{\theta}(X_j)]\\
 S^{f(2)}_s (u,\theta)&=\nabla_\theta S^{f(1)}_s (u,\theta) \\
 &= \sum_{j=1}^s Y_j(u)\exp\{f_{\theta}(X_j)\}{\nabla_\theta} [ f_{\theta}(X_j)] \otimes {\nabla_\theta} [ f_{\theta}(X_j)] +\sum_{j=1}^s Y_j(u)\exp\{f_{\theta}(X_j)\}\nabla^2_\theta[ f_{\theta}(X_j)] \\
&= \sum_{j=1}^s Y_j(u)\exp\{f_{\theta}(X_j)\}{\nabla_\theta} [ f_{\theta}(X_j)] \otimes {\nabla_\theta} [ f_{\theta}(X_j)] +S^{\nabla^2 f}_s (u,\theta;\theta_0)
\end{align*}
Therefore,
\begin{equation}\label{eq:expression of Hessian}
\nabla^2_\theta\mathbb{E}[L_{Cox}^{(s)}(\theta)]|_{\theta=\theta_0} = \int_0^\tau \mathbb{E}\left[-S^{\nabla^2 f}_s (u,\theta;\theta_0)|_{\theta = \theta_0}+S^{f(2)}_s (u,\theta_0)-\frac{S^{f(1)}_s (u,\theta_0)^{\otimes 2}}{S^{f(0)}_s (u,\theta_0)} \right]\frac{\lambda_0(u)}{s}\mathrm{d}u
\end{equation}
Then we derive the expression for the right-hand side of Eq. (\ref{eq: Hessian and covariance})
\begin{equation}
\label{eq:var of martingale}
\begin{split}
& \mathbb{V}[\nabla_\theta L_{Cox}^{(s)}(\theta)]|_{\theta = \theta_0} \\
= & \left.\mathbb{V}\left[\int_0^\tau \frac{1}{s}\sum_{i=1}^s \left( -\nabla_\theta f_\theta(X_i)+ \frac{S^{f(1)}_s (u,\theta)}{S^{f(0)}_s (u,\theta)} \right) \mathrm{d}N_i(u)\right]\right|_{\theta = \theta_0} \\
=& \left.\mathbb{V}\left[\int_0^\tau \frac{1}{s}\sum_{i=1}^s \left( -\nabla_\theta f_\theta(X_i)+ \frac{S^{f(1)}_s (u,\theta)}{S^{f(0)}_s (u,\theta)} \right) \mathrm{d}M_i(u)\right]\right|_{\theta = \theta_0} \\ 
(1)= & \left.\frac{1}{s^2} \int_0^\tau \sum_{i=1}^s \mathbb{E}\left[\left( -\nabla_\theta f_\theta(X_i)+ \frac{S^{f(1)}_s (u,\theta)}{S^{f(0)}_s (u,\theta)} \right)^{\otimes 2} \mathrm{d}\langle M_i(u), M_i(u)\rangle\right]\right|_{\theta = \theta_0} \\
= & \frac{1}{s^2} \int_0^\tau \sum_{i=1}^s \mathbb{E}\left[\left\{ \left(\nabla_\theta f_\theta(X_i)\right)^{\otimes 2} - \nabla_\theta f_\theta(X_i)\otimes \frac{S^{f(1)}_s (u,\theta)}{S^{f(0)}_s (u,\theta)} \right.\right.\\
& \quad\quad\quad\quad\quad\quad\quad-\left.\left.\left.\frac{S^{f(1)}_s (u,\theta)}{S^{f(0)}_s (u,\theta)}\otimes \nabla_\theta f_\theta(X_i)  + \left(\frac{S^{f(1)}_s (u,\theta)}{S^{f(0)}_s (u,\theta)}\right)^{\otimes 2} \right\} \mathrm{d}\langle M_i(u), M_i(u)\rangle\right]\right|_{\theta = \theta_0},
\end{split}
\end{equation}
where (1) follows the Theorem 2.4.2 and Theorem 2.5.2 in \citet{fleming2013counting}. Additionally, the Theorem 2.5.2 in \citet{fleming2013counting} and (\ref{ap-eq:compensator}) implies that
\begin{align*}
& \mathbb{V}[\nabla_\theta L_{Cox}^{(s)}(\theta)]|_{\theta = \theta_0}\\
= & \frac{1}{s^2} \int_0^\tau  \mathbb{E}\left.\left[ \sum_{i=1}^sY_i(u)\exp\{f_{\theta_0}(X_j)\}\left[\nabla_\theta f_{\theta}(X_i)\right]^{\otimes 2}\right|_{\theta = \theta_0} - S^{f(1)}_s (u,\theta_0)\otimes \frac{S^{f(1)}_s (u,\theta_0)}{S^{f(0)}_s (u,\theta_0)} \right.\\
& \quad\quad\quad\quad\quad\quad\quad\quad\quad\quad\quad\quad\quad\quad-\left.\frac{S^{f(1)}_s (u,\theta_0)}{S^{f(0)}_s (u,\theta_0)}\otimes S^{f(1)}_s (u,\theta_0)  + \frac{S^{f(1)}_s (u,\theta_0)^{\otimes 2}}{S^{f(0)}_s (u,\theta_0)} \right]\lambda_0(u)\mathrm{d}u\\
= & \frac{1}{s^2} \int_0^\tau \mathbb{E}\left[-S^{\nabla^2 f}_s (u,\theta;\theta_0)|_{\theta=\theta_0} +S^{f(2)}_s (u,\theta_0) -  \frac{(S^{f(1)}_s (u,\theta_0))^{\otimes 2}}{S^{f(0)}_s (u,\theta_0)}\right]\lambda_0(u)\mathrm{d}u.
\end{align*}
Compare this expression with (\ref{eq:expression of Hessian}) and we have
\begin{equation*}
\nabla^2_\theta \mathbb{E}[L_{Cox}^{(s)}(\theta)]|_{\theta=\theta_0} = s\mathbb{V}[\nabla_\theta L_{Cox}^{(s)}(\theta)]|_{\theta=\theta_0}.
\end{equation*}
\end{proof}

\subsubsection*{Proof of (\ref{eq: batch size and Hessian})}
\begin{proof}
From the proof of (\ref{eq: Hessian and covariance}), we have
\begin{equation*}
\begin{split}
& \nabla^2_\theta\mathbb{E}[L_{Cox}^{(s)}(\theta)]|_{\theta=\theta_0} \\
= &\int_0^\tau \mathbb{E}\left[-S^{\nabla^2 f}_s (u,\theta;\theta_0)|_{\theta = \theta_0}+S^{f(2)}_s (u,\theta_0)-\frac{S^{f(1)}_s (u,\theta_0)^{\otimes 2}}{S^{f(0)}_s (u,\theta_0)} \right]\frac{\lambda_0(u)}{s}\mathrm{d}u \\
=& \int_0^\tau \mathbb{E}\left [Y(u)\exp(X^T\theta_0)\left\{ (\nabla_\theta f_{\theta}(X_i))^{\otimes 2}|_{\theta = \theta_0}\right \}\right]\lambda_0(u)\mathrm{d}u-\int_0^\tau \mathbb{E}\left[\frac{S^{f(1)}_s (u,\theta_0)^{\otimes 2}}{S^{f(0)}_s (u,\theta_0)} \right]\frac{\lambda_0(u)}{s}\mathrm{d}u\\
=& \int_0^\tau \mathbb{E}_{\mathcal{F}^{(s)}_u}\left [Y(u)\exp(X^T\theta_0)\left\{ (\nabla_\theta f_{\theta}(X_i))^{\otimes 2}|_{\theta = \theta_0}\right \}\right]\lambda_0(u)\mathrm{d}u-\int_0^\tau \mathbb{E}_{\mathcal{F}^{(s)}_u}\left[\frac{S^{f(1)}_s (u,\theta_0)^{\otimes 2}}{S^{f(0)}_s (u,\theta_0)} \right]\frac{\lambda_0(u)}{s}\mathrm{d}u.
\end{split}
\end{equation*}

Notice that the first item does not depend on $s$.

Recall that the filtration $\mathcal{F}^{(s)}_u$ is the information up to time $u$ for $s$ i.i.d. samples, we consider a larger filtration $\mathcal{F}^{(2s)}_u$ with additional $s$ i.i.d. samples, that is $\mathcal{F}^{(s)}_u \subset \mathcal{F}^{(2s)}_u$. Let $j=1,2,\dots,s$ denote the original $s$ subjects and $j=s+1,s+2,\dots,2s$ denote the additional $s$ subjects. For $j=1,2,\dots,s$, we have $\mathbb{E}[\mathrm{d}N_j(u)|\mathcal{F}^{(2s)}_u] = \mathbb{E}[\mathbb{E}(\mathrm{d}N_j(u)|\mathcal{F}^{(s)}_u)|\mathcal{F}^{(2s)}_u]= \mathbb{E}[\mathrm{d}N_j(u)|\mathcal{F}^{(s)}_u] $ and $\mathbb{E}_{\mathcal{F}^{(s)}_u}[f(X_j,Y_j(u))] = \mathbb{E}_{\mathcal{F}^{(2s)}_u}[f(X_j,Y_j(u))]$ by property of conditional expectation. Therefore, the derivation above still holds under the filtration $\mathcal{F}^{(2s)}_u$. Hence, in order to prove $\mathbb{E}[{\nabla^2_\theta}L_{Cox}^{(2s)}(\theta)]|_{\theta=\theta_0} \succeq \mathbb{E}[{\nabla^2_\theta}L_{Cox}^{(s)}(\theta)]|_{\theta=\theta_0}$, it suffices to show 
\begin{equation}
\begin{split}
& \mathbb{E}[{\nabla^2_\theta}L_{Cox}^{(s)}(\theta)]|_{\theta=\theta_0} - \mathbb{E}[{\nabla^2_\theta}L_{Cox}^{(2s)}(\theta)]|_{\theta=\theta_0}\\
= & \int_0^\tau \mathbb{E}_{\mathcal{F}^{(2s)}_u}\left[\frac{S^{f(1)}_{2s} (u,\theta_0)^{\otimes 2}}{S^{f(0)}_{2s} (u,\theta_0)} \right]\frac{\lambda_0(u)}{2s}\mathrm{d}u-\int_0^\tau \mathbb{E}_{\mathcal{F}^{(s)}_u}\left[\frac{S^{f(1)}_s (u,\theta_0)^{\otimes 2}}{S^{f(0)}_s (u,\theta_0)} \right]\frac{\lambda_0(u)}{s}\mathrm{d}u \\
= & \int_0^\tau \mathbb{E}_{\mathcal{F}^{(2s)}_u}\left[\frac{S^{f(1)}_{2s} (u,\theta_0)^{\otimes 2}}{S^{f(0)}_{2s} (u,\theta_0)} \right]\frac{\lambda_0(u)}{2s}\mathrm{d}u-\int_0^\tau \mathbb{E}_{\mathcal{F}^{(2s)}_u}\left[\frac{S^{f(1)}_s (u,\theta_0)^{\otimes 2}}{S^{f(0)}_s (u,\theta_0)} \right]\frac{\lambda_0(u)}{s}\mathrm{d}u  \\
\preceq & \ 0 \label{eq:target}
\end{split}
\end{equation}
with the filtration $\mathcal{F}^{(2s)}_u$.

By Lemma \ref{lemma: change variable}, we can replace $\frac{S^{f(1)}_{2s} (u,\theta_0)}{{2s}}$ by $\frac{S^{f(1)}_{s} (u,\theta_0)}{{s}}$. That is, 
\begin{align*}
& \int_0^\tau \mathbb{E}_{\mathcal{F}^{(2s)}_u}\left[ \frac{S^{f(1)}_{2s} (u,\theta_0)}{S^{f(0)}_{2s} (u,\theta_0)}   \otimes \frac{S^{f(1)}_{2s} (u,\theta_0)}{{2s}}\right]\lambda_0(u)\mathrm{d}u \\
& = \int_0^\tau \mathbb{E}_{\mathcal{F}^{(2s)}_u}\left[ \frac{S^{f(1)}_{2s} (u,\theta_0)}{S^{f(0)}_{2s} (u,\theta_0)} \otimes \frac{\sum_{j=1}^{2s} Y_j(u)\exp(f_{\theta_0}(X_j))\nabla_\theta f_{\theta_0}(X_j)|_{\theta=\theta_0}}{{2s}}\right]\lambda_0(u)\mathrm{d}u \\
& = \int_0^\tau \mathbb{E}_{\mathcal{F}^{(2s)}_u}\left[ \frac{S^{f(1)}_{2s} (u,\theta_0)}{S^{f(0)}_{2s} (u,\theta_0)}  \otimes Y_1(u)\exp(f_{\theta_0}(X_1))\nabla_\theta f_{\theta_0}(X_j)|_{\theta=\theta_0}\right]\lambda_0(u)\mathrm{d}u \quad \text{ by Lemma \ref{lemma: change variable}}\\
& = \int_0^\tau \mathbb{E}_{\mathcal{F}^{(2s)}_u}\left[ \frac{S^{f(1)}_{2s} (u,\theta_0)}{S^{f(0)}_{2s} (u,\theta_0)} \otimes \frac{\sum_{j=1}^{s} Y_j(u)\exp(f_{\theta_0}(X_j))\nabla_\theta f_{\theta_0}(X_j)|_{\theta=\theta_0}}{{s}}\right]\lambda_0(u)\mathrm{d}u \quad \text{ by Lemma \ref{lemma: change variable}}\\
& = \int_0^\tau \mathbb{E}_{\mathcal{F}^{(2s)}_u}\left[ \frac{S^{f(1)}_{2s} (u,\theta_0)}{S^{f(0)}_{2s} (u,\theta_0)}  \otimes \frac{S^{f(1)}_{s} (u,\theta_0)}{{s}}\right]\lambda_0(u)\mathrm{d}u.
\end{align*}

Hence, (\ref{eq:target}) is equivalent to 
\begin{align*}
& \int_0^\tau \mathbb{E}_{\mathcal{F}^{(2s)}_u}\left[ \frac{S^{f(1)}_{2s} (u,\theta_0)}{S^{f(0)}_{2s} (u,\theta_0)}   \otimes \frac{S^{f(1)}_{s} (u,\theta_0)}{{s}}\right]\lambda_0(u)\mathrm{d}u -\int_0^\tau \mathbb{E}_{\mathcal{F}^{(2s)}_u}\left[\frac{S^{f(1)}_{s} (u,\theta_0)}{S^{f(0)}_{s} (u,\theta_0)}  \otimes \frac{S^{f(1)}_{s} (u,\theta_0)}{{s}}\right]\lambda_0(u)\mathrm{d}u \\
& = \int_0^\tau \mathbb{E}_{\mathcal{F}^{(2s)}_u}\left[\left( \frac{S^{f(1)}_{2s} (u,\theta_0)}{S^{f(0)}_{2s} (u,\theta_0)}  - \frac{S^{f(1)}_{s} (u,\theta_0)}{S^{f(0)}_{s} (u,\theta_0)}  \right) \otimes \frac{S^{f(1)}_{s} (u,\theta_0)}{{s}}\right]\lambda_0(u)\mathrm{d}u
\end{align*}

And we are going to show
\begin{equation}
\label{eq: H2s-Hs}
\int_0^\tau \mathbb{E}_{\mathcal{F}^{(2s)}_u}\left[\left( \frac{S^{f(1)}_{2s} (u,\theta_0)}{S^{f(0)}_{2s} (u,\theta_0)}  - \frac{S^{f(1)}_{s} (u,\theta_0)}{S^{f(0)}_{s} (u,\theta_0)}  \right) \otimes \frac{S^{f(1)}_{s} (u,\theta_0)}{{s}}\right]\lambda_0(u)\mathrm{d}u\preceq 0.
\end{equation}
We use $S_\Delta$ to denote the summation over the additional $s$ subjects ($j=s+1,s+2,\dots,2s$) only. Therefore we can decompose $S^{f(k)}_{2s} (u,\theta_0)$ by $S^{f(k)}_{2s} (u,\theta_0) = S^{f(k)}_{\Delta} (u,\theta_0)+S^{f(k)}_{s} (u,\theta_0)$.
\begin{align*}
& \int_0^\tau \mathbb{E}_{\mathcal{F}^{(2s)}_u}\left[\left( \frac{S^{f(1)}_{2s} (u,\theta_0)}{S^{f(0)}_{2s} (u,\theta_0)}  - \frac{S^{f(1)}_{s} (u,\theta_0)}{S^{f(0)}_{s} (u,\theta_0)}  \right) \otimes S^{f(1)}_{s} (u,\theta_0)\right]\frac{\lambda_0(u)}{{s}}\mathrm{d}u \\
= & \int_0^\tau \mathbb{E}_{\mathcal{F}^{(2s)}_u}\left[\left( \frac{S^{f(1)}_{\Delta} (u,\theta_0)+S^{f(1)}_{s} (u,\theta_0)}{S^{f(0)}_{\Delta} (u,\theta_0)+S^{f(0)}_{s} (u,\theta_0)}  - \frac{S^{f(1)}_{s} (u,\theta_0)}{S^{f(0)}_{s} (u,\theta_0)}  \right) \otimes S^{f(1)}_{s} (u,\theta_0)\right]\frac{\lambda_0(u)}{{s}}\mathrm{d}u \\
= & \int_0^\tau \mathbb{E}_{\mathcal{F}^{(2s)}_u}\left[\left( \frac{S^{f(1)}_{\Delta} (u,\theta_0)S^{f(0)}_{s} (u,\theta_0)-S^{f(0)}_{\Delta} (u,\theta_0)S^{f(1)}_{s} (u,\theta_0)}{[S^{f(0)}_{\Delta} (u,\theta_0)+S^{f(0)}_{s} (u,\theta_0)]S^{f(0)}_{s} (u,\theta_0)}\right) \otimes S^{f(1)}_{s} (u,\theta_0)\right] \frac{\lambda_0(u)}{{s}}\mathrm{d}u \\
= & \int_0^\tau \mathbb{E}_{\mathcal{F}^{(2s)}_u}\left[\left\{ \left(\frac{S^{f(1)}_{\Delta} (u,\theta_0)}{S^{f(0)}_{\Delta} (u,\theta_0)} - \frac{S^{f(1)}_{s} (u,\theta_0)}{S^{f(0)}_{s} (u,\theta_0)}\right) \times \frac{S^{f(0)}_{\Delta} (u,\theta_0)}{S^{f(0)}_{\Delta} (u,\theta_0)+S^{f(0)}_{s} (u,\theta_0)} \right\} \otimes S^{f(1)}_{s} (u,\theta_0)\right]\frac{\lambda_0(u)}{{s}}\mathrm{d}u  \\
= & \int_0^\tau \mathbb{E}_{\mathcal{F}^{(2s)}_u}\left[\left\{ \left(\frac{S^{f(1)}_{\Delta} (u,\theta_0)}{S^{f(0)}_{\Delta} (u,\theta_0)} - \frac{S^{f(1)}_{s} (u,\theta_0)}{S^{f(0)}_{s} (u,\theta_0)}\right) \times \frac{S^{f(0)}_{\Delta} (u,\theta_0)S^{f(0)}_{s} (u,\theta_0)}{S^{f(0)}_{\Delta} (u,\theta_0)+S^{f(0)}_{s} (u,\theta_0)} \right\} \otimes \frac{S^{f(1)}_{s} (u,\theta_0)}{{S^{f(0)}_{s} (u,\theta_0)}}\right]\frac{\lambda_0(u)}{s}\mathrm{d}u  \\
 = & \int_0^\tau \mathbb{E}_{\mathcal{F}^{(2s)}_u}[\left\{ h_{\theta_0}(Z_1(u))-h_{\theta_0}(Z_2(u)) \right\}g_{\theta_0}(Z_1(u),Z_2(u)) \otimes h_{\theta_0}(Z_2(u)) ]\frac{\lambda_0(u)}{s}\mathrm{d}u \\
  = & \int_0^\tau \mathbb{E}_{\mathcal{F}^{(2s)}_u}[\left\{ h_{\theta_0}(Z_1(u))-h_{\theta_0}(Z_2(u)) \right\} \otimes h_{\theta_0}(Z_2(u)) g_{\theta_0}(Z_1(u),Z_2(u)) ]\frac{\lambda_0(u)}{s}\mathrm{d}u,
\end{align*}
where $Z_1(u) = \{(Y_{s+1}(u),X_{s+1}),\dots,(Y_{2s}(u),X_{2s})\} , Z_2(u) = \{(Y_{1}(u),X_{1}),\dots,(Y_{s}(u),X_{s})\}$, 
$$ h_{\theta_0}(Z_1(u)) = \frac{S^{f(1)}_{\Delta} (u,\theta_0)}{S^{f(0)}_{\Delta} (u,\theta_0)},\ h_{\theta_0}(Z_2(u)) = \frac{S^{f(1)}_{s} (u,\theta_0)}{S^{f(0)}_{s} (u,\theta_0)}, $$
$$ g_{\theta_0}(Z_1(u),Z_2(u)) = \frac{S^{f(0)}_{\Delta} (u,\theta_0)S^{f(0)}_{s} (u,\theta_0)}{S^{f(0)}_{\Delta} (u,\theta_0)+S^{f(0)}_{s} (u,\theta_0)}. $$
Since $g$ is a positive symmetric real-valued function and $Z_1, Z_2$ are i.i.d. samples, we have
$\mathbb{E}_{\mathcal{F}^{(2s)}_u}[\left( h_{\theta_0}(Z_1)-h_{\theta_0}(Z_2) \right) \otimes h_{\theta_0}(Z_2) g_{\theta_0}(Z_1,Z_2) ]\preceq 0$ by Lemma \ref{lemma: final step}.

Notice that this inequality holds for any $u$. Therefore, 
$$\int_0^\tau \mathbb{E}_{\mathcal{F}^{(2s)}_u}[\left( h_{\theta_0}(Z_2)-h_{\theta_0}(Z_1) \right) \otimes h_{\theta_0}(Z_1)g_{\theta_0}(Z_1,Z_2) ]\frac{\lambda_0(u)\mathrm{d}u}{s}\preceq 0$$
and this implies $\mathbb{E}[{\nabla^2_\theta}L_{Cox}^{(2s)}(\theta)]|_{\theta=\theta_0} \succeq \mathbb{E}[{\nabla^2_\theta}L_{Cox}^{(s)}(\theta)]|_{\theta=\theta_0}$.
\end{proof}

\subsection{Proof of Theorem 3}
\begin{theorem}\label{thm:asymp-dist-SBFB}
Under the Cox model, with assumptions (A1)-(A3), (R1), (R2) and for any integer $s\geq 2$, we have
\begin{align}
\sqrt{n}(\tilde{\theta}_{n}^{SB(s)}-\theta_0) & \to^d N(0,s^2H^{-1}_s\Sigma_{(s|1)}(H_s^{-1})^T), \label{eq:asymp-dist-SBGD} \\
\sqrt{n}(\tilde{\theta}_{n}^{FB(s)}-\theta_0) & \to^d N(0,sH^{-1}_s\Sigma_s(H_s^{-1})^T),
\label{eq:asymp-dist-FBGD}
\end{align}
when $n\to \infty$, where $H_s = \mathbb{E}[\nabla^2_\theta L_{Cox}^{(s)}(\theta)]|_{\theta=\theta_0}$, $\Sigma_s = \mathbb{V}[\nabla_\theta L_{Cox}^{(s)}(\theta)]|_{\theta=\theta_0}$, and
$$ \left . \Sigma_{(s|1)} = \mathbb{V}\left \{ \nabla_\theta L_{Cox}^{(s)}(D_{i_1},D_{i_2},\dots,D_{i_s}|\theta),\nabla_\theta L_{Cox}^{(s)}(D_{i_1},\Tilde D_{{i}_2},\dots,\Tilde D_{{i}_s}|\theta)\right \}\right \rvert_{\theta=\theta_0}, $$
which is the covariance of $\nabla_\theta L_{Cox}^{(s)}$ on two mini-batches $D(s)$ sharing the same sample $D_{i_1}$ but different rest $s-1$ samples (denoted by $\Tilde D$).
\end{theorem}
\begin{proof}
We first establish the consistency of the mb-MPLE and then derive its asymptotic distribution.
\subsubsection{Consistency}
Under Cox-regression, by (\ref{ap-eq: E of score}) in the proof of Theorem \ref{thm: batch size and Hessian}, we have 
$\nabla_\theta \mathbb{E}[L_{Cox}^{(s)}(\theta)]|_{\theta=\theta_0} = 0$. Moreover, Lemma \ref{lemma: strongly convex} demonstrates that the minimizer of $\mathbb{E}[L_{Cox}^{(s)}(\theta)]$ is unique. Hence, $\theta_0$ is the unique minimizer of $\mathbb{E}[L_{Cox}^{(s)}(\theta)]$.

Notice that
\begin{align*}
\nabla^2_\theta L^{(s)}_{Cox}(\theta) & = \frac{1}{s}\sum_{i=1}^s \Delta_i \left\{\frac{S^{(2)}_s (T_i,\theta)}{S^{(0)}_s (T_i,\theta)}-\frac{S^{(1)}_s (T_i,\theta)^{\otimes 2}}{S^{(0)}_s (T_i,\theta)^{\otimes 2}} \right\}\\
& = \frac{1}{s}\sum_{i=1}^s \Delta_i \frac{S^{(2)}_s (T_i,\theta)S^{(0)}_s (T_i,\theta)-S^{(1)}_s (T_i,\theta)^{\otimes 2}}{S^{(0)}_s (T_i,\theta)^{\otimes 2}} \succeq 0
\end{align*}
by Cauchy–Schwarz inequality. Therefore, both $\frac{1}{\binom{n}{s}}\displaystyle\sum_{D(s)\subset D(n)} L_{Cox}^{(s)}(\theta)$ and $\frac{1}{m}\displaystyle\sum_{D(s)\in D(n|s)} L_{Cox}^{(s)}(\theta)$ are random convex functions of $\theta$. By the law of large numbers, for any $\theta$,
$$\lim_{n\to\infty}\frac{1}{\binom{n}{s}}\displaystyle\sum_{D(s)\subset D(n)} L_{Cox}^{(s)}(\theta) \to^p \mathbb{E}[L_{Cox}^{(s)}(\theta)],$$
and
$$\lim_{n\to\infty}\frac{1}{m}\displaystyle\sum_{D(s)\in D(n|s)} L_{Cox}^{(s)}(\theta) \to^p \mathbb{E}[L_{Cox}^{(s)}(\theta)].$$
Hence, following the arguments in Appendix II of \citet{andersen1982cox}, their minimizers
\begin{equation}
\tilde{\theta}^{SB(s)}_{n} := \arg\min_{\theta}\frac{1}{\binom{n}{s}}\sum_{D(s)\subset D(n)} L_{Cox}^{(s)}(\theta),\label{eq:SBGD estimator}
\end{equation}
and
\begin{equation}
\tilde{\theta}^{FB(s)}_{n} := \arg\min_{\theta} \frac{1}{m}\sum_{D(s)\in D(n|s)} L_{Cox}^{(s)}(\theta)
\label{eq:FBGD estimator}
\end{equation} 
converge to $\theta_0$ in probability.

\subsubsection{Asymptotic normality}
We first derive the asymptotic normality of $\tilde{\theta}^{SB(s)}_{n}$. 
Let $\mathcal{N}_{\theta_0}$ denote a neighborhood of $\theta_0$. Recall that
$${\nabla_\theta}L_{Cox}^{(s)}(\theta) = \frac{1}{s}\sum_{i:D_i\in D(s)} \Delta_i\left\{ -X_i+\frac{\sum_{j:D_j\in D(s)} Y_j(T_i)\exp(X_j^T \theta)X_j}{\sum_{j:D_j\in D(s)} Y_j(T_i)\exp(X_j^T \theta)} \right\}.$$
Then for any $k,l,m\in \{1,\dots,p\}$,
$$\frac{\partial^2}{\partial\theta_l\partial\theta_m}\left\{{\nabla_\theta}L_{Cox}^{(s)}(\theta)\right\}_k = \frac{1}{s}\sum_{i:D_i\in D(s)} \Delta_i \frac{\partial^2}{\partial\theta_l\partial\theta_m}\left\{\frac{\sum_{j:D_j\in D(s)} Y_j(T_i)\exp(X_j^T \theta)X_{jk}}{\sum_{j:D_j\in D(s)} Y_j(T_i)\exp(X_j^T \theta)} \right\},$$
where $\{\cdot\}_k$ denotes the $k$th element of a vector. Notice that $\frac{\partial^2}{\partial\theta_l\partial\theta_m}\left\{{\nabla_\theta}L_{Cox}^{(s)}(\theta)\right\}_k$ is uniformly upper-bounded by a constant on $\mathcal{N}_{\theta_0}$ because $Y_i(T_i) = 1$ and $\lVert X\rVert$ is bounded (by A3). Therefore, by Taylor expansion of $\sqrt{n}\frac{1}{\binom{n}{s}}\sum_{D(s)\subset D(n)} \nabla_\theta L_{Cox}^{(s)}(\tilde{\theta}^{SB(s)}_{n})$ at $\theta_0$, we have
\begin{equation}\label{ap-eq: asymptotic normal}
\begin{split}
0 &= \sqrt{n}\frac{1}{\binom{n}{s}}\sum_{D(s)\subset D(n)} \nabla_\theta L_{Cox}^{(s)}(\tilde{\theta}^{SB(s)}_{n}) \\
& = \sqrt{n}\frac{1}{\binom{n}{s}}\sum_{D(s)\subset D(n)} \nabla_\theta L_{Cox}^{(s)}(\theta_0) +\left\{\frac{1}{\binom{n}{s}}\sum_{D(s)\subset D(n)} \nabla^2_\theta L_{Cox}^{(s)}(\theta_0)+o_p(1)\right\} \sqrt{n}(\tilde{\theta}^{SB(s)}_{n}-\theta_0),
\end{split}
\end{equation}
where the term $o_p(1)$ is due to the boundedness of $\frac{\partial^2}{\partial\theta_l\partial\theta_m}\left\{{\nabla_\theta}L_{Cox}^{(s)}(\theta)\right\}_k$ and the consistency of $\tilde{\theta}^{SB(s)}_{n}$.

For the first item in (\ref{ap-eq: asymptotic normal}), by Hoeffding decomposition \citep{hoeffding1992class}, we have
\begin{equation}\label{ap-eq:asymptotic normal clt}
\sqrt{n}\frac{1}{\binom{n}{s}}\sum_{D(s)\subset D(n)} \nabla_\theta L_{Cox}^{(s)}(\theta_0) 
 = \sqrt{n}\frac{s}{n}\sum_{i=1}^n r(D_i;\theta_0)+o_p(1),
\end{equation}
where $r(D_1;\theta_0) = \mathbb{E}\left . \left\{\nabla_\theta L_{Cox}^{(s)}(D_1,D_{2},\dots,D_{s};\theta_0)\right \rvert D_1 \right\}$. Note that $\mathbb{E}[r(D_i;\theta_0)]=\mathbb{E}[\nabla_\theta L_{Cox}^{(s)}(\theta_0)]=0$. By the central limit theorem we have
 $$\sqrt{n}\frac{s}{n}\sum_{i=1}^n r(D_i;\theta_0)\to^d N(0,s^2\mathbb{V}\{r(D_i;\theta_0)\}) $$
 and $\mathbb{V}\{r(D_i;\theta_0)\}  = \left .\mathbb{V}\left \{ \nabla_\theta L_{Cox}^{(s)}(D_{i_1},D_{i_2},\dots,D_{i_s}|\theta),\nabla_\theta L_{Cox}^{(s)}(D_{i_1},\Tilde D_{{i}_2},\dots,\Tilde D_{{i}_s}|\theta)\right \}\right \rvert_{\theta=\theta_0}=  \Sigma_{(s|1)}.$

 On the other hand, for the second item in (\ref{ap-eq: asymptotic normal}), by the law of large numbers,
 \begin{equation}\label{ap-eq:asymptotic normal lln}
\frac{1}{\binom{n}{s}}\sum_{D(s)\subset D(n)} \nabla^2_\theta L_{Cox}^{(s)}(\theta_0)\to^p \mathbb{E}[\nabla^2_\theta L_{Cox}^{(s)}(\theta)]|_{\theta=\theta_0}=H_s,
\end{equation}
which is positive definite by Lemma \ref{lemma: strongly convex}.

By (\ref{ap-eq: asymptotic normal}), (\ref{ap-eq:asymptotic normal clt}), (\ref{ap-eq:asymptotic normal lln}), and Slutsky's theorem, we have
$$\sqrt{n}(\tilde{\theta}_{n}^{SB(s)}-\theta_0) \to^d N(0,s^2H^{-1}_s\Sigma_{(s|1)}(H_s^{-1})^T).$$

By a similar derivation, we can establish the asymptotic normality of $\tilde{\theta}^{FB(s)}_{n}$ in terms of $m$. That is,
$$\sqrt{m}(\tilde{\theta}_{n}^{FB(s)}-\theta_0) \to^d N(0,H^{-1}_s\Sigma_s(H_s^{-1})^T).$$
Because $s$ is fixed and $n = m\times s$, the asymptotic variance of $\tilde{\theta}_{n}^{FB(s)}$ in terms of the sample size $n$ is $\displaystyle\lim_{n\to\infty}n\mathbb{V}[\tilde{\theta}_{n}^{FB(s)}] = s\displaystyle\lim_{m\to\infty}m\mathbb{V}[\tilde{\theta}_{n}^{FB(s)}]  = sH^{-1}_s\Sigma_s(H_s^{-1})^T$. This shows
$$\sqrt{n}(\tilde{\theta}_{n}^{FB(s)}-\theta_0) \to^d N(0,sH^{-1}_s\Sigma_s(H_s^{-1})^T).$$

\end{proof}

\subsection{Proof of Theorem 4}
\begin{theorem}\label{thm:non-asymp-bound}
Consider the SGD procedure \begin{equation}
\hat\theta_{t+1} = \Pi_{\mathbb{R}^p_B}[ \hat\theta_t - \gamma_t \nabla_\theta L_{Cox}^{(s)}(\hat\theta_t)],    
\end{equation} with learning rate $\gamma_t = \frac{C}{t^\alpha}$ where the constant $C>0$ and $\alpha\in[0,1]$. Under Cox model, with assumptions (A1)-(A3), (R1), (R2), assume $\lVert \theta_0\rVert \leq B$, then for any integer $s\geq 2$, we have 
\begin{equation}\label{eq:non-asymp-bound}
\mathbb{E}\lVert \hat\theta_t-\theta_0\rVert^2\leq  
\begin{cases}
    \{\delta_0^2+D^2C^2\varphi_{1-2\alpha}(t)\}\exp(-\frac{\mu C}{2}t^{1-\alpha})+\frac{2D^2C^2}{\mu t^\alpha}       & \quad \text{if } \alpha\in [0,1);\\
    \delta_0^2 t^{-\mu C}+2D^2C^2t^{-\mu C}\varphi_{\mu C-1}(t)  & \quad \text{if } \alpha=1,
\end{cases}
\end{equation}
where $\delta_0$ is the distance between the initial point of SGD and $\theta_0$, $D=\displaystyle\max_{\theta\in \mathbb{R}^p_B,D(s)} \lVert \nabla_\theta L_{Cox}^{(s)}(\theta) \rVert$, and $\mu$ is a constant in Lemma \ref{lemma: strongly convex}. The function $\varphi_{\beta}(t):\mathbb{R}^+ \setminus \{0\}\to\mathbb{R}$ is given by
$$\varphi_{\beta}(t) = 
\begin{cases}
    \frac{t^\beta-1}{\beta}    & \quad \text{if } \beta\neq 0,\\
    \log t  & \quad \text{if } \beta=0.
\end{cases}
$$
\end{theorem}
\begin{proof}
We verify the conditions (H1), (H3), (H5) of Theorem 2 in \citet{moulines2011non}:
\subsubsection*{Condition (H1):} For any $\theta\in \mathbb{R}^p_B$, it can be verified that under the assumption (A1)-(A3), $\mathbb{E}[\{\nabla_\theta L_{Cox}^{(s)}(\theta)\}^2]<\infty$ and is therefore square-integrable. Since $D(s)$ is generated by randomly picking $s$ i.i.d. samples from the population, therefore for $\forall \theta\in \mathbb{R}^p_B$ and $\forall t\geq 1$,
$$ \mathbb{E}_{D_t(s)}[\nabla_\theta L_{Cox}^{(s)}(\theta)|D_k(s),k=1,\dots,t-1]=\mathbb{E}_{D_t(s)}[\nabla_\theta L_{Cox}^{(s)}(\theta)]=\nabla_\theta \mathbb{E}[L_{Cox}^{(s)}(\theta)]. $$
\subsubsection*{Condition (H3):} The strong convexity of $\mathbb{E}[L_{Cox}^{(s)}(\theta)]$ over $\mathbb{R}^p_B$ is guaranteed by Lemma \ref{lemma: strongly convex}.
\subsubsection*{Condition (H5):} For any $t$ and any $\theta\in \mathbb{R}^p_B$ , it can be verified that under the assumption (A1)-(A3), $ L_{Cox}^{(s)}(D_t(s);\theta)$ is convex, differentiable. Moreover, 
\begin{align*}
\left\rVert {\nabla_\theta}L_{Cox}^{(s)}(D_t(s);\theta) \right\rVert & = \left\rVert\frac{1}{s}\sum_{i:D_i\in D_t(s)} \Delta_i\left\{ -X_i+\frac{\sum_{j:D_j\in D_t(s)} Y_j(T_i)\exp(X_j^T \theta)X_j}{\sum_{j:D_j\in D_t(s)} Y_j(T_i)\exp(X_j^T \theta)} \right\}\right\rVert \\
& \leq  \frac{1}{s}\sum_{i:D_i\in D_t(s)} \left\{\left\rVert X_i\right\rVert+\left\rVert\frac{\sum_{j:D_j\in D_t(s)} Y_j(T_i)\exp(X_j^T \theta)X_j}{\sum_{j:D_j\in D_t(s)} Y_j(T_i)\exp(X_j^T \theta)} \right\rVert \right\} \\
& \leq  \frac{1}{s}\sum_{i:D_i\in D_t(s)} \left\{\left\rVert X_i\right\rVert+\sum_{j:D_j\in D_t(s)} \left\rVert X_j \right\rVert \right\}\leq p+sp.
\end{align*}
This shows that the gradient is uniformly bounded on $\theta\in \mathbb{R}^p_B$ for any $t$.

Therefore, the conditions required by theorem 2 in \citet{moulines2011non} hold, and Theorem \ref{thm:non-asymp-bound} follows. 
\end{proof}

\section{Technical Lemmas and Proof}
\subsection{Lemma 1}

\begin{lemma}\label{lemma: minimizer does not depend on s}
Let $L^{(s)}_{0}(f) := \mathbb{E}[L^{(s)}_{Cox}(f)]$. Under the Cox model, with assumptions (A1)-(A3), (N1) and for any integer $s \geq 2$ and for any constant $c>0$, we have
$$L^{(s)}_{0}(f)-L^{(s)}_{0}(f_0)\asymp  d^2(f,f_0)$$
for all $f\in \{f:\lVert f\rVert_\infty\leq c, \mathbb{E}[f(X)]=0\}$, where $d(f,f_0)=[\mathbb{E}\{f(X)-f_0(X)\}^2]^{1/2}$.
\end{lemma}

\begin{proof}
Let $f_* = f-f_0$. Since $\mathbb{E}[f(X)]=\mathbb{E}[f_0(X)]=0$, we have $\mathbb{E}[f_*(X)]=0$. Then 
\begin{equation}\label{eq: difference of population loss}
\begin{split}
& L^{(s)}_0(f) - L^{(s)}_0(f_0) \\
= & -\mathbb{E}\left[ \frac{1}{s}\sum_{j:D_j\in D(s)}\Delta_jf_*(X_j)\right]+\\
& \mathbb{E}\left[ \frac{1}{s}\sum_{j:D_j\in D(s)}\Delta_j\left\{\log \sum_{k:D_k\in D(s)}Y_k(T_j)\exp(f(X_k))-\log \sum_{k:D_k\in D(s)}Y_k(T_j)\exp(f_0(X_k))\right\}\right] \\
=  & -\mathbb{E}\left[ \Delta f_*(X)\right]+ \mathbb{E}\left[ \frac{1}{s}\sum_{j:D_j\in D(s)}\Delta_j\left\{K_{D(s)}(T_j,1)-K_{D(s)}(T_j,0)\right\}\right],
\end{split}
\end{equation}
where $K_{D(s)}(u,v) = \log \sum_{k:D_k\in D(s)}Y_k(u)\exp\{f_0(X_k)+vf_*(X_k)\} $. 
Denote
$$\frac{\partial}{\partial v} K_{D(s)}(u,v) = h_{D(s)}(u,v)=\frac{\sum_{k:D_k\in D(s)}Y_k(u)\exp\{f_0(X_k)+vf_*(X_k)\} f_*(X_k)}{\sum_{k:D_k\in D(s)}Y_k(u)\exp\{f_0(X_k)+vf_*(X_k)\} },$$
then we have
\begin{align*}
\frac{\partial}{\partial v}h_{D(s)}(u,v) = &   \frac{\sum_{k:D_k\in D(s)}Y_k(u)\exp\{f_0(X_k)+vf_*(X_k)\} f^2_*(X_k)}{ \sum_{k:D_k\in D(s)}Y_k(u)\exp\{f_0(X_k)+vf_*(X_k)\} } - \\
 & \left [\frac{ \sum_{k:D_k\in D(s)}Y_k(u)\exp\{f_0(X_k)+vf_*(X_k)\} f_*(X_k)}{ \sum_{k:D_k\in D(s)}Y_k(u)\exp\{f_0(X_k)+vf_*(X_k)\} }\right] ^2 \\
 =& \left\{\sum_{k:D_k\in D(s)}w_k(u,v)f^2_*(X_k)\right\}-\left\{\sum_{k:D_k\in D(s)}w_k(u,v)f_*(X_k)\right \}^2,
\end{align*}
where $w_k(u,v) = \frac{ Y_k(u)\exp\{f_0(X_k)+vf_*(X_k)\} }{ \sum_{k:D_k\in D(s)}Y_k(u)\exp\{f_0(X_k)+vf_*(X_k)\} }$ and $\sum_{k:D_k\in D(s)}w_k(u,v) = 1$.

Because $Y_j(T_j)=1$, therefore 
\begin{align*}
& \sum_{k:D_k\in D(s)}Y_k(T_j)\exp\{f_0(X_k)+vf_*(X_k)\} \\
=& \left [ \exp\{f_0(X_j)+vf_*(X_j)\} + \sum_{\substack{k:D_k\in D(s)\\ k \neq j }}Y_k(u)\exp\{f_0(X_k)+vf_*(X_k)\}\right ]
\end{align*}
is bounded away from zero due to the boundedness of $f$. This ensures that $K_{D(s)}(T_j,v)$, $h_{D(s)}(T_j,v)$, and $\frac{\partial}{\partial v}h_{D(s)}(T_j,v)$ are well-defined for $j:D_j\in D(s)$.
By Taylor expansion of $K_{D(s)}(T_j,v)$ at $v = 0$, there is a $\Bar{v}\in [0,1]$, such that
$$K_{D(s)}(T_j,1)-K_{D(s)}(T_j,0) = h_{D(s)}(T_j,0) + \frac{1}{2}\frac{\partial}{\partial v}h_{D(s)}(T_j,\Bar{v}).$$
Hence (\ref{eq: difference of population loss}) can be re-expressed as
\begin{equation}\label{eq: difference of population loss2}
\begin{split}
& L^{(s)}_0(f) - L^{(s)}_0(f_0) \\
=  & -\mathbb{E}\left[ \Delta f_*(X)\right]+ \mathbb{E}\left[ \frac{1}{s}\sum_{j:D_j\in D(s)}\Delta_j\left\{h_{D(s)}(T_j,0) + \frac{1}{2}\frac{\partial}{\partial v}h_{D(s)}(T_j,\Bar{v})\right\}\right]\\
=  & -\mathbb{E}\left[ \Delta f_*(X)\right]+ \mathbb{E}\left[ \frac{1}{s}\sum_{j:D_j\in D(s)}\Delta_jh_{D(s)}(T_j,0)\right]+ 
\mathbb{E}\left[ \frac{1}{2s}\sum_{j:D_j\in D(s)}\Delta_j\frac{\partial}{\partial v}h_{D(s)}(T_j,\Bar{v})\right].
\end{split}
\end{equation}

We will first show that $\mathbb{E}f^2_*(X)\lesssim\mathbb{E}\left[ \frac{1}{2s}\sum_{j:D_j\in D(s)}\Delta_j\frac{\partial}{\partial v}h_{D(s)}(T_j,\Bar{v})\right]\lesssim \mathbb{E}f^2_*(X)$ and then demonstrate $\mathbb{E}\left[ \Delta f_*(X)\right]= \mathbb{E}\left[ \frac{1}{s}\sum_{j:D_j\in D(s)}\Delta_jh_{D(s)}(T_j,0)\right]$. These results lead to our conclusion $L^{(s)}_{0}(f)-L^{(s)}_{0}(f_0)\asymp  d^2(f,f_0)$.

For any $j:D_j\in D(s)$, because $\frac{\partial}{\partial v}h_{D(s)}(T_j,\Bar{v})\leq \sum_{k:D_k\in D(s)}f^2_*(X_k)$, hence
\begin{equation}\label{eq: upper bound for loss difference}
\mathbb{E}\left[ \frac{1}{2s}\sum_{j:D_j\in D(s)}\Delta_j\frac{\partial}{\partial v}h_{D(s)}(T_j,\Bar{v})\right]\lesssim \mathbb{E}f^2_*(X).
\end{equation}
Now we show $\mathbb{E}f^2_*(X)\lesssim\mathbb{E}\left[ \frac{1}{2s}\sum_{j:D_j\in D(s)}\Delta_j\frac{\partial}{\partial v}h_{D(s)}(T_j,\Bar{v})\right]$. Notice that
\begin{align*}
&\frac{\partial}{\partial v}h_{D(s)}(T_j,\Bar{v}) \\
&= \left\{\sum_{k:D_k\in D(s)}w_k(T_j,\Bar{v})f^2_*(X_k)\right\}\left\{\sum_{k:D_k\in D(s)}w_k(T_j,\Bar{v})\right\}-\left\{\sum_{k:D_k\in D(s)}w_k(T_j,\Bar{v})f_*(X_k)\right\}^2\\
& = \sum_{\substack{i,l\in D(s) \\ i<l }}w_i(T_j,\Bar{v})w_l(T_j,\Bar{v})\{f^2_*(X_i)+f^2_*(X_l)-2f_*(X_i)f_*(X_l)\}.
\end{align*}
Let $X^{(s)}=\{X_i: X_i\in D_i \text{ where }D_i\in D(s)\}$. Notice that $\mathbb{E}[\Delta_j w_i(T_j,\Bar{v})w_l(T_j,\Bar{v})|X^{(s)}] = $
\begin{align*}
\mathbb{E}\left . \left[\Delta_j\frac{ Y_i(T_j)\exp\{f_0(X_i)+\Bar{v}f_*(X_i)\} }{ \sum_{k:D_k\in D(s)}Y_k(T_j)\exp\{f_0(X_k)+\Bar{v}f_*(X_k)\} }\frac{ Y_l(T_j)\exp\{f_0(X_l)+\Bar{v}f_*(X_l)\} }{ \sum_{k:D_k\in D(s)}Y_k(T_j)\exp\{f_0(X_k)+\Bar{v}f_*(X_k)\} }\right|X^{(s)}\right],
\end{align*} is non-negative and is non-zero only when $\Delta_j = 1, T_i\geq T_j, T_l\geq T_j$. Assumption (A1) and (A2) imply that there exists a constant $0<a_1<\infty$ such that
\begin{equation}\label{ap-eq: prob is bounded from below}
\begin{split}
P(\Delta_j = 1, T_i\geq T_j, T_l \geq T_j|X^{(s)})& \geq P(\Delta_j = 1, T_i\geq \tau, T_l \geq \tau|X_i,X_j,X_l) \\
&= P(\Delta_j = 1|X_j)P(T_i\geq \tau|X_i)P(T_l\geq \tau|X_l)\\
& > a_1.
\end{split}
\end{equation}
Hence, due to the boundedness of X from (A3) and the smoothness of $f_0$ from (N1), $\frac{ Y_i(T_j)\exp\{f_0(X_i)+\Bar{v}f_*(X_i)\} }{ \sum_{k:D_k\in D(s)}Y_k(T_j)\exp\{f_0(X_k)+\Bar{v}f_*(X_k)\} }$ is bounded from below when $T_i\geq T_j$.
Together with (\ref{ap-eq: prob is bounded from below}) it indicates that $\mathbb{E}[\Delta_j w_i(T_j,\Bar{v})w_l(T_j,\Bar{v})|X]>a_2$ where $a_2$ is a positive constant. 

This shows that
\begin{equation}\label{eq: lower bound for loss difference}
\begin{split}
& \mathbb{E}\left[\Delta_j \frac{\partial}{\partial v}h_{D(s)}(T_j,\Bar{v})\right] \\
& =\mathbb{E}\left[\Delta_j\sum_{\substack{i,l\in D(s), i<l }}w_i(T_j,\Bar{v})w_l(T_j,\Bar{v})\{f^2_*(X_i)+f^2_*(X_l)-2f_*(X_i)f_*(X_l)\}\right ] \\
& =\mathbb{E}\left[\sum_{\substack{i,l\in D(s), i<l }}\Delta_j w_i(T_j,\Bar{v})w_l(T_j,\Bar{v})\{f^2_*(X_i)+f^2_*(X_l)-2f_*(X_i)f_*(X_l)\}\right ] \\
& =\mathbb{E}\left[\sum_{\substack{i,l\in D(s), i<l }}\mathbb{E}\left.\left[\Delta_j w_i(T_j,\Bar{v})w_l(T_j,\Bar{v})\right\rvert X^{(s)}\right]\{f^2_*(X_i)+f^2_*(X_l)-2f_*(X_i)f_*(X_l)\}\right ] \\
& \geq a_2\mathbb{E}\left[\sum_{\substack{i,l\in D(s), i<l }}\{f^2_*(X_i)+f^2_*(X_l)-2f_*(X_i)f_*(X_l)\}\right ] \\
& = a_2\mathbb{E}\left[\sum_{\substack{i,l\in D(s), i<l }}\{f^2_*(X_i)+f^2_*(X_l)\}\right ]\gtrsim \mathbb{E}f^2_*(X_k),
\end{split}
\end{equation}
where the last equation holds since $i,l$ are i.i.d. samples and $$\mathbb{E}[f_*(X_i)f_*(X_l)]=\mathbb{E}[f_*(X_i)]\mathbb{E}[f_*(X_l)]=0.$$

So far, we have shown that $\mathbb{E}f^2_*(X)\lesssim\mathbb{E}\left[ \frac{1}{2s}\sum_{j:D_j\in D(s)}\Delta_j\frac{\partial}{\partial v}h_{D(s)}(T_j,\Bar{v})\right]\lesssim \mathbb{E}f^2_*(X)$. What remains is to demonstrate $\mathbb{E}\left[ \Delta f_*(X)\right]= \mathbb{E}\left[ \frac{1}{s}\sum_{j:D_j\in D(s)}\Delta_jh_{D(s)}(T_j,0)\right]$, which is proved below:

By (\ref{ap-eq:compensator}), we have
\begin{equation}\label{eq: martingale equivalence}
\begin{split}
&\mathbb{E}\left[ \frac{1}{s}\sum_{j:D_j\in D(s)}\Delta_j h_{D(s)}(T_j,0) \right ]  \\
& = \mathbb{E}\left[ \frac{1}{s}\sum_{j:D_j\in D(s)}\int_0^\tau\frac{\sum_{k:D_k\in D(s)}Y_k(u)\exp\{f_0(X_k)\} f_*(X_k)}{\sum_{k:D_k\in D(s)}Y_k(u)\exp\{f_0(X_k)\} }d N_j(u)\right ] \\
& = \mathbb{E}\left[ \frac{1}{s}\int_0^\tau\frac{\sum_{k:D_k\in D(s)}Y_k(u)\exp\{f_0(X_k)\} f_*(X_k)}{\sum_{k:D_k\in D(s)}Y_k(u)\exp\{f_0(X_k)\} }\sum_{j:D_j\in D(s)}d N_j(u)\right ] \\
& = \mathbb{E}\left[ \frac{1}{s}\int_0^\tau\sum_{k:D_k\in D(s)}Y_k(u)\exp\{f_0(X_k)\} f_*(X_k)\lambda_0(u)du\right ] \\
& = \mathbb{E}\left[ \frac{1}{s}\int_0^\tau\sum_{k:D_k\in D(s)}f_*(X_k)d N_k(u)\right ] = \mathbb{E}\left[ \frac{1}{s}\sum_{j:D_j\in D(s)}\Delta_j f_*(X_j) \right ].
\end{split}
\end{equation}
Therefore, combining (\ref{eq: difference of population loss}), (\ref{eq: difference of population loss2}), (\ref{eq: upper bound for loss difference}), (\ref{eq: lower bound for loss difference}) and (\ref{eq: martingale equivalence}), we have
$$\mathbb{E}f^2_*(X) \lesssim L^{(s)}_0(f) - L^{(s)}_0(f_0) \lesssim \mathbb{E}f^2_*(X).$$
The result follows. 
\end{proof}

\subsection{Lemma 2}
\begin{lemma}\label{lemma: strongly convex}
Suppose the integer $s\geq 2$, the constant $B>0$, and $\theta_0\in \mathbb{R}^p_B$. Then under Cox model, with assumptions (A1)-(A3), (R1), (R2), there exist a constant $\mu>0$ such that for any $\theta\in \mathbb{R}^p_B$
$$\nu^T\mathbb{E}[\nabla^2_\theta L^{(s)}_{Cox}(\theta)]\nu \geq \mu >0\quad  \forall \nu\in \{\nu:\lVert \nu \rVert_2 = 1\}. $$
\end{lemma}
\begin{proof}
Let $S^{(k)}_s (u,\theta) := \sum_{j=1}^s Y_j(u)\exp(X_j^T \theta)X_j^{\otimes k}$. Under Cox regression, we have
$$\nabla^2_\theta L^{(s)}_{Cox}(\theta) = \frac{1}{s}\sum_{i=1}^s \Delta_i \left\{\frac{S^{(2)}_s (T_i,\theta)}{S^{(0)}_s (T_i,\theta)}-\frac{S^{(1)}_s (T_i,\theta)^{\otimes 2}}{S^{(0)}_s (T_i,\theta)^{\otimes 2}} \right\},$$
and it is bounded over $\theta\in \mathbb{R}^p_B$ due to the boundedness of the $X_i$. 

\begin{align*}
\mathbb{E}\left[\nabla^2_\theta L_{Cox}^{(s)}(\theta)\right] \geq & \left.\mathbb{E}\left[\frac{1}{s}\sum_{i=1}^s \left\{\frac{S^{(2)}_s (T_i,\theta)}{S^{(0)}_s (T_i,\theta)}-\frac{S^{(1)}_s (T_i,\theta)^{\otimes 2}}{S^{(0)}_s (T_i,\theta)^{\otimes 2}} \right\}\right|\sum_{i=1}^s\Delta_i=s\right]\mathbb{P}\left(\sum_{i=1}^s\Delta_i=s\right) \\
= & \left.\mathbb{E}\left[\frac{1}{s}\sum_{i=1}^s \frac{C_s(T_i,\theta)}{S^{(0)}_s (T_i,\theta)^{\otimes 2}}\right|\sum_{i=1}^s\Delta_i=s\right]\mathbb{P}\left(\sum_{i=1}^s\Delta_i=s\right),
\end{align*}
where $C_s(u,\theta):=S^{(2)}_s (u,\theta)S^{(0)}_s (u,\theta)-S^{(1)}_s (u,\theta)^{\otimes 2} $. 

By assumption (A2), there is a constant $0<b_1<\infty$ such that $\mathbb{P}(\Delta=1)>b_1$, so that
$$\mathbb{P}\left(\sum_{i=1}^s\Delta_i=s\right)= \prod_{i=1}^s\mathbb{P}(\Delta_i=1) = b_1^s >0,$$
for $s$ i.i.d. samples and
$$\mathbb{E}\left[\nabla^2_\theta L_{Cox}^{(s)}(\theta)\right] \geq \left.\mathbb{E}\left[\frac{b_1^s}{s}\sum_{i=1}^s \frac{C_s(T_i,\theta)}{S^{(0)}_s (T_i,\theta)^{\otimes 2}}\right|\sum_{i=1}^s\Delta_i=s\right].$$

We introduce the bounds of several terms as preparation for the proof. By assumptions (A3) and (R1), we have 
$$ S^{(0)}_s (u,\theta) \leq s\exp(pB), $$
and there exists $0<b_2<\infty$ such that
$$\quad \lVert C_s(u,\theta) \rVert_F < b_2 <\infty \quad \forall u\in (0,\tau],\ \forall \theta\in \mathbb{R}^p_B.$$

We prove Lemma 2 by contradiction through the following two claims:
\begin{itemize}
\item Claim 1: If Lemma 2 does not hold, then there exists a $\nu^*\in S_\nu=\{\nu:\lVert \nu \rVert_2 = 1\}$ such that 
$$ \left.\mathbb{P}\left(\sum_{i=1}^s {\nu^*}^TC_s(T_i,\theta)\nu^*\right|\sum_{i=1}^s\Delta_i=s\right)= 1.$$
\item Claim 2: Under the assumptions of Lemma 2, we have 
$$ \left.\mathbb{P}\left(\sum_{i=1}^s {\nu}^TC_s(T_i,\theta)\nu\right|\sum_{i=1}^s\Delta_i=s\right)<1. $$
for any $\nu\in S_\nu=\{\nu:\lVert \nu \rVert_2 = 1\}$.
\end{itemize}
The contradiction of two claims shows that Lemma 2 must hold.
\subsubsection*{Proof of claim 1}

If Lemma 2 does not hold, then for any $\mu>0$, we can find $\nu_\mu\in S_\nu$ such that
\begin{align*}
\mu > & \nu_\mu^T\mathbb{E}[\nabla^2_\theta L_{Cox}^{(s)}(\theta)]\nu_\mu \\
\geq & \left.\mathbb{E}\left[\frac{b_1^s}{s}\sum_{i=1}^s \frac{\nu_\mu^TC_s(T_i,\theta)\nu_\mu}{S^{(0)}_s (T_i,\theta)^{\otimes 2}}\right|\sum_{i=1}^s\Delta_i=s\right] \\
\geq & \frac{b_1^s}{s^3\exp(2pB)}\left.\mathbb{E}\left[\sum_{i=1}^s \nu_\mu^TC_s(T_i,\theta)\nu_\mu\right|\sum_{i=1}^s\Delta_i=s\right] \\
\geq & \frac{b_1^s}{s^3\exp(2pB)}\left.\mathbb{E}\left[\left(\sum_{i=1}^s \nu_\mu^TC_s(T_i,\theta)\nu_\mu\right)I_{(\sum_{i=1}^s \nu_\mu^TC_s(T_i,\theta)\nu_\mu>\epsilon)}\right|\sum_{i=1}^s\Delta_i=s\right] \\
> & \frac{b_1^s\epsilon}{s^3\exp(2pB)}\left.\mathbb{P}\left(\sum_{i=1}^s \nu_\mu^TC_s(T_i,\theta)\nu_\mu>\epsilon\right|\sum_{i=1}^s\Delta_i=s\right) ,
\end{align*}
where $\epsilon$ is an arbitrary positive value. 

Denote $b_3:=\frac{s^3\exp(2pB)}{b_1^s}$. Therefore, for any $\mu$, we can find $\nu_\mu$ such that
$$\left.\mathbb{P}\left(\sum_{i=1}^s \nu_\mu^TC_s(T_i,\theta)\nu_\mu>\epsilon \right|\sum_{i=1}^s\Delta_i=s\right) <   b_3\frac{\mu}{\epsilon} \quad \forall \epsilon >0  $$
or equivalently,
\begin{align*}
\left.\mathbb{P}\left(\sum_{i=1}^s \nu_\mu^TC_s(T_i,\theta)\nu_\mu\leq \epsilon \right|\sum_{i=1}^s\Delta_i=s\right) > 1 -   b_3\frac{\mu}{\epsilon} \quad \forall\epsilon >0 .
\end{align*}

Let $\delta>0$ and consider an infinite sequence $\nu_1, \dots, \nu_k,\dots $ with 
$$\epsilon_k = \frac{1}{k} \quad \mu_k(\delta) = \frac{\delta}{kb_3},$$
then for $\forall \delta>0, \forall k$, we have
\begin{equation}
\left.\mathbb{P}\left(\sum_{i=1}^s \nu_\mu^TC_s(T_i,\theta)\nu_\mu\leq \frac{1}{k} \right|\sum_{i=1}^s\Delta_i=s\right) > 1 -   \delta. \label{cd: claim 1}
\end{equation}
Since $S_\nu=\{\nu:\lVert \nu \rVert_2 = 1\}$ is a compact space, the sequence $\{\nu_k\}$ has an infinite sub-sequence $\{\nu'_k\}$ converging to a point $\nu^*\in S_\nu$. For this converging point $\nu^*$ and all $k$,
\begin{align*}
\nu^{*T}C_s(u,\theta)\nu^* = & (\nu^{*}-\nu'_k+\nu'_k)^TC_s(u,\theta)(\nu^{*}-\nu'_k+\nu'_k) \\
= & (\nu^{*}-\nu'_k)^TC_s(u,\theta)(\nu^{*}-\nu'_k) + {\nu'}^{T}_k C_s(u,\theta)\nu'_k + 2 {\nu'}^{T}_k C_s(u,\theta)(\nu^{*}-\nu'_k)  \\
\leq & \lVert\nu^{*}-\nu'_k\rVert_2^2 \lVert C_s(u,\theta)\rVert_F + {\nu'}^{T}_k C_s(u,\theta)\nu'_k + 2 \lVert\nu'_k\rVert_2 \lVert C_s(u,\theta)\rVert_F \lVert\nu^{*}-\nu'_k\rVert_2  \\
\leq &{\nu'}^{T}_k C_s(u,\theta)\nu'_k + 4 b_2 \lVert\nu^{*}-\nu'_k\rVert_2.
\end{align*}
The last inequality follows $\lVert\nu^{*}-\nu'_k\rVert_2\leq \lVert\nu^{*}\rVert_2+\lVert\nu'_k\rVert_2 =2 $ and $\lVert C_s(u,\theta)\rVert_F < b_2$.  The first inequality comes from the fact that for vectors $a,b$ and matrix $M$,
\begin{align*}
a^TMb =  \langle a, Mb \rangle
\leq  \lVert a \rVert_2 \lVert Mb \rVert_2 
\leq  \lVert a \rVert_2 \lVert M \rVert_F \lVert b \rVert_2.
\end{align*}

For $\forall \delta>0, \forall \zeta >0$, consider $k>K_1>\frac{2}{\zeta} $ then by (\ref{cd: claim 1}),
\begin{equation}
\begin{split}
& \left.\mathbb{P}\left(\sum_{i=1}^s {\nu'}^{T}_k C_s(T_i,\theta){\nu'}_k\leq \frac{\zeta}{2} \right|\sum_{i=1}^s\Delta_i=s\right) \\
\geq & \left.\mathbb{P}\left(\sum_{i=1}^s {\nu'}^{T}_k C_s(T_i,\theta){\nu'}_k\leq \frac{1}{k} \right|\sum_{i=1}^s\Delta_i=s\right)   \quad\quad \text{(because } \frac{1}{k}<\frac{\zeta}{2})\\
> & 1 - \delta.
\end{split} \label{cd: claim 1-1}
\end{equation}

On the other hand, since $\nu'_k \to \nu^*$ when $k\to \infty$, we have 
\begin{equation*}
\forall \zeta>0, \exists K_2>0, s.t. \ \forall k>K_2, \lVert\nu^{*}-\nu'_k\rVert_2 \leq \frac{\zeta}{4b_2}.
\end{equation*}

Then $\forall \delta>0$ and $\forall \zeta > 0$, we have that for all $k>\max(K_1,K_2)$,
\begin{align*}
& \left.\mathbb{P}\left(\sum_{i=1}^s {\nu^*}^{T} C_s(T_i,\theta){\nu^*}\leq \zeta \right|\sum_{i=1}^s\Delta_i=s\right) \\
\geq & \left.\mathbb{P}\left(\sum_{i=1}^s {\nu'}^{T}_k C_s(T_i,\theta){\nu'}_k+4b_2\lVert\nu^{*}-\nu'_k\rVert_2\leq \zeta \right|\sum_{i=1}^s\Delta_i=s\right) \\
\geq & \left.\mathbb{P}\left(\sum_{i=1}^s {\nu'}^{T}_k C_s(T_i,\theta){\nu'}_k\leq \frac{\zeta}{2} \text{ and } 4b_2\lVert\nu^{*}-\nu'_k\rVert_2\leq \frac{\zeta}{2} \right|\sum_{i=1}^s\Delta_i=s\right) \\
= & \left.\mathbb{P}\left(\sum_{i=1}^s {\nu'}^{T}_k C_s(T_i,\theta){\nu'}_k\leq \frac{\zeta}{2}\right|\sum_{i=1}^s\Delta_i=s\right) \\
\geq &1 - \delta.
\end{align*}

Let $\zeta,\delta\to 0$, we have 
$$ \left.\mathbb{P}\left(\sum_{i=1}^s {\nu^*}^{T} C_s(T_i,\theta){\nu^*}=0 \right|\sum_{i=1}^s\Delta_i=s\right) = 1. $$
This completes the proof of claim 1.

\subsubsection*{Proof of claim 2}
We first investigate when does $\sum_{i=1}^s {\nu}^{T} C_s(T_i,\theta){\nu}=0$. Consider $u\in (0,\tau)$ and $\nu\in S_\nu$. We have
\begin{align*}
\nu^{T} C_s(u,\theta)\nu = & \nu^T\{S_s^{(2)} (u,\theta)S_s^{(0)} (u,\theta) - (S_s^{(1)} (u,\theta))^{\otimes 2}\}\nu  \\
= & \nu^TS_s^{(2)} (u,\theta)S_s^{(0)} (u,\theta)\nu^T - \nu^TS_s^{(1)} (u,\theta)S_s^{(1)} (u,\theta)^T\nu\\
= & \nu^T\{\sum_{j=1}^s Y_j(u)\exp(X_j^T \theta)X_jX_j^T\}\{\sum_{j=1}^s Y_j(u)\exp(X_j^T \theta)\}\nu - \\
&  \quad  \quad  \quad  \quad \nu^T\{\sum_{j=1}^s Y_j(u)\exp(X_j^T \theta)X_j\}\{\sum_{j=1}^s Y_j(u)\exp(X_j^T \theta)X_j\}^T\nu\\
= & \{\sum_{j=1}^s Y_j(u)\exp(X_j^T \theta)\nu^TX_jX_j^T\nu\}\{\sum_{j=1}^s Y_j(u)\exp(X_j^T \theta)\} - \\
& \quad  \quad  \quad  \quad \{\sum_{j=1}^s Y_j(u)\exp(X_j^T \theta)\nu^TX_j\}\{\sum_{j=1}^s Y_j(u)\exp(X_j^T \theta)X_j^T\nu\}\\
= & \{\sum_{j=1}^s\left(\sqrt{ Y_j(u)\exp(X_j^T \theta)}\nu^TX_j\right)^2\}\{\sum_{j=1}^s \left(\sqrt{ Y_j(u)\exp(X_j^T \theta)}\right)^2\} - \\
&  \quad  \quad  \quad  \quad \{\sum_{j=1}^s Y_j(u)\exp(X_j^T \theta)\nu^TX_j \}^2 \\
\geq & 0 \quad \quad (\text{by Cauchy–Schwarz Inequality})
\end{align*}
For any realization $\{(T_i,X_i,\Delta_i = 1)\}_{i=1}^s$, suppose no ties of event, there exist $k\in\{1,\dots,s\}$ such that $Y_i(T_k) = 1$ for all $i$. Hence, $\sum_{i=1}^s {\nu}^{T} C_s(T_i,\theta){\nu}=0$ if and only if $X_1^T\nu  = X_2^T\nu = \dots = X_s^T\nu$, conditional on $\sum_{i=1}^s\Delta_i = s$, so that for any $\nu\in S_\nu$,
\begin{equation*}
\left.\mathbb{P}\left(\sum_{i=1}^s {\nu}^{T} C_s(T_i,\theta){\nu}=0 \right|\sum_{i=1}^s\Delta_i=s\right) = \mathbb{P}\left(X_1^T\nu  = X_2^T\nu = \cdots = X_s^T\nu \right|\sum_{i=1}^s\Delta_i=s).
\end{equation*}
By assumption (R2), the conditional probability on the right-hand side must be less than 1, therefore $\left.\mathbb{P}\left(\sum_{i=1}^s {\nu}^{T} C_s(T_i,\theta){\nu}=0 \right|\sum_{i=1}^s\Delta_i=s\right) < 1$. This completes the proof of claim 2 and indicates
$$ \left.\mathbb{P}\left(\sum_{i=1}^s {\nu^*}^TC_s(T_i,\theta)\nu^*\right|\sum_{i=1}^s\Delta_i=s\right)<1,$$
which contradicts claim 1. Therefore, Lemma 2 must hold.
\end{proof}

\subsection{Lemma 3}
\begin{lemma}
\label{lemma: change variable}
Let $n$ be a fixed number and $Z_1,\dots,Z_n$ $\in \mathcal{Z}$ which are independently and identically distributed (i.i.d.) from $F_\mathcal{Z}$, and the function $f_1:\mathcal{Z}^n\mapsto \mathbb{R}^{d_1}$ is symmetric to all the arguments. Another function $f_2:\mathcal{Z}\mapsto \mathbb{R}^{d_2}$. Then 
\begin{equation}
\mathbb{E}[f_1(Z_1,\dots,Z_n)\otimes f_2(Z_1)] = \mathbb{E}\left[f_1(Z_1,\dots,Z_n)\otimes f_2(Z_i)\right], \label{ap-eq: lemma2.1}
\end{equation}
\begin{equation}
\mathbb{E}\left[f_1(Z_1,\dots,Z_n)\otimes \left(\frac{\sum_i^n f_2(Z_i)}{n}\right)\right]  = \mathbb{E}[f_1(Z_1,\dots,Z_n)\otimes f_2(Z_1)],\label{ap-eq: lemma2.2}
\end{equation}
for any $i \in [n]$ when the expectation exists.

\begin{proof}
We first prove that Eq. (\ref{ap-eq: lemma2.1}) holds:
\begin{align*}
\mathbb{E}[f_1(Z_1,\dots,Z_i,\dots,Z_n)\otimes f_2(Z_1)] = & \mathbb{E}[f_1(Z_i,\dots,Z_1,\dots,Z_n)\otimes f_2(Z_1)] \\
= & \mathbb{E}[f_1(Z_1,\dots,Z_i,\dots,Z_n)\otimes f_2(Z_i)].
\end{align*}

The first equation is due to the symmetry of the function $f_1$. The second equation holds because switching the indices of i.i.d. samples (sample $1$ and sample $i$) will not change the expectation. 

This implies that
\begin{align*}
\mathbb{E}[f_1(Z_1,\dots,Z_n)\otimes \left(\frac{\sum_i^n f_2(Z_i)}{n}\right)] & = \frac{1}{n}\sum_i^n \mathbb{E}[f_1(Z_1,\dots,Z_n)\otimes f_2(Z_i)] \\
& = \frac{1}{n}\sum_i^n \mathbb{E}[f_1(Z_1,\dots,Z_n)\otimes f_2(Z_1)] \\
& = \mathbb{E}[f_1(Z_1,\dots,Z_n)\otimes f_2(Z_1)],
\end{align*}
where the second equation holds by applying the Eq. (\ref{ap-eq: lemma2.1}) to each term in the summation. This ends the proof of Lemma \ref{lemma: change variable}. 
\end{proof}
\end{lemma}

\subsection{Lemma 4}
\begin{lemma}
\label{lemma: final step}
$Z_1,Z_2$ $\in \mathcal{Z}$ are independently and identically distributed random variables from $F_\mathcal{Z}$. $f:\mathcal{Z}\mapsto \mathbb{R}^{d}$. $g(Z_1, Z_2):\mathcal{Z}^2\mapsto \mathbb{R}$ is a positive real-valued function and is symmetric to all arguments. Then
$$\mathbb{E}[[f(Z_1)-f(Z_2)]\otimes f(Z_2)g(Z_1,Z_2)] \preceq 0$$
and the equality holds when $\mathbb{P}(f(Z_1)=f(Z_2))=1$.

\begin{proof}
When $f$ is a real-valued function ($d=1$), because $ab\leq\frac{a^2+b^2}{2}$ for any $a,b \in \mathbb{R}^1$, \begin{align*}
\mathbb{E}[f(Z_1)f(Z_2)g(Z_1,Z_2)] & \leq \mathbb{E}[\frac{f(Z_1)^2+f(Z_2)^2}{2}g(Z_1,Z_2)] \\
& = \mathbb{E}[\frac{f(Z_2)^2+f(Z_2)^2}{2}g(Z_1,Z_2)] \\
& = \mathbb{E}[f(Z_2)^2g(Z_1,Z_2)],
\end{align*}
and $\mathbb{E}[[f(Z_1)-f(Z_2)]f(Z_2)g(Z_1,Z_2)]\leq 0$ follows immediately. The second equation holds because $\mathbb{E}[f(Z_1)^2g(Z_1,Z_2)] = \mathbb{E}[f(Z_2)^2g(Z_1,Z_2)]$
by Lemma \ref{lemma: change variable}.

The proof for $f(Z)\in \mathbb{R}^d$ follows the same logic. For any vector $X$ and $Y$ with the same dimension, we have 
$X^{\otimes 2} - X\otimes Y - Y \otimes X + Y^{\otimes 2} = (X-Y)^{\otimes 2} \succeq 0$. Therefore,
$$ X\otimes Y + Y \otimes X \preceq X^{\otimes 2} + Y^{\otimes 2}.$$
Let $X = f(Z_1)$ and $Y = f(Z_2)$, this yields $$f(Z_1)\otimes f(Z_2) + f(Z_2) \otimes f(Z_1) \preceq f(Z_1)^{\otimes 2} + f(Z_2)^{\otimes 2}.$$
Because $g$ is a positive real-valued function, the direction of inequality holds after multiplying $g(Z_1, Z_2)$ on both sides. Hence, 
$$f(Z_1)\otimes f(Z_2)g(Z_1,Z_2) + f(Z_2) \otimes f(Z_1)g(Z_1,Z_2) \preceq \left [ f(Z_1)^{\otimes 2} + f(Z_2)^{\otimes 2}\right ]g(Z_1,Z_2),$$
and
\begin{equation}
\mathbb{E} \left[ f(Z_1)\otimes f(Z_2)g(Z_1,Z_2) + f(Z_2) \otimes f(Z_1)g(Z_1,Z_2) \right] \preceq \mathbb{E} \left( \left [ f(Z_1)^{\otimes 2} + f(Z_2)^{\otimes 2}\right ]g(Z_1,Z_2) \right). \label{eq:Lemma2}
\end{equation}

The element $(i,j)$ of the matrix $\mathbb{E} \left[ f(Z_1)\otimes f(Z_2)g(Z_1,Z_2) \right]$ is $\mathbb{E} \left[ f(Z_1)_{i} f(Z_2)_{j}g(Z_1,Z_2) \right]$, where $f_{j}$ denotes the $j$th element of vector $f$ and $i,j \in \{1,\dots,d\}.$ Notice that
\begin{align*}
\mathbb{E} \left[ f(Z_1)_{i} f(Z_2)_{j}g(Z_1,Z_2) \right] =& \mathbb{E} \left[ f(Z_1)_{i} f(Z_2)_{j}g(Z_2,Z_1) \right] \\
=& \mathbb{E} \left[ f(Z_2)_{j}f(Z_1)_{i}g(Z_2,Z_1) \right]\\
=& \mathbb{E} \left[ f(Z_1)_{j}f(Z_2)_{i}g(Z_1,Z_2) \right].
\end{align*}
The first equation holds because $g$ is symmetric. The second equation is due to the commutative property of real value multiplication. The third equation is based on the fact that $Z_1, Z_2$ are i.i.d. samples. 

This implies that the element $(i,j)$ of the matrix $\mathbb{E} \left[ f(Z_1)\otimes f(Z_2)g(Z_1,Z_2) \right]$ is equal to its element $(j,i)$. Hence $\mathbb{E} \left[ f(Z_1)\otimes f(Z_2)g(Z_1,Z_2) \right]$ is symmetric and $$\mathbb{E} \left[ f(Z_1)\otimes f(Z_2)g(Z_1,Z_2) \right] = \mathbb{E} \left[ f(Z_2)\otimes f(Z_1)g(Z_1,Z_2) \right].$$ 
Therefore, equation (\ref{eq:Lemma2}) can be re-written as
$$\mathbb{E} \left[ 2f(Z_1)\otimes f(Z_2)g(Z_1,Z_2)\right] \preceq \mathbb{E} \left( \left [ f(Z_1)^{\otimes 2} + f(Z_2)^{\otimes 2}\right ]g(Z_1,Z_2) \right).$$
Moreover,
\begin{align*}
\mathbb{E}[f(Z_1)\otimes f(Z_2)g(Z_1,Z_2)] & \preceq \mathbb{E}[\frac{f(Z_1)^{\otimes 2}+f(Z_2)^{\otimes 2}}{2}g(Z_1,Z_2)] \\
& = \mathbb{E}[\frac{f(Z_2)^{\otimes 2}+f(Z_2)^{\otimes 2}}{2}g(Z_1,Z_2)] \\
& = \mathbb{E}[f(Z_2)^{\otimes 2}g(Z_1,Z_2)],
\end{align*}
and $\mathbb{E}[[f(Z_1)-f(Z_2)]\otimes f(Z_2)g(Z_1,Z_2)]\preceq 0$ follows immediately. The second equation holds because, by Lemma \ref{lemma: change variable}, we have $\mathbb{E}[g(Z_1,Z_2)\otimes f(Z_1)^{\otimes 2}] = \mathbb{E}[g(Z_1,Z_2)\otimes f(Z_2)^{\otimes 2}].$ This is equivalent to $\mathbb{E}[f(Z_1)^{\otimes 2}g(Z_1,Z_2)] = \mathbb{E}[f(Z_2)^{\otimes 2}g(Z_1,Z_2)]$, since $g$ is a real-valued function. 
\end{proof}
\end{lemma}

\subsection{Lemma 5}
\begin{lemma}
\label{lemma: convergence rate}
Let $\mathcal{B}_\delta=\{f-\mathbb{E}_X[f]:f\in\mathcal{F}, d(f_0,f-\mathbb{E}_X[f])\leq \delta\}$. Define $\mathbb{G}_n = \sqrt{n}(\mathbb{P}_n-\mathbb{P})$, then
\begin{equation}
\label{eq: Gn1}
\mathbb{E}^*\sup_{f\in\mathcal{B}_\delta}\left\lvert\mathbb{G}_n(\Tilde{g}^{(1)}(.;f)-\Tilde{g}^{(1)}(.;f_0))\right\rvert = O( \delta\sqrt{\varsigma\log \frac{U}{\delta}}+\frac{\varsigma}{\sqrt{n}}\log \frac{U}{\delta})
\end{equation}
\begin{equation}
\label{eq: Gn2}
\mathbb{E}^*\sup_{f\in \mathcal{F}(K,\varsigma,\mathbf{p},\mathcal{D})}\left\lvert\mathbb{G}_nf\right\rvert = O(\mathcal{D}\sqrt{\varsigma\log \frac{U}{\mathcal{D}}}+\frac{\varsigma \mathcal{D}}{\sqrt{n}}\log \frac{U}{\mathcal{D}}),
\end{equation}
where $\Tilde{g}^{(1)}(D_1;f) = \mathbb{E}[g(D_1,D_2,\dots,D_s;f)\lvert D_1]$, $\mathbb{E}^*$ is the outer measure and $U=K\prod_k^{K}(p_k+1)\sum_{k=0}^Kp_kp_{k+1}.$
\end{lemma}

\subsubsection{Proof of (\ref{eq: Gn1})}
\begin{proof}
Let $$\mathcal{G}_\delta=\{g(.;f):f\in\mathcal{F}, d(f_0,f-\mathbb{E}_X[f])\leq \delta\}$$ and 
$$\tilde{\mathcal{G}}_\delta=\{\Tilde{g}^{(1)}(.;f)-\Tilde{g}^{(1)}(.;f_0):f\in\mathcal{F}, d(f_0,f-\mathbb{E}_X[f])\leq \delta\}.$$

We first show that
\begin{equation}
    \label{eq: g tilt and g}
    \mathcal{N}_{[]}(\epsilon,\Tilde{\mathcal{G}}_\delta,L_2(P^s)) \lesssim \mathcal{N}_{[]}(\epsilon,{\mathcal{G}}_\delta,L_2(P^s))
\end{equation}

For any $f_1,f_2\in \mathcal{B}_\delta $ and let $D_1$ be a random sample,
\begin{align*}
& \mathbb{E}_{D_1}[ \Tilde{g}^{(1)}(D_1;f_1)-\Tilde{g}^{(1)}(D_1;f_0)- \{\Tilde{g}^{(1)}(D_1;f_2)-\Tilde{g}^{(1)}(D_1;f_0)\}]^2 \\
= & \mathbb{E}_{D_1}[ \Tilde{g}^{(1)}(D_1;f_1)- \Tilde{g}^{(1)}(D_1;f_2)]^2 \\
= &  \mathbb{E}_{D_1}[ \Tilde{g}^{(1)}(D_1;f_1)- \Tilde{g}^{(1)}(D_1;f_2)]^2 -\{\mathbb{E}[g(D(s);f_1)-g(D(s);f_2)]\}^2+\{\mathbb{E}[g(D(s);f_1)-g(D(s);f_2)]\}^2\\
= &  \mathbb{V}_{D_1}[ \Tilde{g}^{(1)}(D_1;f_1)- \Tilde{g}^{(1)}(D_1;f_2)]+\{\mathbb{E}_{D(s)}[g(D(s);f_1)-g(D(s);f_2)]\}^2\\
\leq & \mathbb{V}_{D(s)}[g(D(s);f_1)-g(D(s);f_2)]+\{\mathbb{E}_{D(s)}[g(D(s);f_1)-g(D(s);f_2)]\}^2\\
= & \mathbb{E}_{D(s)}[g(D(s);f_1)-g(D(s);f_2)]^2.
\end{align*}
Next, we show that 
\begin{equation}
\label{eq: dg and df}
\mathbb{E}_{D(s)}[g_1(D(s);f_1)-g_2(D(s);f_2)]^2\lesssim d^2(f_1,f_2).
\end{equation}
For any two $g_1,g_2 \in \mathcal{G}_\delta$ (indexed by $f_1$ and $f_2$, respectively), 
\begin{align*}
& \mathbb{E}[g(D(s);f_1)-g(D(s);f_2)]^2 \\ 
\leq & 2\mathbb{E}\left [\frac{1}{s}\sum_{j=1}^s\Delta_j\left \{ f_1(X_j)-f_2(X_j)\right \} \right ]^2+\\
&\quad 2\mathbb{E}\left [\frac{1}{s}\sum_{j=1}^s\Delta_j\left \{\log \sum_{k=1}^sY_k(T_j)\exp(f_1(X_k))-\log \sum_{k=1}^sY_k(T_j)\exp(f_2(X_k))\right\} \right ]^2\\
(1)\leq & 2\mathbb{E}\left [\frac{s}{s^2}\sum_{j=1}^s\Delta_j\left \{ f_1(X_j)-f_2(X_j)\right \}^2 \right ]+\\
&\quad 2\mathbb{E}\left [\frac{s}{s^2}\sum_{j=1}^s\Delta_j\left \{\log \sum_{k=1}^sY_k(T_j)\exp(f_1(X_k))-\log \sum_{k=1}^sY_k(T_j)\exp(f_2(X_k))\right\}^2 \right ]\\
(2)\lesssim &  \mathbb{E}\left [\left \{ f_1(X)-f_2(X)\right \}^2 \right ] + \mathbb{E}\left [\sum_{j=1}^s\left \{ \sum_{k=1}^sY_k(T_j)\exp(f_1(X_k))- \sum_{k=1}^sY_k(T_j)\exp(f_2(X_k))\right\}^2 \right ] \\
= &  \mathbb{E}\left [\left \{ f_1(X)-f_2(X)\right \}^2 \right ] + \mathbb{E}\left [\sum_{j=1}^s\left \{ \sum_{k=1}^sY_k(T_j)\left \{\exp(f_1(X_k))-\exp(f_2(X_k))\right \}\right\}^2 \right ]\\
(3)\leq &  \mathbb{E}\left [\left \{ f_1(X)-f_2(X)\right \}^2 \right ] + s\mathbb{E}\left [\sum_{j=1}^s\sum_{k=1}^sY_k(T_j)\left \{\exp(f_1(X_k))-\exp(f_2(X_k))\right\}^2 \right ] \\
(4)\lesssim &  \mathbb{E}\left [\left \{ f_1(X)-f_2(X)\right \}^2 \right ] + \mathbb{E}\left [\left \{\exp(f_1(X))-\exp(f_2(X))\right\}^2 \right ] \\
(5)\lesssim &  \mathbb{E}\left [\left \{ f_1(X)-f_2(X)\right \}^2 \right ] + \mathbb{E}\left [\left \{f_1(X)-f_2(X)\right\}^2 \right ] \\
\lesssim &  d^2(f_1,f_2).
\end{align*}
Inequalities (1) and (3) are due to the Cauchy-Schwarz inequality. Inequality (2) holds because $s$ is a finite number, and $\Delta$ only takes values 0 and 1, as well as the boundedness of $\sum_{k=1}^sY_k(T_j)\exp(f(X_k))$.
Inequality (4) holds because $s$ is a finite number, and $Y$ only takes values 0 and 1. Inequality (5) is due to the boundedness of $f$.

The relationships (\ref{eq: g tilt and g}) and (\ref{eq: dg and df}) imply $\mathcal{N}_{[]}(\epsilon,\Tilde{\mathcal{G}}_\delta,L_2(P^s)) \lesssim \mathcal{N}_{[]}(\epsilon,{\mathcal{G}}_\delta,L_2(P^s))\lesssim \mathcal{N}_{[]}(\epsilon,{\mathcal{B}}_\delta,\lVert. \rVert_{L^2})$. For any $f_1-\mathbb{E}_X[f_1],f_2-\mathbb{E}_X[f_2]\in \mathcal{B}_\delta$,
\begin{align*}
\lVert f_1-\mathbb{E}_X[f_1]-(f_2-\mathbb{E}_X[f_2] )\rVert_{L^\infty} & =\lVert f_1-f_2-(\mathbb{E}_X[f_1]-\mathbb{E}_X[f_2] )\rVert_{L^\infty}\\
& \leq \lVert f_1-f_2\rVert_{L^\infty}+\lVert\mathbb{E}_X[f_1]-\mathbb{E}_X[f_2] \rVert_{L^\infty} \\
& \lesssim  \lVert f_1-f_2\rVert_{L^\infty}.
\end{align*}
Hence, $\mathcal{N}_{[]}(\epsilon,{\mathcal{B}}_\delta,\lVert. \rVert_{L^\infty})\lesssim \mathcal{N}_{[]}(\epsilon,\mathcal{F},\lVert. \rVert_{L^\infty})$.
By the proof of Lemma 6 in \citet{zhong2022deep}, we have
$\log \mathcal{N}_{[]}(\epsilon,\mathcal{F},\lVert . \rVert_{L^\infty})\lesssim \varsigma\log \frac{U}{\epsilon}$ and therefore
$$\log \mathcal{N}_{[]}(\epsilon,\Tilde{\mathcal{G}}_\delta,L_2(P^s))\lesssim \varsigma\log \frac{U}{\epsilon}.$$
Moreover, we obtain
\begin{align*}
J_{[]}(\delta,\Tilde{\mathcal{G}}_\delta):&=\int_0^\delta\sqrt{1+\log \mathcal{N}_{[]}(\epsilon,\Tilde{\mathcal{G}}_\delta,L_2(P^s))}d\epsilon \\
& \lesssim \int_0^\delta\sqrt{\varsigma\log \frac{U}{\epsilon}}d\epsilon \\
(\text{Let }v=\sqrt{2\log\frac{U}{\epsilon}})& = \sqrt{\frac{\varsigma}{2}}U\int_{\sqrt{2\log\frac{U}{\delta}}}^\infty v^2\exp(-v^2/2)dv \\
(\text{Integration by part})&\asymp \delta\sqrt{\varsigma\log\frac{U}{\delta}}.
\end{align*}
The element of $\Tilde{\mathcal{G}}_\delta$ are uniformly bounded (e.g., $\lVert\cdot\rVert_\infty\lesssim M$) and for any $f_1-\mathbb{E}_X[f_1], f_2-\mathbb{E}_X[f_2] \in \mathcal{B}_\delta$, 
$$d(f_1-\mathbb{E}_X[f_1],f_2-\mathbb{E}_X[f_2])\leq d(f_1-\mathbb{E}_X[f_1],f_0)+d(f_0,f_2-\mathbb{E}_X[f_2]) \leq 2\delta.$$
This, (\ref{eq: g tilt and g}), and (\ref{eq: dg and df}) show that $\mathbb{E}[\Tilde{g}^{(1)}(.;f_1)-\Tilde{g}^{(1)}(.;f_2)]^2 \leq K\delta$ for any two $\Tilde{g}^{(1)}(.;f_1),\Tilde{g}^{(1)}(.;f_2) \in \Tilde{\mathcal{G}}_\delta$ with a universal constant $K$ so that the diameter of $\tilde{\mathcal{G}}_\delta$ is $K\delta$.

Then by Lemma 3.4.2 in \cite{van1996weak}, we have
\begin{align*}
\mathbb{E}^*\lVert \mathbb{G}_n\rVert_{\Tilde{\mathcal{G}}_\delta}&\lesssim J_{[]}(\delta,\Tilde{\mathcal{G}}_\delta)\left\{1+\frac{J_{[]}(\delta,\Tilde{\mathcal{G}}_\delta)}{\delta^2 \sqrt{n}}M\right\}\\
& \lesssim \delta\sqrt{\varsigma\log \frac{U}{\delta}}+\frac{\varsigma}{\sqrt{n}}\log \frac{U}{\delta}.
\end{align*}
\end{proof}

\subsubsection{Proof of (\ref{eq: Gn2})}
\begin{proof}
By the proof of Lemma 6 in \citet{zhong2022deep}, $\log \mathcal{N}_{[]}(\epsilon,\mathcal{F},L_2(P)) \lesssim \varsigma\log(\frac{U}{\epsilon})$. Therefore, the bracketing integral of $\mathcal{F}$ is
\begin{align*}
J_{[]}(\mathcal{D},\mathcal{F},L_2(P)):&=\int_0^\mathcal{D}\sqrt{1+\log \mathcal{N}_{[]}(\epsilon,\mathcal{F},L_2(P))}d\epsilon \\
& \lesssim \int_0^\mathcal{D}\sqrt{\varsigma\log \frac{U}{\epsilon}}d\epsilon \\
(\text{Let }v=\sqrt{2\log\frac{U}{\mathcal{D}}}) &= \sqrt{\frac{\varsigma}{2}}U\int_{\sqrt{2\log\frac{U}{\mathcal{D}}}}^\infty v^2\exp(-v^2/2)dv \\
(\text{Integration by part}) &\asymp \mathcal{D}\sqrt{\varsigma\log\frac{U}{\mathcal{D}}}.
\end{align*}
Because $\mathcal{F}$ is uniformly bounded by $\mathcal{D}$, then by Lemma 3.4.2 in \cite{van1996weak}, we have
\begin{align*}
\mathbb{E}^*\lVert \mathbb{G}_n\rVert_{\mathcal{F}}&\lesssim J_{[]}(\mathcal{D},\mathcal{F},L_2(P))\left\{1+\frac{J_{[]}(\mathcal{D},\mathcal{F},L_2(P))}{\mathcal{D}^2 \sqrt{n}}\mathcal{D}\right\}\\
& \lesssim \mathcal{D}\sqrt{\varsigma\log \frac{U}{\mathcal{D}}}+\frac{\varsigma \mathcal{D}}{\sqrt{n}}\log \frac{U}{\mathcal{D}}.
\end{align*}
\end{proof}

\subsection{Proof of Proposition 1}
\begin{proposition}\label{prop:trace and batch size}
Under the Cox model with the assumptions in Theorem \ref{thm: batch size and Hessian} hold,
$$ \text{Tr}(H_{2s})-\text{Tr}(H_{s}) = O(1/s),$$
where $H_s = \nabla^2_\theta\mathbb{E}[L_{Cox}^{(s)}(\theta)]|_{\theta=\theta_{0}}$ and Tr($\cdot$) is the trace. 
\end{proposition}
\begin{proof}
In the proof of Theorem 2, we have used Lemma \ref{lemma: final step} to show that
\begin{align*}
& \mathbb{E}[{\nabla^2_\theta}L_{Cox}^{(s)}(\theta)]|_{\theta=\theta_0}-\mathbb{E}[{\nabla^2_\theta}L_{Cox}^{(2s)}(\theta)]|_{\theta=\theta_0} \\
& = \int_0^\tau \mathbb{E}_{\mathcal{F}^{(2s)}_u}[\left( h_{\theta_0}(Z_2)-h_{\theta_0}(Z_1) \right) \otimes h_{\theta_0}(Z_1)g_{\theta_0}(Z_1,Z_2) ]\frac{\lambda_0(u)\mathrm{d}u}{s} \preceq 0,
\end{align*}
where $Z_1(u) = \{(Y_{s+1}(u),X_{s+1}),\dots,(Y_{2s}(u),X_{2s})\} , Z_2(u) = \{(Y_{1}(u),X_{1}),\dots,(Y_{s}(u),X_{s})\}$, 
$$ h_{\theta_0}(Z_1(u)) = \frac{S^{f(1)}_{\Delta} (u,\theta_0)}{S^{f(0)}_{\Delta} (u,\theta_0)},\ h_{\theta_0}(Z_2(u)) = \frac{S^{f(1)}_{s} (u,\theta_0)}{S^{f(0)}_{s} (u,\theta_0)}, $$
$$ g_{\theta_0}(Z_1(u),Z_2(u)) = \frac{S^{f(0)}_{\Delta} (u,\theta_0)S^{f(0)}_{s} (u,\theta_0)}{S^{f(0)}_{\Delta} (u,\theta_0)+S^{f(0)}_{s} (u,\theta_0)}, $$
and
\begin{align*}
S^{f(0)}_s (u,\theta) = &\sum_{j=1}^s Y_j(u)\exp\{f_{\theta}(X_j)\} \\
S^{f(1)}_s (u,\theta) = & \sum_{j=1}^s Y_j(u)\exp\{f_{\theta}(X_j)\}{\nabla_\theta} [ f_{\theta}(X_j)].
\end{align*}
This is equivalent to 
$$I_s : =\int_0^\tau \mathbb{E}_{\mathcal{F}^{(2s)}_u}[\left( h_{\theta_0}(Z_1)-h_{\theta_0}(Z_2) \right) \otimes h_{\theta_0}(Z_1)g_{\theta_0}(Z_1,Z_2) ]\frac{\lambda_0(u)\mathrm{d}u}{s} \succeq 0,$$
and we are going to show $Tr(I_s) = O(1/s)$ when the assumptions of Theorem \ref{thm: batch size and Hessian} hold. We will first show that
$$\int_0^\tau \mathbb{E}_{\mathcal{F}^{(2s)}_u}\left[\left( h_{\theta_0}(Z_1)-h_{\theta_0}(Z_2) \right) \otimes h_{\theta_0}(Z_1) \right]\lambda_0(u)\mathrm{d}u \succeq I_s,$$
and then demonstrate the trace of the matrix on the left-hand side is $O(1/s)$.

Notice that
\begin{equation}\label{ap-eq:trace1}
\frac{g_{\theta_0}(Z_1,Z_2)}{s} = \frac{1}{s}\frac{S^{f(0)}_{\Delta} (u,\theta_0)S^{f(0)}_{s} (u,\theta_0)}{S^{f(0)}_{\Delta} (u,\theta_0)+S^{f(0)}_{s} (u,\theta_0)} \leq \frac{1}{s}\frac{S^{f(0)}_{\Delta} (u,\theta_0)+S^{f(0)}_{s} (u,\theta_0)}{4}
\end{equation}
which follows from the inequality $\frac{2ab}{a+b}<\frac{a+b}{2}$. Under the assumptions of Theorem \ref{thm: batch size and Hessian}, the $s$ summands in both $S^{f(0)}_{\Delta}$ and $S^{f(0)}_{s}$ are bounded and $\frac{S^{f(0)}_{\Delta} (u,\theta_0)+S^{f(0)}_{s} (u,\theta_0)}{4s}$ is upper-bounded by a constant $C$ which does not depend on $s$.
Therefore,
\begin{align*}
I_s\preceq C\int_0^\tau \mathbb{E}_{\mathcal{F}^{(2s)}_u}\left[\left( \frac{S^{f(1)}_{\Delta} (u,\theta_0)}{S^{f(0)}_{\Delta} (u,\theta_0)}-\frac{S^{f(1)}_{s} (u,\theta_0)}{S^{f(0)}_{s} (u,\theta_0)} \right) \otimes \frac{S^{f(1)}_{\Delta} (u,\theta_0)}{S^{f(0)}_{\Delta} (u,\theta_0)} \right]\lambda_0(u)\mathrm{d}u,
\end{align*}

and we have
$$Tr(I_s) \lesssim Tr\left(\int_0^\tau \mathbb{E}_{\mathcal{F}^{(2s)}_u}\left[\left( \frac{S^{f(1)}_{\Delta} (u,\theta_0)}{S^{f(0)}_{\Delta} (u,\theta_0)}-\frac{S^{f(1)}_{s} (u,\theta_0)}{S^{f(0)}_{s} (u,\theta_0)} \right) \otimes \frac{S^{f(1)}_{\Delta} (u,\theta_0)}{S^{f(0)}_{\Delta} (u,\theta_0)} \right ]\lambda_0(u)\mathrm{d}u\right).$$

Moreover, because $\frac{S^{f(1)}_{\Delta} (u,\theta_0)}{S^{f(0)}_{\Delta} (u,\theta_0)}$ and $\frac{S^{f(1)}_{s} (u,\theta_0)}{S^{f(0)}_{s} (u,\theta_0)}$ are i.i.d. distributed,
\begin{align*}
& \mathbb{E}_{\mathcal{F}^{(2s)}_u}\left[\left( \frac{S^{f(1)}_{\Delta} (u,\theta_0)}{S^{f(0)}_{\Delta} (u,\theta_0)}-\frac{S^{f(1)}_{s} (u,\theta_0)}{S^{f(0)}_{s} (u,\theta_0)} \right) \otimes \frac{S^{f(1)}_{\Delta} (u,\theta_0)}{S^{f(0)}_{\Delta} (u,\theta_0)} \right]  = \mathbb{V}_{\mathcal{F}^{(2s)}_u}\left[\frac{S^{f(1)}_{\Delta} (u,\theta_0)}{S^{f(0)}_{\Delta} (u,\theta_0)} \right] \\
& \preceq \mathbb{E}_{\mathcal{F}^{(2s)}_u}\left[\left\{\frac{S^{f(1)}_{\Delta} (u,\theta_0)}{S^{f(0)}_{\Delta} (u,\theta_0)}\right\}^{\otimes 2}\right]= \mathbb{E}\left[\left\{\frac{S^{f(1)}_{s} (u,\theta_0)}{S^{f(0)}_{s} (u,\theta_0)}\right\}^{\otimes 2}\right].
\end{align*}

This shows that
\begin{align*}
Tr(I_s) & \lesssim Tr\left(\int_0^\tau \mathbb{E}\left[\left\{\frac{S^{f(1)}_{s} (u,\theta_0)}{S^{f(0)}_{s} (u,\theta_0)}\right\}^{\otimes 2}\right]\lambda_0(u)\mathrm{d}u\right) \\
& = \sum_{i=1}^p \left(\int_0^\tau \mathbb{E}\left[\left\{\frac{S^{f(1)}_{s} (u,\theta_0)}{S^{f(0)}_{s} (u,\theta_0)}\right\}^{\otimes 2}\right]\lambda_0(u)\mathrm{d}u\right)_{i,i} \\
& \lesssim \left(\int_0^\tau \mathbb{E}\left[\left\{\frac{S^{f(1)}_{s} (u,\theta_0)}{S^{f(0)}_{s} (u,\theta_0)}\right\}^{\otimes 2}\right]\lambda_0(u)\mathrm{d}u\right)_{1,1} \\
& = \int_0^\tau \mathbb{E}\left[\left\{\frac{S^{f(1)}_{s} (u,\theta_0)}{S^{f(0)}_{s} (u,\theta_0)}\right\}^{\otimes 2}_{1,1}\right]\lambda_0(u)\mathrm{d}u,
\end{align*}
where $(\cdot)_{i,j}$ denotes the $(i,j)$ element of the matrix.

Next, we will show that for $u\in(0,\tau)$,
$$ \mathbb{E}\left[\left\{\frac{S^{f(1)}_{s} (u,\theta_0)}{S^{f(0)}_{s} (u,\theta_0)}\right\}^{\otimes 2}_{1,1}\right] = \mathbb{E}\left[\left\{\frac{\sum_{j=1}^s Y_j(u)\exp\{f_{\theta}(X_j)\}  {\nabla_\theta}f_{\theta}(X_j)_1}{\sum_{j=1}^s Y_j(u)\exp\{f_{\theta}(X_j)\}}\right\}^2\right] = O(1/s),$$
where $ {\nabla_\theta}f_{\theta}(X_j)_1$ is the first element of $ {\nabla_\theta}f_{\theta}(X_j)$. Notice that both the numerator and the denominator would be zero when $Y^{(s)}(u) = 0$. This is the case where all subjects are censored at time $u$ and the expectation is over an empty filtration $\mathcal{F}^{(s)}_u$, which we manually defined the value as 0. Hence
\begin{align*}
& \mathbb{E}\left[\left\{\frac{\sum_{j=1}^s Y_j(u)\exp\{f_{\theta}(X_j)\}  {\nabla_\theta}f_{\theta}(X_j)_1}{\sum_{j=1}^s Y_j(u)\exp\{f_{\theta}(X_j)\}}\right\}^2\right] \\
&= \mathbb{E}\left[\mathbb{E}\left[\left . \left\{\frac{\sum_{j=1}^s Y_j(u)\exp\{f_{\theta}(X_j)\}  {\nabla_\theta}f_{\theta}(X_j)_1}{\sum_{j=1}^s Y_j(u)\exp\{f_{\theta}(X_j)\}}\right\}^2\right |Y^{(s)}(u)\right]\right] \\
&= \sum_{k=1}^s\mathbb{E}\left[\left . \left\{\frac{\sum_{j=1}^s Y_j(u)\exp\{f_{\theta}(X_j)\}  {\nabla_\theta}f_{\theta}(X_j)_1}{\sum_{j=1}^s Y_j(u)\exp\{f_{\theta}(X_j)\}}\right\}^2\right |Y^{(s)}(u) = k\right]\mathbb{P}(Y^{(s)}(u) = k) \\
&= \sum_{k=1}^s\mathbb{E}\left[\left . \left\{\frac{\sum_{j=1}^k \exp\{f_{\theta}(X_j)\}  {\nabla_\theta}f_{\theta}(X_j)_1}{\sum_{j=1}^k \exp\{f_{\theta}(X_j)\}}\right\}^2\right |\sum_{j=1}^kY_j(u) = k\right]\mathbb{P}(Y^{(s)}(u) = k) \\
(1) & \lesssim \sum_{k=1}^s\mathbb{E}\left[\left . \left\{\frac{\sum_{j=1}^k   {\nabla_\theta}f_{\theta}(X_j)_1}{k}\right\}^2\right |\sum_{j=1}^kY_j(u) = k\right]\mathbb{P}(Y^{(s)}(u) = k) \\
& \lesssim \sum_{k=1}^s \frac{\mathbb{P}(Y^{(s)}(u) = k)}{k} \\
& \leq 2\mathbb{P}(Y^{(s)}(u) = 0)+\sum_{k=1}^s \frac{\mathbb{P}(Y^{(s)}(u) = k)}{k} \leq 2\mathbb{P}(Y^{(s)}(u) = 0)+\sum_{k=1}^s \frac{2\mathbb{P}(Y^{(s)}(u) = k)}{k+1} \\
& = 2\sum_{k=0}^s \binom{s}{k}\frac{p_u^k(1-p_u)^{s-k}}{k+1} \\
& = \frac{2}{(s+1)p_u}\sum_{k=0}^s \binom{s+1}{k+1}\frac{p_u^{k+1}(1-p_u)^{s-k}}{k+1} = 2\frac{1-(1-p_u)^{s+1}}{(s+1)p_u} \lesssim 1/s,
\end{align*}
where $p_u := \mathbb{P}(Y_i(u) = 1)>0$ by assumptions (A1)-(A2) and inequality $(1)$ is due to the boundedness of $\nabla_\theta f$ over $X$ in assumption. This completes the proof. 
\end{proof}

\section{Linear Scaling Rule}
\subsection{Numerical Experiment: Linear Scaling Rule in NN training}
\begin{figure}[htbp]
\centering
\includegraphics[width=1\textwidth]{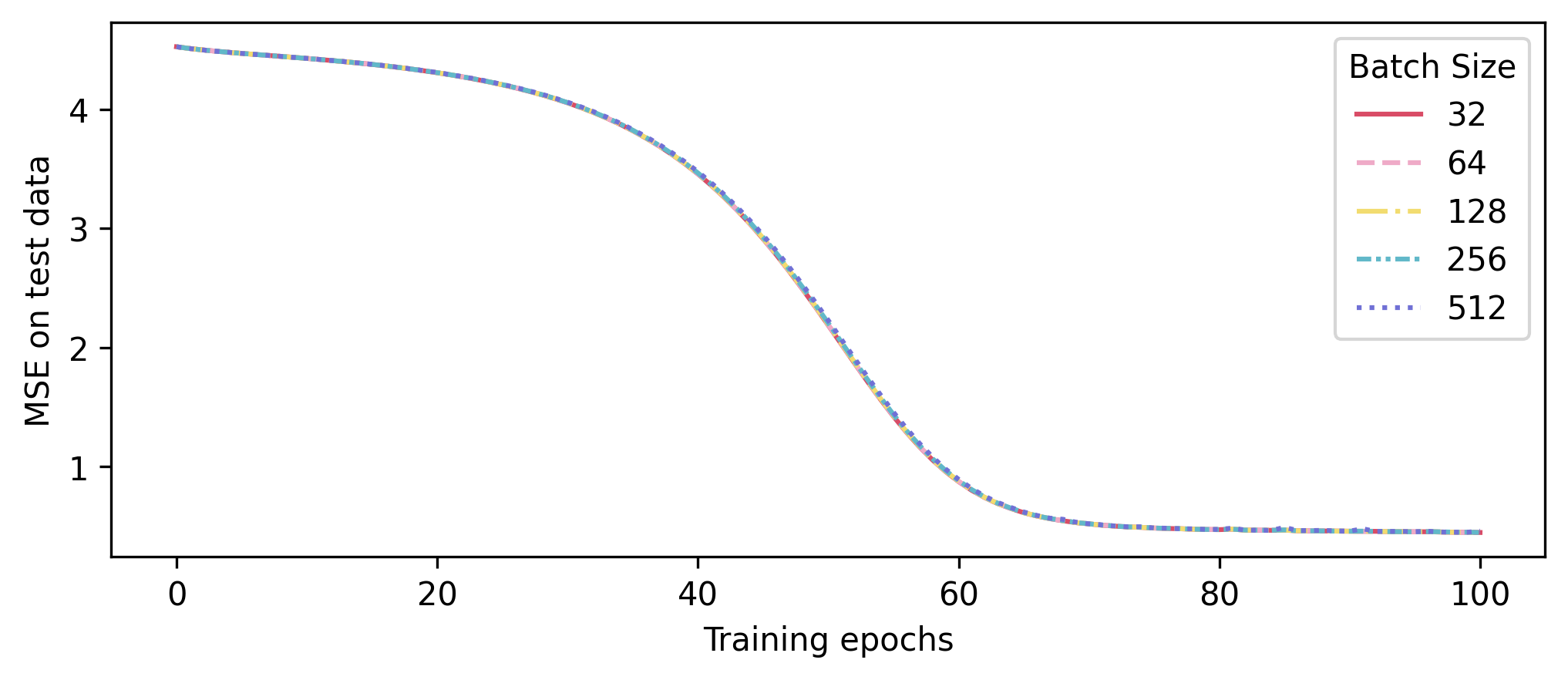}
\caption{Linear Scaling Rule in NN training. The Mean Squared Error (MSE) over the training epochs are perfectly matched under the linear scaling rule when optimizing a neural network for a standard regression problem. The MSE is calculated on separate test data ($n= 2,048$). The learning rate $\gamma$ is 0.001/16 when the batch size is 32. $\gamma$ is doubled when doubling the batch size, and it reaches 0.001 when the batch size is 512 ($=32\times 16$). All the other hyperparameters are kept the same.}
\label{fig:NN-with-LS}
\end{figure}

Considering a nonparametric regression problem where 
\begin{equation*}
y = f_0(X) = x_1^2x_2^3+\log(x_3+1)+\sqrt{x_4x_5+1}+\exp(x_5/2)-8.6,\ x_i\sim U(0,1).
\end{equation*} The observed data is $(y_i,x_{i1},\dots,,x_{i5})$ and we use a neural network to approximate $f_0$ with mean squared error as loss. Fig. (\ref{fig:NN-with-LS}) presents the loss curves on separate test data when adjusting the learning rate by the linear scaling rule. Specifically, the learning rate $\gamma$ is 0.001/16 when the batch size is 32. $\gamma$ is doubled when doubling the batch size, and it reaches 0.001 when the batch size is 512 ($=32\times 16$). The fully connected neural network consists of one hidden layer with 16 nodes and uses ReLU as an activation function. Neither the dropout layer nor the batch-normalization layer is used. The initialization of the neural network parameters is kept the same throughout the simulations.

The curves are perfectly matched when applying the scaling rule. Identical to the spirit of \citet{goyal2017accurate}, the goal is to match the test errors across minibatch sizes by using a general strategy that avoids hyperparameter tuning for each minibatch size, and even better results for any specific minibatch size could be obtained by optimizing
hyperparameters for that case.

\subsection{Review of existing result}
For completeness, we first present the mathematical sketch of SGD dynamic by \citet{jastrzkebski2017three} where the loss function is the average of i.i.d. individual loss. That is $L(\theta) = \frac{1}{n}\sum_{i=1}^n L_i(\theta)$ and the SGD step is
\begin{equation}\label{ap-eq:sgd dynamics 1}
\theta_{t+1} = \theta_t-\eta \nabla_\theta L^{(s)}(\theta_t),
\end{equation}
where $L^{(s)}(\theta) = \frac{1}{s}\sum_{i:D_i\in D(s)} L_i(\theta)$ and $D(s)$ is a randomly generated minibatch of size $s<n$.

The SGD (\ref{ap-eq:sgd dynamics 1}) can be written as
\begin{equation}\label{ap-eq:sgd dynamics 2}
\theta_{t+1} = \theta_t-\eta \nabla_\theta L(\theta_t)+\eta (\nabla_\theta L(\theta_t)-\nabla_\theta L^{(s)}(\theta_t)).
\end{equation}
For large $s$, $ (\nabla_\theta L(\theta_t)-\nabla_\theta L^{(s)}(\theta_t))$ is close to a zero mean Gaussian random noise with variance $\Sigma(\theta) = \frac{1}{s}C(\theta)$ where $C(\theta):=\mathbb{V}[\nabla_\theta L_i(\theta)]$. Hence (\ref{ap-eq:sgd dynamics 2}) can be re-written as
\begin{equation}\label{ap-eq:sgd dynamics 3}
\theta_{t+1} = \theta_t-\eta \nabla_\theta L(\theta_t)+\frac{\eta}{\sqrt{s}}\epsilon,
\end{equation}
where $\epsilon$ is a zero mean Gaussian random variable with covariance $C(\theta)$. This is a discretization of the stochastic differential equation (SDE) of the form
\begin{equation}\label{ap-eq:sgd dynamics 4}
d\theta = -\nabla_\theta L(\theta)dt+\sqrt{\frac{\eta}{s}}R(\theta)dW(t),
\end{equation}
where $R(\theta)^TR(\theta) = C(\theta).$ Let $H:= \nabla^2_\theta L(\theta)$ and consider the eigendecomposition $H = V\Lambda V^T$ with $\Lambda$ being the diagonal matrix of positive eigenvalues and $V$ an orthonormal matrix. With the assumption that $H=C$, the (\ref{ap-eq:sgd dynamics 4}) can be parameterized in terms of a new variable $z$ defined by $z:=V^T(\theta-\theta_0)$ where $\theta_0$ are the parameters at the minimum.

Specifically, starting from the SDE (\ref{ap-eq:sgd dynamics 4}), and making the quadratic approximation of the loss $L(\theta)\approx \frac{1}{2}(\theta-\theta_0)^TH(\theta-\theta_0)$ and the change of variables, results in an Ornstein-Unhlenbeck (OU) process for $z$
\begin{equation}\label{ap-eq:sgd dynamics 5}
\begin{split}
dz & = V^Td\theta \\
& = V^T\left\{-\nabla_\theta L(\theta)dt+\sqrt{\frac{\eta}{s}}R(\theta)dW(t)\right\} \\
& = V^T\left\{-V\Lambda V^T(\theta-\theta_0)dt+\sqrt{\frac{\eta}{s}}V\Lambda^{1/2}dW(t)\right\} \\
& = -\Lambda zdt+\sqrt{\frac{\eta}{s}}\Lambda^{1/2}dW(t).
\end{split}
\end{equation}

The quantity $\sqrt{\frac{\eta}{s}}\Lambda^{1/2}$ in the last term shows how learning rate, batch size, and Hessian (aka gradient covariance) influence the SGD dynamics. Moreover, in terms of the new parameters $z$, the expected loss can be written as
\begin{equation}\label{ap-eq:sgd dynamics 6}
\mathbb{E}(L) = \frac{1}{2}\sum_{i=1}^p \lambda_i \mathbb{E}(z^2_i) = \frac{\eta}{4S}\text{Tr}(\Lambda) = \frac{\eta}{4S}\text{Tr}(H),
\end{equation}
where the expectation is over the stationary distribution of the OU process, and the second equality follows from the expression for the OU covariance.

The relationship (\ref{ap-eq:sgd dynamics 6}) shows that the learning rate to batch size ratio determines the trade-off between width and expected loss associated with SGD dynamics within a minimum. A wider minima often leads to better generalization. Following this work, \citet{xie2020diffusion} further shows that SGD favors flat minima exponentially more than sharp minima, demonstrating the powerfulness of SGD in NN training.

\subsection{Linear scaling rule in Cox-NN training}
Two assumptions used in \citet{jastrzkebski2017three} do not hold for Cox models: (1) The loss of Cox models can not be written as the average of i.i.d. individual loss, and (2) the Hessian does not equal the covariance of gradient as depicted by Theorem \ref{thm: batch size and Hessian}. However, we note that we can follow the idea of \citet{jastrzkebski2017three} and derive a similar result for Cox-NN based on the relationship (\ref{eq: Hessian and covariance}).

Specifically, \citet{jastrzkebski2017three} uses the following relationships on the neighbor of $\theta_0$:
$$ \mathbb{V}[\nabla_\theta L^{(s)}(\theta_t)-\nabla_\theta L(\theta_t)] \approx  \frac{1}{s}\mathbb{V}[\nabla_\theta L_i(\theta)] = \frac{1}{s} \nabla^2_\theta L(\theta) \approx \frac{1}{s} \mathbb{E}[\nabla^2_\theta L^{(s)}(\theta_t)],$$
where the first and the last approximations are due to the large sample size, and the equation in the middle holds by the assumption in \citet{jastrzkebski2017three} that $H=C$. The key is that the variance of random noise $\mathbb{V}[\nabla_\theta L^{(s)}(\theta_t)-\nabla_\theta L(\theta_t)]$ is approximately the Hessian $\mathbb{E}[\nabla^2_\theta L^{(s)}(\theta_t)]$ divided by $s$, which holds at $\theta_0$ for the loss function of the Cox model given by (\ref{eq: Hessian and covariance}). Hence, even though the intermediate equivalences fail for Cox models, a similar result can be obtained because the variance of random noise is approximately the Hessian divided by $s$.

Specifically, the target function of SGD for Cox model is $$ L^{SGD}_{Cox}(\theta):= \frac{1}{\binom{n}{s}}\sum_{D(s)\subset D(n)} L_{Cox}^{(s)}(\theta)\approx \mathbb{E}[L_{Cox}^{(s)}(\theta)] $$
and the SGD step is
\begin{equation}\label{ap-eq:sgd dynamics 7}
\theta_{t+1} = \theta_t-\eta \nabla_\theta L_{Cox}^{(s)}(\theta_t).
\end{equation}
It can be written as
\begin{equation}\label{ap-eq:sgd dynamics 8}
\theta_{t+1} = \theta_t-\eta \nabla_\theta L^{SGD}_{Cox}(\theta)+\eta (\nabla_\theta L^{SGD}_{Cox}(\theta)-\nabla_\theta L_{Cox}^{(s)}(\theta_t)).
\end{equation}
For large $s$, by martingale central limit theorem $(\nabla_\theta L^{SGD}_{Cox}(\theta)-\nabla_\theta L_{Cox}^{(s)}(\theta_t))$ is close to a zero mean Gaussian random noise with variance $\mathbb{V}[\nabla_\theta L_{Cox}^{(s)}(\theta)]$.

Hence (\ref{ap-eq:sgd dynamics 8}) can be re-written as
\begin{equation}\label{ap-eq:sgd dynamics 9}
\theta_{t+1} = \theta_t-\eta \nabla_\theta  L^{SGD}_{Cox}(\theta)+\frac{\eta}{\sqrt{s}}\epsilon,
\end{equation}
where $\epsilon$ is a zero mean Gaussian random variable with variance $s\mathbb{V}[\nabla_\theta L_{Cox}^{(s)}(\theta)]$. Note that $s\mathbb{V}[\nabla_\theta L_{Cox}^{(s)}(\theta)] = \mathbb{E}[\nabla^2_\theta L^{(s)}_{Cox}(\theta)]$ at $\theta_0$ by equation (\ref{eq: Hessian and covariance}) and $\mathbb{E}[\nabla^2_\theta L^{(s)}_{Cox}(\theta)]$ is as approximately invariant to $s$ for large $s$. Therefore, we expect the variance $s\mathbb{V}[\nabla_\theta L_{Cox}^{(s)}(\theta)]$ depends on $s$ but this dependency shall be mild when $s$ is large. The iteration (\ref{ap-eq:sgd dynamics 9}) is a discretization of the stochastic differential equation (SDE) of the form
\begin{equation}\label{ap-eq:sgd dynamics 10}
d\theta = -\nabla_\theta  L^{SGD}_{Cox}(\theta)dt+\sqrt{\frac{\eta}{s}}R_s(\theta)dW(t),
\end{equation}
where $R_s(\theta)^TR_s(\theta) = s\mathbb{V}[\nabla_\theta L_{Cox}^{(s)}(\theta)].$ Consider the eigendecomposition $\nabla^2_\theta L^{SGD}_{Cox}(\theta) = V_s\Lambda_s V_s^T$ with $\Lambda_s$ being the diagonal matrix of positive eigenvalues and $V_s$ an orthonormal matrix. By relationship given in (\ref{eq: Hessian and covariance}), we have $\nabla^2_\theta L^{SGD}_{Cox}(\theta)\approx s\mathbb{V}[\nabla_\theta L_{Cox}^{(s)}(\theta)]$ around $\theta_0$, the (\ref{ap-eq:sgd dynamics 10}) can be parameterized in terms of a new variable $z$ defined by $z:=V_s^T(\theta-\theta_0)$ where $\theta_0$ are the parameters at the minimum.

Specifically, starting from the SDE (\ref{ap-eq:sgd dynamics 10}), and making the quadratic approximation of the loss $L^{SGD}_{Cox}(\theta)\approx \frac{1}{2}(\theta-\theta_0)^T\nabla^2_\theta L^{SGD}_{Cox}(\theta)(\theta-\theta_0)$ and the change of variables, results in an Ornstein-Unhlenbeck (OU) process for $z$
\begin{equation}\label{ap-eq:sgd dynamics 11}
\begin{split}
dz & = V_s^Td\theta \\
& = V_s^T\left\{-\nabla_\theta L^{SGD}_{Cox}(\theta)dt+\sqrt{\frac{\eta}{s}}R_s(\theta)dW(t)\right\} \\
& = V_s^T\left\{-V_s\Lambda_s V_s^T(\theta-\theta_0)dt+\sqrt{\frac{\eta}{s}}V_s\Lambda_s^{1/2}dW(t)\right\} \\
& = -\Lambda_s zdt+\sqrt{\frac{\eta}{s}}\Lambda_s^{1/2}dW(t),
\end{split}
\end{equation}
which is the same as the result (\ref{ap-eq:sgd dynamics 5}) except that $\Lambda_s$ depends on $s$. This demonstrates that three factors: learning rate, batch size and the Hessian (or the gradient covariance) influence the SGD performance in Cox-NN. When $s$ is large, the change of $\text{Tr}(H_s)$ would be small (see Proposition \ref{prop:trace and batch size}). This explains why the linear scaling rule holds in Cox-NN training, especially when the batch size is large. Finally
\begin{equation}\label{ap-eq:sgd dynamics 12}
\mathbb{E}(L^{SGD}_{Cox}) = \frac{1}{2}\sum_{i=1}^p \lambda_i \mathbb{E}(z^2_i) = \frac{\eta}{4S}\text{Tr}(\Lambda_s) = \frac{\eta}{4S}\text{Tr}(H_s).
\end{equation}
This gives a theoretical relationship between the expected loss value, the level of stochastic noise in SGD ($\frac{\eta}{S}$), and the Hessian trace at this final stage of training.

Finally, we provide additional simulation details for the Cox-NN linear scaling rule discussed in the main text: Cox-NN uses the fully connected layers consisting of one hidden layer with 16 nodes and uses ReLU as an activation function. Neither the dropout layer nor the batch-normalization layer is used. The initialization of the neural network parameters is kept the same throughout the simulations.

\section{Details of Numerical Implementations}

\subsection{Simulations}

All simulations are implemented in Python. The implementation of Cox-NN and the training use Pytorch.

\subsubsection{Cox-NN Simulation\label{sec:Cox-NN Simulation}}
We conducted the simulations to evaluate the performance of the mb-MPLE (estimated by SGD) for Cox-NN. The true event time $T^*_i$ is generated from a distribution with true hazard function $\lambda(t|X_i) = \lambda_0(t)\exp(f_0(X_i))$ where $\lambda_0(t) = 1,X\in \mathbb{R}^5$ and 
\begin{equation*}
f_0(X) = x_1^2x_2^3+\log(x_3+1)+\sqrt{x_4x_5+1}+\exp(x_5/2)-8.6,\ x_i\sim U(0,1), 
\end{equation*} where $-8.6$ is to center $\mathbb{E}[f_0(X)]$ towards zero.  The observed data is $(X_i,\min(T^*_i,C^*_i),I(T^*_i<C^*_i))$ where $C^*_i$ is generated from an independent  exponential distribution. We adjust the parameter of the exponential distribution to make the censoring rate $30\%$.

To tune the hyperparameters of the Cox-NN, we held out a validation set (20\%) from the training set, and the training is early stopped once the C-index on the validation set does not improve for ten consecutive epochs. Therefore, the best validation C-index is at the tenth epoch from the end, and the corresponding Cox-NN is the output model. Among the different configurations listed in Table \ref{tb:simulation configurations}, we chose the hyperparameters that maximized the C-index on the validation set for our final model, which is further evaluated on a separate test set ($N_{test}=10,000$). We repeated this process 50 times (50 runs) with different training sample sizes $N_{train}$ and reported the mean and standard deviation of the metrics.

\begin{table}[htp]
\centering
\begin{threeparttable}[b]
\caption{Cox-NN simulation configurations}
\begin{tabular}{ccc}
\hline
\multirow{4}{*}{Training Data} & train/validation split  & 0.8/0.2             \\
                                   & censoring rate     & 0.3           \\
                                   & simulation runs     & 50           \\
\hline
\multirow{3}{*}{Cox-NN parameters} & Number of layer$^*$  & \{1,2,4\}             \\
                                   & Number of neuron & \{16,32,64\}          \\
                                   & Dropout rate     & \{0.1,0.3\}           \\
\hline
\multirow{4}{*}{SGD parameters}    & Learning rate    & \{0.1,0.01,0.001\} \\
                                   & Maximum Epoch   & 200         \\
                                   & Early Stopping Patience   & 10         \\
                                   & Early Stopping Metric   & validation C-index\\
\hline
\end{tabular}
\label{tb:simulation configurations}
\begin{tablenotes}
\item [$*$] Each layer consists of the following structures: fully connected layer $\to$ batch normalization layer $\to$ activation function (ReLU) $\to$ dropout layer.
\end{tablenotes}
\end{threeparttable}
\end{table}

We evaluate the performance of estimates $\hat{f}$ using the root mean square error (RMSE)
$$\text{RMSE}(\hat{f}):=\left[\frac{1}{N_{test}}\sum_{i=1}^{N_{test}}\{(\hat{f}(X_i)-\Bar{\hat{f}})-f_0(X_i)\}^2\right]^{1/2}$$

where $\hat{f}$ and $f_0$ are evaluated on the covariates of the test set $\{X_i,i=1,\dots,n_{test}\}$ and $\Bar{\hat{f}} = \frac{1}{N_{train}}\sum_{i=1}^{N_{train}}\hat{f}(X_i)$ is the empirical average of $\hat{f}$ based on the training data. Besides the RMSE, we also evaluated $\hat{f}$ through the concordance index (C-index) by \citet{harrell1982evaluating}, which is a popular method to measure the predictive performance in deep learning. The C-index measures the concordance between the ranks of the predicted survival times and the observed ranks.

Table \ref{tb:Cox-NN simulation} presents the Cox-NN simulation result from 50 runs. The mb-MPLE found by SGD for Cox-NN consistently outperformed Cox-reg (Cox regression model fitted by standard software) in both metrics. Moreover, with increasing training samples, the RMSE of the mb-MPLE decreases for both batch sizes 32 and 128. This numerically verifies the Theorem \ref{thm:convergence rate} that the mb-MPLE of Cox-NN $\hat{f}^{(s)}$ converges to the true function $f_0$, and this holds for different choices of $s$.

\begin{table}[htp]
\centering
\caption{The RMSE and C-index of the Cox-NN, optimized using SGD (with batch sizes of 32 and 128), are reported as the mean (standard deviation) across 50 simulation runs. In each run, we chose the hyperparameters that maximized the C-index on the validation set for our final model and evaluated the RMSE$(\hat{f})$ and C-index on the test set ($N_{test}=10,000$).}
\begin{tabular}{c|c|cc|c}
\hline
                         & \multirow{2}{*}{$N_{train}$} & \multicolumn{2}{c|}{Cox-NN}            & \multirow{2}{*}{Cox-reg} \\
\cline{3-4}
                         &                    & s=32    & s=128   &                        \\
\hline
\multirow{3}{*}{RMSE$(\hat{f}_n^{(s)})$}     & 512  & 0.668(0.156) & 0.664(0.200) & 0.705(0.028) \\
                              & 2048 & 0.575(0.208) & 0.455(0.150) & 0.697(0.017) \\
                              & 8192 & 0.369(0.092) & 0.324(0.114) & 0.694(0.012) \\
\hline
\multirow{3}{*}{C-index} & 512  & 0.837(0.004) & 0.836(0.006) & 0.827(0.002) \\
                              & 2048 & 0.843(0.002) & 0.843(0.002) & 0.828(0.002) \\
                              & 8192 & 0.845(0.002) & 0.845(0.002) & 0.828(0.002)  \\
\hline
\end{tabular}
\label{tb:Cox-NN simulation}
\end{table}

\subsubsection{SGD for offline Cox regression}
We carried out simulations with 200 runs to empirically assess the impact of batch size in Cox regression. The true event time $T^*_i$ is generated from $\lambda(t|X_i) = \lambda_0(t)\exp(X_i^T\theta_0)$ where $\lambda_0(t) = 1$, $\theta_0 = \textbf{1}_{10\times 1}$, and $X_{ip}\displaystyle\operatorname*{\sim}^{i.i.d.}$ Uniform(0,1) for $p \in \{1,2,\dots,10\}$. The true censoring time $C^*_i$ is generated from an independent exponential distribution with a censoring rate $30\%$. 

We performed projected SGD with $B=10^{6}$ (the threshold for projection) and initial point $\hat\theta^{(0)} = \textbf{0}_{10\times 1}$ to estimate $\theta_0$ based on $n = 2,048$ samples. The SGD batch size is set as $2^k$ where $k=2,\dots,9$. The total number of epochs (train the model with all the training data for one cycle) is $200$ and the learning rate is set as $\gamma_E = \frac{2^{k-5}}{E+1}$ at epoch $E$, which is proportional to the choice of batch size and decreases after each epoch. Besides SGD-SB and SGD-FB, we also fit a stratified Cox model (CoxPH-strata) by treating the fixed batches from the FB strategy as strata. Note that CoxPH-strata directly solves
\begin{equation*}
\tilde{\theta}^{FB(s)}_{n} = \arg\min_{\theta\in\mathbb{R}_M^{p}} \frac{1}{m}\sum_{D(s)\in D(n|s)} L_{Cox}^{(s)}(\theta),
\end{equation*} 
and gives $\Tilde\theta^{FB(s)}_{n}$ from GD algorithm. The convergence of SGD to $\displaystyle\arg\min_\theta \mathbb{E}[L^{(s)}_{Cox}(\theta)|D(n)]$ can be evaluated by comparing the estimators from SGD-FB and CoxPH-strata. CoxPH is fitted under the standard function using all samples to obtain the MPLE. The simulation is repeated for 200 runs. 

We note that $\hat\theta^{(t)}$ stays inside of $\mathbb{R}^p_B$ throughout the iterations. Hence, the projection step is redundant in the simulation, and the projected SGD degenerates to the standard SGD. The convergence of SGD is validated by the negligible difference between the output from SGD $\hat\theta_t^{FB(s)}$ and the estimator from CoxPH-strata $\hat\theta^{strata}$ where $\log(\lVert \hat\theta^{FB(s)}-\hat\theta^{strata} \rVert_2^2)<-7.5$ for all $s$ throughout the simulations.

\begin{figure}[htbp]
\includegraphics[width=1\textwidth]{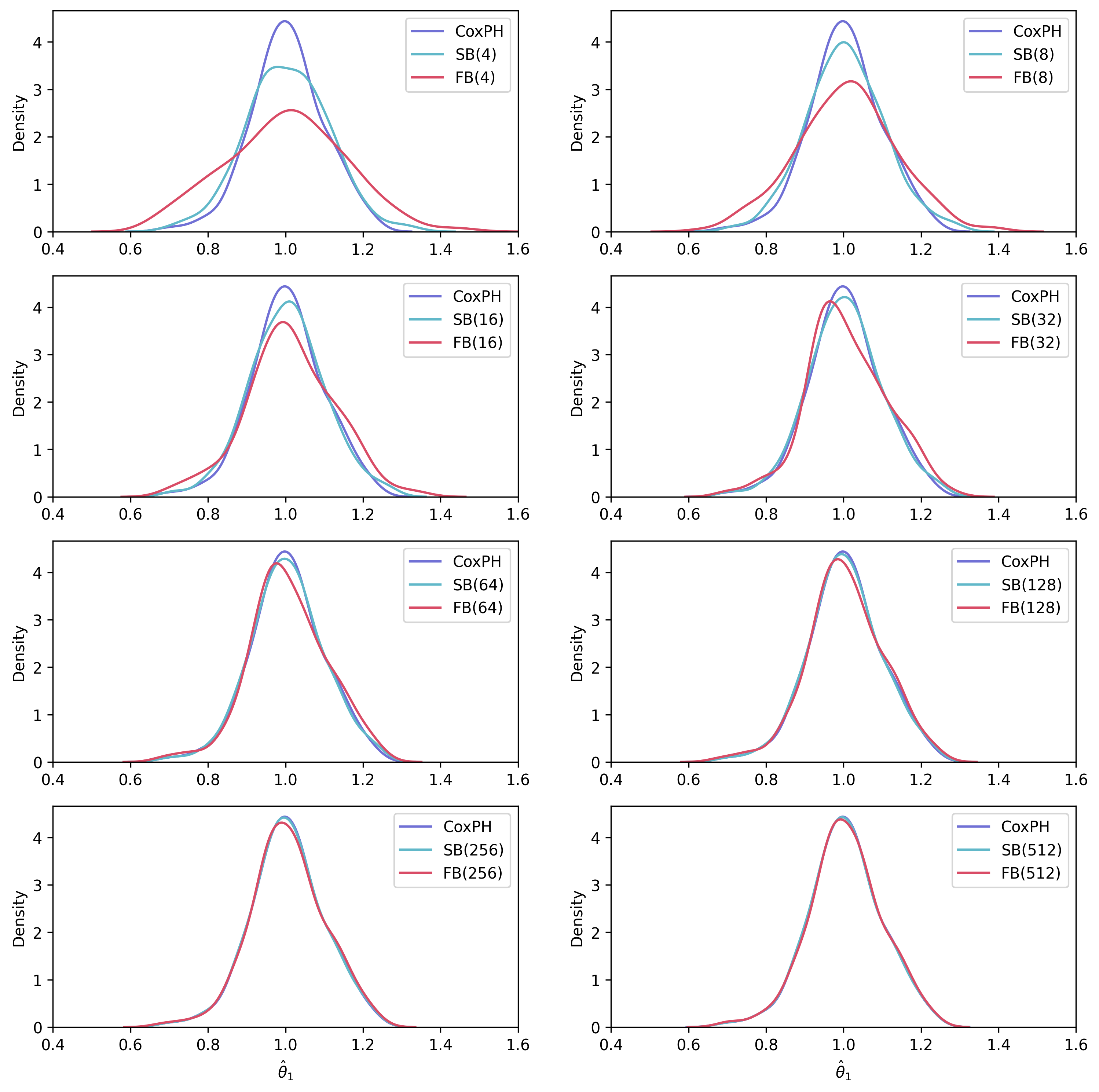}
\caption{The kernel density estimation of the $\hat{\theta}_1$ distribution is based on 200 simulation runs (with truth $\theta_1 = 1$ and $n = 2,048$). Each subfigure corresponds to a different batch size (indicated in brackets) used in SGD. The SGD algorithm was implemented with either a stochastic batch (SB) strategy or a fixed batch (FB) strategy. For reference, the MPLE obtained using standard software (CoxPH), which is independent of batch size, is also included in the figure.} 
\label{fig:PH-reg-simulation-theta1}
\end{figure}

We investigated the asymptotic normality of the mb-MPLE. Figure \ref{fig:PH-reg-simulation-theta1} displays the kernel density distribution of $\hat{\theta}_1$ (the estimator of $\theta_{0,1}$) based on 200 simulation runs with different choices of batch size. As a reference, the MPLE obtained using standard software (CoxPH), which is independent of batch size, is also included in the figure. We observe that the density of $\hat{\theta}^{SB(s)}_1$ and $\hat{\theta}^{FB(s)}_1$ both exhibit a normal shape for all different values of $s$. Moreover, the density gets closer to that of MPLE as $s$ increases.

The bias (Bias) and standard deviation (Std) of the mb-MPLE for $\theta_{0,1}$ from 200 simulation runs are displayed in Table \ref{tb: simulation theta1}. We found that the bias is close to zero, and the empirical variance of the estimated coefficients decreases as the batch size increases. Moreover, across all $s$, the empirical variance of $\hat{\theta}_n^{SB(s)}$ is smaller than that of $\hat{\theta}_n^{FB(s)}$. These simulation results confirm our discussion on the efficiency of mb-MPLE and their relationship with batch size $s$ in Cox regression. 

The theoretical asymptotic standard error (A-SE) of $\hat{\theta}^{SB(s)}_1$ and $\hat{\theta}^{FB(s)}_1$ for various batch sizes $s$ are calculated based on the asymptotic variance derived in Theorem \ref{thm:asymp-dist-SBFB}. Specifically, the asymptotic variance of $\hat{\theta}^{FB(s)}$ is $H^{-1}_s/n$. From (\ref{eq: Hessian and covariance}) in Theorem \ref{thm: batch size and Hessian}, we estimated $\hat{H}_s = s\hat{\mathbb{V}}[\nabla_\theta L^{(s)}_{Cox}(\theta)]\rvert_{\theta = \theta_0}$ by calculating the variance of $\nabla_\theta L^{(s)}_{Cox}(\theta_0)$ at the truth $\theta_0$ from 100,000 randomly generated mini-batches data, each of size $s$. Square root of $\{\hat H^{-1}_s/n\}_{(1,1)}$ is reported as the A-SE for $\hat{\theta}^{FB(s)}_1$.  For $\hat{\theta}^{SB(s)}$ with asymptotic variance $s^2H^{-1}_s\Sigma_{(s|1)}(H_s^{-1})^T/n$, $\hat \Sigma_{(s|1)}$ is estimated by calculating the covariance of $\nabla_\theta L_{Cox}^{(s)}(D_{1},D_{2},\dots,D_{s}|\theta_0)$ and \\ $\nabla_\theta L_{Cox}^{(s)}(D_{1},D_{s+1},\dots,D_{2s-1}|\theta_0)$ at the truth $\theta_0$ from 100,000 randomly generated mini-batches, each of size $2s-1$. Square root of $\{s^2\hat{H}^{-1}_s\hat\Sigma_{(s|1)}(\hat{H}_s^{-1})^T/n\}_{(1,1)}$ is reported as the A-SE for $\hat{\theta}^{SB(s)}_1$. 

Note that there is a computational challenge when numerically evaluating the asymptotic variance of $\hat{\theta}^{SB(s)}$, which is $s^2H^{-1}_s\Sigma_{(s|1)}(H_s^{-1})^T/n$. The term $\Sigma_{(s|1)}$ is the covariance between two scores $\nabla_\theta L_{Cox}^{(s)}(D_{1},D_{2},\dots,D_{s}|\theta_0)$ and $\nabla_\theta L_{Cox}^{(s)}(D_{1},D_{s+1},\dots,D_{2s-1}|\theta_0)$, sharing one common sample but differing in the rest. Notably, when $s$ is large, the covariance is close to zero, so an accurate estimation of $\Sigma_{(s|1)}$ requires a very large number of mini-batches due to its low signal level (close to zero). 
We used 10,000,000 randomly generated mini-batches to approximate $\Sigma_{(s|1)}$ for $s=64$. Due to the computational burden, the approximation of $s^2H^{-1}_s\Sigma_{(s|1)}(H_s^{-1})^T/n$ is not performed for $s \geq 128$ and its lower/upper bounds are provided instead. Mathematically, it is smaller than the asymptotic variance of $\hat{\theta}^{FB(s)}_1$ and greater than the asymptotic variance of MPLE, and these two bounds are very close when $s$ is large. The A-SE of the MPLE is 0.093, calculated as the average of the estimated variances from 1,000 CoxPH runs, where the software provides the variance estimates.

\begin{figure}[htbp]
\includegraphics[width=1\textwidth]{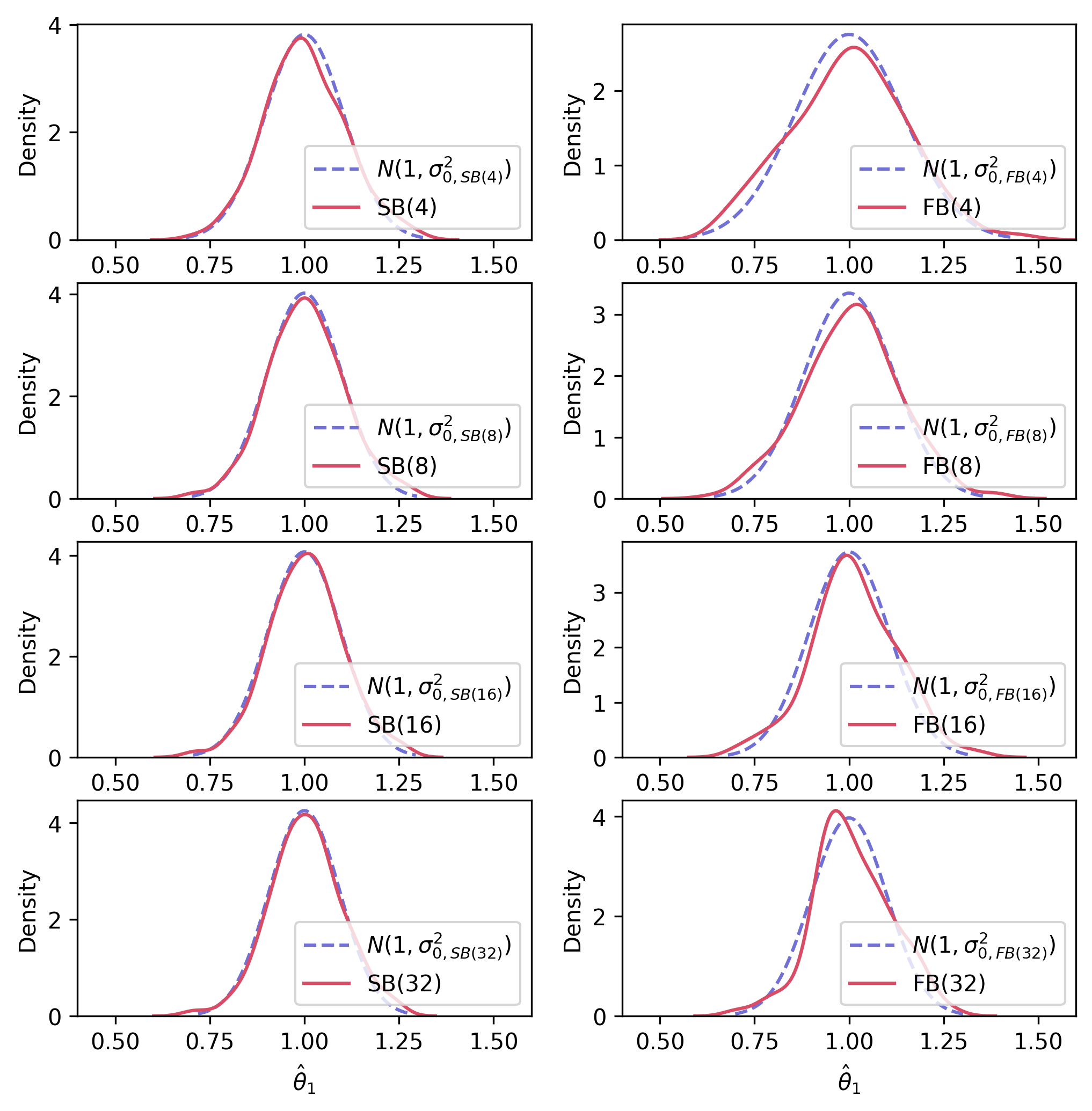}
\caption{The comparison between the theoretical asymptotic distribution and the empirical distribution for batch sizes $s\leq 32$, with the batch size indicated in brackets.. The density of the theoretical asymptotic distribution is based on the density of $N(1,\sigma^2_{0,s})$ with $\sigma_{0,s}$ from the A-SE column in Table \ref{tb: simulation theta1}. The empirical distribution is the kernel density estimation of the $\hat{\theta}_1$ distribution from 200 simulation runs.}
\label{fig:PH-reg-simulation-theta1-normality}
\end{figure}

Table \ref{tb: simulation theta1} demonstrates that the Std is closely aligned with the A-SE. The comparisons between the asymptotic normal distribution (with standard errors from the A-SE column in Table \ref{tb: simulation theta1}) and the empirical distribution for $s\leq 32$ are shown in Figure \ref{fig:PH-reg-simulation-theta1-normality}, supporting the asymptotic result of the mini-batch-based estimators as derived in Theorem \ref{thm:asymp-dist-SBFB}. Moreover, for all values of $s$, the Std of $\hat{\theta}^{SB(1)}_1$ is smaller than that of $\hat{\theta}^{FB(1)}_1$. Lastly, as $s$ doubles, both Std and A-SE decrease, though the reduction becomes negligible when $s$ is large. These simulation results confirm our discussion on the efficiency of the mb-MPLE and its relationship with batch size.

\subsubsection{Discussion}
The learning of $\theta$ is guided by the objective function $$L_{Cox}^{(s)}(\theta):=  -\frac{1}{s}\sum_{i:D_i \in D(s)} \Delta_i \log \frac{\exp\{f_\theta(X_i)\}}{\sum_{j: D_j \in D(s)} I(T_j\geq T_i)\exp\{f_\theta(X_j)\}},$$ which is a loss function depending on the ranking of $T_i$'s instead of the actual value of $T_i$'s. In this formulation, each subject $i\in D(s)$ is compared with all other subjects within $D(s)$. With a stochastic batch, the comparison between any two samples $i,j\in D(n)$ is possible because there is a chance for both $i,j$ to be selected in the same batch $D(s)$. In contrast, samples $i$ and $j$ will never be compared for the fixed batch strategy if they are not assigned to the same batch. Therefore, the fixed batch has no ranking information between samples from different non-overlapping batches. This loss of information consequently reduces the efficiency in estimating $\theta$.

The increase in batch size $s$ leads to an increase in the size of risk sets, which enhances the statistical efficiency in estimating the linear coefficient in the Cox regression. This is because the objective score function gets closer to the efficient score function of Cox regression when $s$ increases closer to $n$. 

Specifically, the efficient score \citep{tsiatis2006semiparametric} for Cox regression is 
$$S_{eff} = \Delta_i\left\{X_i-\frac{\mathbb{E}\{X\exp(\theta_0X)Y(T_i)|T_i\}}{\mathbb{E}\{\exp(\theta_0X)Y(T_i)|T_i\}}\right\}.$$
The estimator $\Tilde{\theta}_{MPLE}$, which has an efficient influence function (i.e., proportional to $S_{eff}$), is therefore semi-parametric efficient. On the other hand, the influence function of $\tilde{\theta}_{mb-MPLE}$ is proportional to
\begin{align*}
S_{s} &= \mathbb{E}\left.\left[\Delta_i\left\{X_i-\frac{\sum_{k=1}^s\{X_k\exp(\theta_0X_k)Y_k(T_i)\}}{\sum_{k=1}^s\{\exp(\theta_0X_k)Y_k(T_i)\}}\right\}\right|\Delta_i,X_i,T_i\right] \\
&= \Delta_i\left\{X_i-\mathbb{E}\left.\left[\frac{\sum_{k=1}^s\{X_k\exp(\theta_0X_k)Y_k(T_i)\}}{\sum_{k=1}^s\{\exp(\theta_0X_k)Y_k(T_i)\}}\right|\Delta_i,X_i,T_i\right]\right\}.
\end{align*}
Notice that when $s\to\infty$, $\frac{\sum_{k=1}^s\{X_k\exp(\theta_0X_k)Y_k(T_i)\}}{\sum_{k=1}^s\{\exp(\theta_0X_k)Y_k(T_i)\}}\to^p\frac{\mathbb{E}\{X\exp(\theta_0X)Y(T_i)|T_i\}}{\mathbb{E}\{\exp(\theta_0X)Y(T_i)|T_i\}}$. Hence, the influence function of $\tilde{\theta}_{mb-MPLE}$ gets closer to the efficient influence function as $s$ increases, leading to an increase in the statistical efficiency of the estimator. This explains why comparing each individual with more individuals provides more information about the parameter.

\begin{table}[htp]
\centering
\caption{The bias (Bias), empirical standard deviation (Std), and asymptotic standard error (A-SE) of $\hat{\theta}^{SB(s)}_1$ and $\hat{\theta}^{FB(s)}_1$ for various batch sizes $s$. The bias and standard deviation of the estimator are calculated from 200 simulation runs with sample size $n = 2,048$ in each run. The A-SE is approximated from 100,000 randomly generated mini-batches using the theoretical asymptotic variance formula presented in Theorem 2. When $s\geq 128$, the approximation of A-SE for $\hat{\theta}^{SB(s)}_1$ is not performed due to large computational burden for $\Sigma_{(s|1)}$. An interval is provided instead since this value is between the A-SE of $\hat{\theta}^{FB(s)}_1$ and the A-SE of MPLE mathematically.\label{tb: simulation theta1}
}
\begin{tabular}{c|ccc|ccc}
\hline
\multirow{2}{*}{$s$} & \multicolumn{3}{c|}{$\hat{\theta}^{SB(s)}_{1}$} & \multicolumn{3}{c}{$\hat{\theta}^{FB(s)}_{1}$} \\
\cline{2-7}
                   & Bias   & Std   & A-SE  & Bias   & Std   & A-SE  \\
\hline
4   & 0.002 & 0.106 & 0.104 & -0.002 & 0.152 & 0.145 \\
8   & 0.004 & 0.101 & 0.099 & 0.006  & 0.128 & 0.119 \\
16  & 0.005 & 0.098 & 0.098 & 0.013  & 0.115 & 0.107 \\
32  & 0.005 & 0.095 & 0.094 & 0.009  & 0.103 & 0.100 \\
64  & 0.005 & 0.094 & 0.094$^*$ & 0.008  & 0.100 & 0.097 \\
128 & 0.005 & 0.093 & 0.093-0.095$\dag$     & 0.007  & 0.097 & 0.095 \\
256 & 0.005 & 0.093 & 0.093-0.094$\dag$     & 0.006  & 0.095 & 0.094 \\
512 & 0.005 & 0.093 & 0.093$\dag$     & 0.006  & 0.094 & 0.093 \\
\hline
\multicolumn{7}{l}{$^*$ use 10,000,000 randomly generated mini-batches}\\
\multicolumn{7}{l}{$\dag$ based on A-SE of MPLE and A-SE of $\hat{\theta}^{FB(s)}_{1}$}
\end{tabular}
\end{table}

\subsection{Real Data Application}
The hyperparameters used to train Cox-NN in the real data application are obtained by grid search based on the candidate values
in Table \ref{tb:hyperparameters}. The tuning of the hyperparameters is identical to the one introduced in Section \ref{sec:Cox-NN Simulation}.

\begin{table}[htp]
\centering
\begin{threeparttable}[b]
\caption{Cox-NN hyperparameter configurations in real data application}
\begin{tabular}{ccc}
\hline
\multirow{3}{*}{Cox-NN parameters} & Number of layer$^*$  & \{1,2,4\}             \\
                                   & Number of neuron & \{16,32,64\}          \\
                                   & Dropout rate     & \{0.1,0.3\}           \\
\hline
\multirow{5}{*}{SGD parameters}    & Learning rate    & \{0.001,0.002,0.004\} \\
                                   & Batch Size       & 32                    \\
                                   & Maximum Epoch   & 200         \\
                                   & Early Stopping Patience   & 10         \\
                                   & Early Stopping Metric   & validation C-index\\
\hline
\end{tabular}
\label{tb:hyperparameters}
\begin{tablenotes}
\item [$*$] Each layer consists of the following structures: fully connected layer $\to$ batch normalization layer $\to$ activation function (ReLU) $\to$ drop out layer.
\end{tablenotes}
\end{threeparttable}
\end{table}

\newpage
\section{Partially Linear Cox Model}\label{sec:PLNN}

\subsection{Introduction and Assumptions}
In this section, we consider the partially linear Cox model 
\begin{equation}\label{eq:partial linear cox model}
\begin{split}
\lambda(t|X) &=  \lambda_0(t)\exp\{\theta_0^TZ+f_0(X)\},
\end{split}
\end{equation}
where $(Z,X)\in \mathbb{R}^p\times \mathbb{R}^r$ and $\theta_0\in\mathbb{R}^p$ is the parameter of interest. \citet{zhong2022deep} established the statistical properties of the estimator
$$(\Tilde{\theta},\Tilde{f}) = \arg\min_{(\theta,f)\in\mathbb{R}^p\times\mathcal{F}}L^{(n)}_{Cox}(\theta,f),$$
where 
$$L^{(n)}_{Cox}(\theta,f) = -\frac{1}{n}\sum_{i=1}^n \Delta_i \log \frac{\exp\{\zeta_{(\theta,f)}(V_i)\}}{\sum_{j=1}^n I(T_j\geq T_i)\exp\{\zeta_{(\theta,f)}(V_j)\}}.$$
Specifically, they showed that
$\lVert\Tilde\theta-\theta_0\rVert=O_p(\Upsilon_n\log^2n)$ and 
$$\left\lVert\left\{\Tilde f-\mathbb{E}_{X}[\Tilde{f}(X)]\right\}-f_0\right\rVert_{L^2}=O_p(\Upsilon_n\log^2n).$$
Further, they demonstrate $\Tilde\theta$ is $\sqrt{n}$-consistent and asymptotically normal with asymptotic variance equal to the information bound of $\theta_0$ (thus semi-parametric efficient). In this section, we employed the following assumptions to establish the statistical properties of mb-MPLE. Those assumptions have also been adopted in \citet{zhong2022deep}.

\begin{itemize}
    \item [(A1$^{\prime}$)] The failure time $T_i^*$ and censoring time $C_i^*$ are independent given the covariates $(Z_i,X_i)$.
    \item [(A2$^{\prime}$)] There is a truncation time $\tau < \infty $ such that, for some constant $\delta>0$, $\mathbb{P}(T^* >  \tau|Z,X)\geq \delta$ and $\mathbb{P}(\Delta=1|Z,X)\geq \delta$ almost surely with respect to the probability measure of $(Z,X)$. The stochastic integrals computed from here on will be truncated at this value $\tau$.
    \item [(A3$^{\prime}$)] The covariate $V = (Z,X)$ takes value in $\Omega_V$, a compact subset of $\mathbb{R}^{p+r}$, with joint probability density function bounded away from zero. Without loss of generality, we assume that the domain of $X$ is taken to be $[0,1]^{p}$. 
    \item[(N1)] The unknown function $f_0$ is an element of $\mathcal{H}_0=\{f \in \mathcal{H}(q,\vec{\alpha},\mathbf{d},\mathbf{\Tilde{d}},M):\mathbb{E}[f(X)] = 0\}.$ 
    \item[(N2)] $K=O(\log n)$, $\varsigma=O(n\Upsilon_n^2\log n)$, $\mathcal{D}>2M$, and $n\Upsilon_n\lesssim \displaystyle\min_{k=1,\dots,K} (p_k)\leq \displaystyle\max_{k=1,\dots,K} (p_k) \lesssim n$.
    \item[(R1$^{\prime}$)] The true parameter $\theta_0$ is an interior point of $\mathbb{R}_M^{p} :=\{\theta \in \mathbb{R}^{p}:\lVert\theta\rVert\leq M\}$.
    \item[(R2$^{\prime}$)] There exist constants $0<c_1<c_2<\infty$ such that the subdensities $p(v,t,\Delta = 1)$ of $(V,T,\Delta=1)$ satisfies $c_1<p(v,t,\Delta = 1)<c_2$ for all $(v,t)\in \Omega_V\times[0,\tau]$.
\end{itemize}

The subdensity $p(v,t,\Delta = 1)$ of measure $\mathbb{P}_{(V,T,\Delta=1)}$ in (R2$^{\prime}$) is defined as
$$p(v,t,\Delta = 1) = \frac{\partial^2 \mathbb{P}(V\leq v, T\leq t,\Delta = 1)}{\partial v \partial t},$$
which is well defined and also adopted in \citet{huang1999efficient}. 

\subsection{mb-MPLE for Partially Linear Cox Model}
We first introduce the necessary notations before presenting the statistical properties of mb-MPLE for the partially linear Cox model.
Let $V = (Z,X)\in \mathbb{R}^p\times \mathbb{R}^r$, $\vartheta:=(\theta,f)$,
$$\zeta_{0}(V):=\zeta_{\vartheta_0}(V)=\theta_0^TZ+f_0(X).$$
Define $d^2(\vartheta_1,\vartheta_2):=\mathbb{E}_V\{\zeta_{\vartheta_1}(V)-\mathbb{E}_V[\zeta_{\vartheta_1}(V)-\zeta_0(V)]-(\zeta_{\vartheta_2}(V)-\mathbb{E}_V[\zeta_{\vartheta_2}(V)-\zeta_0(V)])\}^2$, where the mean shift ensures that $\mathbb{E}_V[\zeta_{\vartheta}(V)] = \mathbb{E}_V[\zeta_0(V)]$.

We consider the following mb-MPLE with stochastic batch (SB) strategy or fixed batch (FB) strategy:
\begin{equation}
\Tilde{\vartheta}^{SB(s)}_n =(\Tilde{\theta}^{SB(s)}_n,\Tilde{f}^{SB(s)}_n) = \displaystyle\arg \min_{(\theta,f)\in \mathbb{R}^p\times\mathcal{F}} \frac{1}{\binom{n}{s}}\displaystyle\sum_{D(s)\subset D(n)}L_{Cox}^{(s)}(\theta,f),
\label{eq:SB PLNN minimizer}
\end{equation}
\begin{equation}
\Tilde{\vartheta}^{FB(s)}_n =(\Tilde{\theta}^{FB(s)}_n,\Tilde{f}^{FB(s)}_n) = \displaystyle\arg \min_{(\theta,f)\in \mathbb{R}^p\times\mathcal{F}} \frac{1}{m}\sum_{D(s)\in D(n|s)}L_{Cox}^{(s)}(\theta,f),
\label{eq:FB PLNN minimizer}
\end{equation}
where we use $D(n|s)$ to denote that $D(n)$ has been partitioned into $m = n/s$ fixed disjoint batches. The element of $D(n|s)$ is a mini-batch containing $s$ i.i.d. samples and $D(s)$ is generated by randomly picking one element from $D(n|s)$.

For any function $h$ of $V$, define
\begin{equation}
\label{eq:rho}
\rho(T_1,V_1,\vartheta)[h] := h(V_1)-\frac{\sum_{j=1}^s Y_j(T_1)\exp(\zeta_\vartheta(V_j))h(V_j)}{\sum_{j=1}^s Y_j(T_1)\exp(\zeta_\vartheta(V_j))},
\end{equation}
\begin{align*}
l^{(s)}(\vartheta)[h]:= -\frac{1}{s}\sum_{k=1}^s\Delta_k \rho(T_k,V_k,\vartheta)[h],
\end{align*}
\begin{align*}
l^{(s)}_0(\vartheta)[h]:&= \mathbb{E}\left[l^{(s)}(\vartheta)[h]\right] = \mathbb{E}\left[-\frac{1}{s}\sum_{k=1}^s\Delta_k \rho(T_k,V_k,\vartheta)[h]\right]\\
&=\mathbb{E}\left[ -\Delta_1\rho(T_1,V_1,\vartheta)[h]\right].
\end{align*}

Then the derivatives of $L_{Cox}^{(s)}(\vartheta)$ with respect to $\theta$ and with respect to $f$ in some direction $\mathbf{h}$ are
$$\nabla_\theta L_{Cox}^{(s)}(\vartheta) = \frac{1}{s}\sum_{k=1}^s\Delta_k \left\{-Z_k+\frac{\sum_{j=1}^s Y_j(T_k)\exp(\zeta_\vartheta(V_j))Z_j}{\sum_{j=1}^s Y_j(T_k)\exp(\zeta_\vartheta(V_j))} \right \} = l^{(s)}(\vartheta)[\mathbf{I}_Z],$$
$$\nabla_f L_{Cox}^{(s)}(\vartheta)[\mathbf{h}] = \frac{1}{s}\sum_{k=1}^s\Delta_k \left\{-\mathbf{h}(V_k)+\frac{\sum_{j=1}^s Y_j(T_k)\exp(\zeta_\vartheta(V_j))\mathbf{h}(V_j)}{\sum_{j=1}^s Y_j(T_k)\exp(\zeta_\vartheta(V_j))} \right \} = l^{(s)}(\vartheta)[\mathbf{h}],$$
 where $\mathbf{I}_Z(V) = Z$. 

Let $\mathcal{H}_{f_0}$ denote the collection of all subfamilies $\{f_b\in L^2([0,1]^r):b\in (-1,1)\}\subset\mathcal{H}_0$ such that $\lim_{b\to 0}\lVert b^{-1}(f_b-f_0)\rVert_{L^2_{[0,1]^r}}=0$, and let 
$$\mathbb{T}_{f_0} = \left\{f\in L^2([0,1]^r): \lim_{b\to 0}\lVert b^{-1}(f_b-f_0)-f\rVert_{L^2_{[0,1]^r}}= 0 \text{ for some subfamilies } \{f_b:b\in (-1,1)\}\subset\mathcal{H}_{f_0}\right\}.$$
Set $\bar{\mathbb{T}}_{f_0}$ be the closed linear span of ${\mathbb{T}}_{f_0}$.

\begin{theorem}
\label{thm: PLNN}
For partially linear Cox model (\ref{eq:partial linear cox model}), under assumptions (A1$^{\prime}$)-(A3$^{\prime}$), (N1)-(N2), and (R1$^{\prime}$), for any $s\geq 2$, we have
$$d(\Tilde\vartheta^{SB(s)}_n,\vartheta_0) = O_p(\Upsilon_n\log^2n)$$
$$d(\Tilde\vartheta^{FB(s)}_n,\vartheta_0) = O_p(\Upsilon_n\log^2n).$$
Additionally, with assumptions (R2$^{\prime}$) and suppose $\mathbf{h^*}$ defined below is also an element of $(\mathcal{H}_0)^p$, if $n\gamma^4_n\to \infty$ as $n\to\infty$,  then we have
$$\sqrt{n}(\tilde{\theta}^{SB(s)}_n-\theta_0)\overset{d}{\to}N(0,s^2{H}_s^{-1}{\Sigma}_{(s|1)}{H}_s^{-1}),$$
$$\sqrt{n}(\tilde{\theta}^{FB(s)}_n-\theta_0)\overset{d}{\to}N(0,s{H}_s^{-1}{\Sigma}_s{H}_s^{-1}),$$
where
$${H}_s = \mathbb{E}\left[\Delta_1 \sum_{j=1}^s {w}_j(T_1,0)\left\{\rho(T_1,V_j,\vartheta_0)[\mathbf{I}_Z-\mathbf{h^*}]\right\}^{\otimes 2}\right]$$
$${\Sigma}_s = \mathbb{E}\left[l^{(s)}(\vartheta_0)[\mathbf{I}_Z-\mathbf{h^*}]^{\otimes 2}\right]$$
$$ {\Sigma}_{(s|1)} = \mathbb{V}\left \{ l^{(s)}(\vartheta_0)[\mathbf{I}_Z-\mathbf{h^*}](D_{i_1},D_{i_2},\dots,D_{i_s}),l^{(s)}(\vartheta_0)[\mathbf{I}_Z-\mathbf{h^*}](D_{i_1},\Tilde D_{{i}_2},\dots,\Tilde D_{{i}_s})\right \}, $$
which is the covariance of $l^{(s)}(\vartheta_0)[\mathbf{I}_Z-\mathbf{h^*}]$ on two mini-batches $D(s)$ sharing the same sample $D_{i_1}$ but different rest $s-1$ samples (denoted by $\Tilde D$). The vector function $\mathbf{h^*}\in (\bar{\mathbb{T}}_{f_0})^p$ minimizes
$$\mathbb{E}\left[\Delta_1 \sum_{j=1}^s {w}_j(T_1,0)\left\lVert\rho(T_1,V_j,\vartheta_0)[\mathbf{I}_Z-\mathbf{h}]\right\rVert^2_c\right] \text{ with }w_j(u,0) = \frac{Y_j(u)\exp(\zeta_{0}(V_j))}{\sum_{k=1}^s Y_k(u)\exp(\zeta_{0}(V_k))},$$
where the notation $\lVert v \rVert_c^2 = (v_1^2,\dots,v_p^2)^T$ for vector $v = (v_1,\dots,v_p)^T$ and the minimization operates component-wise on the vector.

Furthermore, ${H}_s = s{\Sigma}_s,$ and
${H}_{2s} \succeq {H}_{s}.$ Therefore, the asymptotic variance of $\tilde{\theta}^{FB(s)}_n$ is $s{H}_s^{-1}{\Sigma}_s{H}_s^{-1} = H_s^{-1}$ and is non-increasing when $s$ doubles.
\end{theorem}

\begin{proof}
By the same proof of Lemma \ref{lemma: minimizer does not depend on s}, we have 
$$\mathbb{E}[L_{Cox}^{(s)}(\vartheta)]-\mathbb{E}[L_{Cox}^{(s)}(\vartheta_0)]\asymp  d^2(\vartheta,\vartheta_0).$$
Follow the proof of Theorem \ref{thm:convergence rate}, it can be shown that $d(\Tilde\vartheta^{SB(s)}_n,\vartheta_0) = O_p(\Upsilon_n\log^2n)$. Because $s$ is a fixed number and $m = \frac{n}{s}$, we have $d(\Tilde\vartheta^{FB(s)}_n,\vartheta_0) = O_p(\Upsilon_n\log^2n)$.

Next, we derive the asymptotic distribution of $\tilde{\theta}^{SB(s)}_n$. Note that the derivation for the asymptotic distribution of $\tilde{\theta}^{FB(s)}_n$ follows a similar approach.

By definition, $\mathbf{h^*}\in (\bar{\mathbb{T}}_{f_0})^p$ minimizes
\begin{equation}
\label{eq: minimization}
\mathbb{E}\left[\Delta_1 \sum_{j=1}^s {w}_j(T_1,0)\left\lVert\rho(T_1,V_j,\vartheta_0)[\mathbf{I}_Z-\mathbf{h}]\right\rVert^2_c\right]
\end{equation}
component-wise. Therefore, for any $k \in \{1,\dots,p\}$, the $k$th element of $\mathbf{h^*}$ (denoted by $\mathbf{h}^*_k$) minimizes
\begin{align*}
& \mathbb{E}\left[\Delta_1 \sum_{j=1}^s {w}_j(T_1,0)\rho(T_1,V_j,\vartheta_0)[\mathbf{I}_{Z,k}-h]^2\right] \\
& = \langle \mathbf{I}_{Z,k}-h, \mathbf{I}_{Z,k}-h\rangle_s \\
(*) & = \langle \mathbf{I}_{Z,k}-\mathbb{E}_{Z_k}[Z_k]-h, \mathbf{I}_{Z,k}-\mathbb{E}_{Z_k}[Z_k]-h\rangle_s,
\end{align*}
for all $h\in \bar{\mathbb{T}}_{f_0}$ with $\langle \cdot, \cdot \rangle_s$ given in Lemma \ref{PLlemma: inner product}, where $\mathbb{E}_{Z_k}[Z_k]$ is the mean shift. The last equation $(*)$ holds because the operation $\langle \cdot, \cdot \rangle_s$ is invariant under any constant shift. That is $\langle \cdot, \cdot \rangle_s = \langle \cdot+a, \cdot \rangle_s = \langle \cdot, \cdot+a \rangle_s$ for any constant $a\in \mathbb{R}$. Note that $h\in \bar{\mathbb{T}}_{f_0}$ is mean zero by definition of the function space. By Lemma \ref{PLlemma: inner product}, the $\langle \cdot, \cdot \rangle_s$ is an inner product of mean zero functions. Therefore, the least favorable direction $\mathbf{h}^*_k$ is the projection of $I_{Z,k}-\mathbb{E}[Z]$ onto the space $\bar{\mathbb{T}}_{f_0}$ which is a sub-space of the Hilbert space defined in Lemma \ref{PLlemma: inner product}. Hence, 
$$\langle \mathbf{I}_{Z,k}-\mathbf{h}^*_k, f\rangle = \langle \mathbf{I}_{Z,k}-\mathbb{E}_{Z_k}[Z_k]-\mathbf{h}^*_k, f\rangle = 0,$$
which indicates
\begin{equation}
\label{eq: orthogonal elementwise}
\mathbb{E}\left[\Delta_1 \sum_{j=1}^s {w}_j(T_1,0)\rho(T_1,V_j,\vartheta_0)[\mathbf{I}_{Z,k}-\mathbf{h}^*_k]\rho(T_1,V_j,\vartheta_0)[f]\right] = 0
\end{equation}
for all $f\in \bar{\mathbb{T}}_{f_0}$ and all $k\in\{1,\dots,p\}$. This implies
\begin{equation}
\label{eq: orthogonal}
\mathbb{E}\left[\Delta_1 \sum_{j=1}^s {w}_j(T_1,0)\rho(T_1,V_j,\vartheta_0)[\mathbf{I}_Z-\mathbf{h^*}]\rho(T_1,V_j,\vartheta_0)[f]\right] = 0_{p\times 1}
\end{equation}
for all $f\in \bar{\mathbb{T}}_{f_0}$. 

Because
\begin{equation}
\begin{split}
&\mathbb{E}\left[\Delta_1 \sum_{j=1}^s {w}_j(T_1,0)\left\lVert\rho(T_1,V_j,\vartheta_0)[\mathbf{I}_Z-\mathbf{h}]\right\rVert^2_c\right] \\
&= \mathbb{E}\left[\frac{1}{s}\sum_{k=1}^s\Delta_k \sum_{j=1}^s {w}_j(T_k,0)\left\lVert\rho(T_k,V_j,\vartheta_0)[\mathbf{I}_Z-\mathbf{h}]\right\rVert^2_c\right],
\end{split}
\end{equation}
in practice, we can estimate $\mathbf{h^*}\in (\bar{\mathbb{T}}_{f_0})^p$ through a neural network $\mathbf{h}$ by minimizing
$$
\frac{1}{\binom{n}{s}}\displaystyle\sum_{D(s)\subset D(n)}\left[\frac{1}{s}\sum_{k=1}^s\Delta_k \sum_{j=1}^s {w}_j(T_k,0)\left\lVert\rho(T_k,V_j,\vartheta_0)[\mathbf{I}_Z-\mathbf{h}]\right\rVert^2_c\right],
$$
with plugging in the consistent estimator $\hat{\vartheta}$ for $\vartheta_0$, for example, $\hat{\vartheta}^{SB(s)}_n$.

According to Lemma \ref{PLlemma: empirical}, we have
\begin{align*}
    & \frac{1}{\binom{n}{s}}\displaystyle\sum_{D(s)\subset D(n)}l^{(s)}(\Tilde\vartheta^{SB(s)}_n)[\mathbf{I}_Z-\mathbf{h}^*]-\frac{1}{\binom{n}{s}}\displaystyle\sum_{D(s)\subset D(n)}l^{(s)}(\vartheta_0)[\mathbf{I}_Z-\mathbf{h}^*] \\
    & =  \mathbb{E}\left[\Delta_1 \sum_{j=1}^s {w}_j(T_1,0)\left\{\rho(T_1,V_j,\vartheta_0)[\mathbf{I}_Z-\mathbf{h^*}]\right\}^{\otimes 2}\right](\Tilde{\theta}^{SB(s)}_n-\theta_0) + o_p(1/\sqrt{n}).
\end{align*}
By definition of $\Tilde\vartheta^{SB(s)}_n = (\Tilde{\theta}^{SB(s)}_n,\Tilde{f}^{SB(s)}_n)$, we have 
$\frac{1}{\binom{n}{s}}\displaystyle\sum_{D(s)\subset D(n)}l^{(s)}(\Tilde\vartheta^{SB(s)}_n)[\mathbf{I}_Z] = 0$ and $\frac{1}{\binom{n}{s}}\displaystyle\sum_{D(s)\subset D(n)}l^{(s)}(\Tilde\vartheta^{SB(s)}_n)[\mathbf{h}^*] = o_p(1/\sqrt{n})$ by Lemma \ref{PLlemma: least favorable direction}. This shows that the first term is $o_p(1/\sqrt{n})$ and
$$  H_s(\Tilde{\theta}^{SB(s)}_n-\theta_0) = -\frac{1}{\binom{n}{s}}\displaystyle\sum_{D(s)\subset D(n)}l^{(s)}(\vartheta_0)[\mathbf{I}_Z-\mathbf{h}^*] + o_p(1/\sqrt{n}).$$
The proof of Lemma \ref{PLlemma: AVar} shows that $l^{(s)}_0(\vartheta_0)[\mathbf{I}_Z-\mathbf{h}^*]=0_{p\times 1}$. By the central limit theorem for U-statistics, we have
$$\sqrt{n} \frac{1}{\binom{n}{s}}\displaystyle\sum_{D(s)\subset D(n)}l^{(s)}(\vartheta_0)[\mathbf{I}_Z-\mathbf{h}^*]\overset{d}{\to}N(0,s^2\Sigma_{(s|1)}).$$
By assumptions (A3$^\prime$), (R2$^\prime$) and Lemma \ref{PLlemma: inner product}, $H_s$ is not singular. This gives the asymptotic distribution of $\hat{\theta}^{SB(s)}_n$. 

Furthermore, by Lemma \ref{PLlemma: AVar}, we have
 ${H}_s = s{\Sigma}_s,$ and
${H}_{2s} \succeq {H}_{s}.$
\end{proof}

\subsection{Technical Lemmas for Partially Linear Cox Model}

\begin{lemma}\label{PLlemma: derivative}
Let
$$w_j(u,v,d) = \frac{Y_j(u)\exp(\zeta_{0}(V_j)+vd(V_j))}{\sum_{k=1}^s Y_k(u)\exp(\zeta_{0}(V_k)+vd(V_k))},$$
$$W(u,v)[h] = \sum_{k=1}^s w_k(u,v,d)h(V_k),$$
$$ W(u,v)[h_1h_2] = \sum_{k=1}^s w_k(u,v,d)h_1(V_k)h_2(V_k),$$
then
\begin{align*}
&  \frac{\partial}{\partial v}\mathbb{E}\left[\Delta_1 \left\{-h(V_1)+\sum_{j=1}^s w_j(T_1,v)h(V_j) \right \}\right] \\
= & \mathbb{E}\left[\Delta_1 \left\{W(T_1,v)[hd]-W(T_1,v)[h]W(T_1,v)[d]\right \}\right],
\end{align*}
and
\begin{align*}
&  \frac{\partial^2}{\partial v^2}\mathbb{E}\left[\Delta_1 \left\{-h(V_1)+\sum_{j=1}^s w_j(T_1,v)h(V_j) \right \}\right] \\
= & \mathbb{E}\left[\Delta_1 \left\{W(T_1,v)[hd^2]\right \}\right]-\mathbb{E}\left[\Delta_1 \left\{W(T_1,v)[hd]W(T_1,v)[d]\right \}\right] \\
& -\mathbb{E}\left[\Delta_1 \left\{W(T_1,v)[hd]W(T_1,v)[d^2]\right \}\right]+\mathbb{E}\left[\Delta_1 \left\{W(T_1,v)[h]W(T_1,v)[d]W(T_1,v)[d^2]\right \}\right]\\
& +\mathbb{E}\left[\Delta_1 \left\{W(T_1,v)[hd]W(T_1,v)[d]W(T_1,v)[d]\right \}\right] \\
& -\mathbb{E}\left[\Delta_1 \left\{W(T_1,v)[h]W(T_1,v)[d]W(T_1,v)[d]W(T_1,v)[d]\right \}\right],
\end{align*}
\end{lemma}

\begin{proof}
The result follows from direct calculation.
\end{proof}

\begin{lemma}\label{PLlemma: inner product}
Let $\Omega_V\subset \mathbb{R}^{p+r}$ denote the sample space of $V$, 
$$H = \left\{h:h\in L^\infty(\Omega_V,\mathcal{B}(\Omega_V),\mathbb{P}_{(V,T,\Delta=1)}) \text{ and } \int h(V) \mathrm{d}\mathbb{P}_{(V,T,\Delta=1)} = 0\right\},$$ which is the collection of real-valued random functions with mean zero with respect to the probability measure of $(V,T,\Delta=1)$ (defined in Assumption (R2$^\prime$)), and is finite almost everywhere. For any $h_1,h_2\in H$, define
$$\langle h_1,h_2\rangle_s:=\mathbb{E}\left[\Delta_1 \sum_{j=1}^s {w}_j(T_1,0)\rho(T_1,V_j,\vartheta_0)[h_1]\rho(T_1,V_j,\vartheta_0)[h_2]\right].$$
Then $H$ with $\langle \cdot , \cdot \rangle_s$ is a Hilbert space.
\end{lemma}

\begin{proof} Because $H$ is a vector space, what remains is to show $\langle \cdot , \cdot \rangle_s$ is an inner product.
Notice that for any random function $h\in L^\infty(\Omega_V,\mathcal{B}(\Omega_V),\mathbb{P}_{(V,T,\Delta=1)})$, we have $\tilde{h} = h-\int h(V) \mathrm{d}\mathbb{P}_{(V,T,\Delta=1)} = 0 \in H$.

The symmetry and linearity can be easily verified. We will show that  $\langle h,h\rangle_s\geq 0$ and the equality holds if only if $h=0$ almost everywhere. 
\begin{align*}
\langle h,h\rangle_s & =\mathbb{E}\left[\Delta_1 \sum_{j=1}^s {w}_j(T_1,0)\{\rho(T_1,V_j,\vartheta_0)[h]\}^2\right] \\
& = \frac{1}{s}\mathbb{E}\left[\sum_{k=1}^s\Delta_k \sum_{j=1}^s {w}_j(T_k,0)\{\rho(T_k,V_j,\vartheta_0)[h]\}^2\right] \\
& \geq \frac{1}{s}\mathbb{E}\left[\sum_{k=1}^s \sum_{j=1}^s {w}_j(T_k,0)\{\rho(T_k,V_j,\vartheta_0)[h]\}^2\left|\sum_{k=1}^s\Delta_k = s\right]\right.\mathbb{P}\left[\sum_{k=1}^s\Delta_k = s\right].
\end{align*}
Note that $\mathbb{P}\left[\sum_{k=1}^s\Delta_k = s\right]>0$ under assumption (A1$^{\prime}$) and (A2$^{\prime}$). Therefore $\langle h,h\rangle_s = 0$ implies that, for any $k\in\{1,\dots,s\}$,
\begin{equation}
\label{eq: h=0}
\begin{split}
    & \mathbb{E}\left[\sum_{j=1}^s {w}_j(T_{k},0)\{\rho(T_{k'},V_j,\vartheta_0)[h]\}^2\left|\sum_{k=1}^s\Delta_k = s\right]\right. \\
    = & \int  \sum_{j=1}^s {w}_j(T_{k},0)\{\rho(T_{k},V_j,\vartheta_0)[h]\}^2 \mathrm{d} \left(\mathbb{P}_{(V_1,T_1,\Delta_1=1)} \times \mathbb{P}_{(V_2,T_2,\Delta_2=1)}\times\cdots \times\mathbb{P}_{(V_s,T_s,\Delta_s=1)} \right) \\
    = & 0.
\end{split}
\end{equation}
For any realization $\{(V_i,T_i,\Delta_i=1)\}_{i=1}^s$, suppose no tie, there exists $k'\in\{1,\dots,s\}$ such that ${w}_j(T_{k'},0)>0$ for all $j=1,\dots,s$.
For simplicity, we denote the positive weight ${w}_j(T_{k'},0)$ by $w_j$ and denote $h(V_j)$ by $h_j$, then by definition of $\rho$ in (\ref{eq:rho})
\begin{align*}
&\sum_{j=1}^s {w}_j(T_{k'},0)\{\rho(T_{k'},V_j,\vartheta_0)[h]\}^2 \\
=& \sum_{j=1}^s \left\{w_j\left(h_j - \sum_{k=1}^s w_k h_k\right)\left(h_j - \sum_{k=1}^s w_kh_k\right)\right \} \\
=& \sum_{j=1}^s \left\{w_j h_j \left(h_j - \sum_{k=1}^s w_kh_k\right)\right \} \\
=& \sum_{j=1}^s w_j h_j^2  - \left(\sum_{j=1}^s w_j h_j\right)\left(\sum_{k=1}^s w_kh_k\right) \\
=& \left(\sum_{j=1}^s w_j h_j^2\right) \left(\sum_{j=1}^s w_j\right) - \left(\sum_{j=1}^s w_j h_j\right)^2.
\end{align*}
When $w_j$s are positive, by Cauchy–Schwarz inequality, $\sum_{j=1}^s {w}_j(T_{k'},0)\{\rho(T_{k'},V_j,\vartheta_0)[h]\}^2 \geq 0$ and the equality holds if and only if $h_1 = h_2 = \dots = h_s$. Therefore, consider two events
\begin{align*}
 E_1:&=\left\{\{(V_i,T_i,\Delta_i=1)\}_{i=1}^s:\sum_{k=1}^s \sum_{j=1}^s {w}_j(T_k,0)\{\rho(T_k,V_j,\vartheta_0)[h]\}^2 = 0\right\}, \\
 E_2: &= \left\{\{(V_i,T_i,\Delta_i=1)\}_{i=1}^s:h(V_1)=h(V_2) = \dots=h(V_s)\right\},
\end{align*}
and we have $E_1\subseteq E_2$. By (\ref{eq: h=0}) and the Assumption (R2$^\prime$) that $\mathrm{d} \mathbb{P}_{(V,T,\Delta=1)}$ is bounded from below, the condition $\langle h,h\rangle_s = 0$ implies $E_1$ happens with probability 1. This, along with $E_1\subseteq E_2$, implies $h(V_i) = h(V_j)$ almost everywhere with respect to the probability measure of $(V,T,\Delta=1)$ and it follows
\begin{equation}
\label{eq: inner product}
    \begin{split}
        0 & = \int (h(V_i)-h(V_j))^2 \mathrm{d}\mathbb{P}_{(V_i,T_i,\Delta_i=1)\times (V_j,T_j,\Delta_j=1)} \\
& = 2\int h(V_i)^2 \mathrm{d}\mathbb{P}_{(V_i,T_i,\Delta_i=1)} -2\int h(V_i) \mathrm{d}\mathbb{P}_{(V_i,T_i,\Delta_i=1)}\int h(V_j) \mathrm{d}\mathbb{P}_{(V_j,T_j,\Delta_j=1)}  \\
& = 2\int h(V_i)^2 \mathrm{d}\mathbb{P}_{(V_i,T_i,\Delta_i=1)},
    \end{split}
\end{equation}
where the second equation is due to $i$ and $j$ are two i.i.d. samples and the Fubini's thereom. The last equation holds because $\int h(V) \mathrm{d}\mathbb{P}_{(V,T,\Delta=1)} = 0$ for $h\in H$.
Equation (\ref{eq: inner product}) implies $h=0$ almost everywhere under $\mathbb{P}_{(V,T,\Delta=1)}$. This concludes that $\langle h,h\rangle_s = 0 \iff h = 0 \ a.s.$ under $\mathbb{P}_{(V,T,\Delta=1)}$. Hence $\langle \cdot , \cdot \rangle_s$ is an inner product. 

Note that, by following the proof strategy of Lemma \ref{PLlemma: AVar}, one can show that $\langle h,h\rangle_{2s} \geq \langle h,h\rangle_s$ for $h\in H$.
\end{proof}

\begin{lemma}\label{PLlemma: empirical}
\begin{align*}
    & \frac{1}{\binom{n}{s}}\displaystyle\sum_{D(s)\subset D(n)}l^{(s)}(\Tilde\vartheta^{SB(s)}_n)[\mathbf{I}_Z-\mathbf{h}^*]-\frac{1}{\binom{n}{s}}\displaystyle\sum_{D(s)\subset D(n)}l^{(s)}(\vartheta_0)[\mathbf{I}_Z-\mathbf{h}^*] \\
    & =  \mathbb{E}\left[\Delta_1 \sum_{j=1}^s {w}_j(T_1,0)\left\{\rho(T_1,V_j,\vartheta_0)[\mathbf{I}_Z-\mathbf{h^*}]\right\}^{\otimes 2}\right](\Tilde{\theta}^{SB(s)}_n-\theta_0) + o_p(1/\sqrt{n}).
\end{align*}

\end{lemma}

\begin{proof}
We simplify the notation $\Tilde\vartheta^{SB(s)}_n$ by $\Tilde\vartheta$ in this proof and first show that 
\begin{equation}
\label{eq: PLlemma: empirical 1}
\begin{split}
    & \frac{1}{\binom{n}{s}}\displaystyle\sum_{D(s)\subset D(n)}l^{(s)}(\Tilde\vartheta)[\mathbf{I}_Z-\mathbf{h}^*]-\frac{1}{\binom{n}{s}}\displaystyle\sum_{D(s)\subset D(n)}l^{(s)}(\vartheta_0)[\mathbf{I}_Z-\mathbf{h}^*] \\
    & =  l^{(s)}_0(\Tilde\vartheta)[\mathbf{I}_Z-\mathbf{h^*}]-l^{(s)}_0(\vartheta_0)[\mathbf{I}_Z-\mathbf{h^*}] + o_p(1/\sqrt{n}).
\end{split}
\end{equation}
To prove (\ref{eq: PLlemma: empirical 1}), it is sufficient to verify
\begin{equation}
\label{eq: PLlemma: empirical 2}
\begin{split}
    &\frac{1}{\binom{n}{s}}\displaystyle\sum_{D(s)\subset D(n)}l^{(s)}(\Tilde\vartheta)[\mathbf{I}_Z]-\frac{1}{\binom{n}{s}}\displaystyle\sum_{D(s)\subset D(n)}l^{(s)}(\vartheta_0)[\mathbf{I}_Z] \\
    =&  l^{(s)}_0(\Tilde\vartheta)[\mathbf{I}_Z]-l^{(s)}_0(\vartheta_0)[\mathbf{I}_Z] + o_p(1/\sqrt{n}),
\end{split}
\end{equation}
and
\begin{equation}
\label{eq: PLlemma: empirical 3}
\begin{split}
    &\frac{1}{\binom{n}{s}}\displaystyle\sum_{D(s)\subset D(n)}l^{(s)}(\Tilde\vartheta)[\mathbf{h^*}]-\frac{1}{\binom{n}{s}}\displaystyle\sum_{D(s)\subset D(n)}l^{(s)}(\vartheta_0)[\mathbf{h^*}] \\
    =&  l^{(s)}_0(\Tilde\vartheta)[\mathbf{h^*}]-l^{(s)}_0(\vartheta_0)[\mathbf{h^*}] + o_p(1/\sqrt{n}).
\end{split}
\end{equation}

We only prove (\ref{eq: PLlemma: empirical 3}) since the proof of (\ref{eq: PLlemma: empirical 2}) is similar. By reorganizing the terms in (\ref{eq: PLlemma: empirical 3}), it is equivalent to showing that
$$\frac{1}{\binom{n}{s}}\displaystyle\sum_{D(s)\subset D(n)}\left\{\left(l^{(s)}(\Tilde\vartheta)[\mathbf{h^*}]-l^{(s)}_0(\Tilde\vartheta)[\mathbf{h^*}]\right)-\left(l^{(s)}(\vartheta_0)[\mathbf{h^*}]-l^{(s)}_0(\vartheta_0)[\mathbf{h^*}]\right)\right\} = o_p(1/\sqrt{n}).$$
Following the similar proof of Lemma \ref{lemma: convergence rate}, we obtain that the order of the left-hand side is
$ O(\frac{d(\Tilde\vartheta,\vartheta_0)\sqrt{\varsigma\log U}}{\sqrt{n}})$. Because $d(\Tilde\vartheta,\vartheta_0) = O_p(\Upsilon_n\log^2n)$, with the assumption that $\sqrt{n}\Upsilon_n^2\to 0$ as $n\to\infty$, we have $ O(\frac{d(\Tilde\vartheta,\vartheta_0)\sqrt{\varsigma\log U}}{\sqrt{n}}) = o_p(1/\sqrt{n})$. This proves (\ref{eq: PLlemma: empirical 3}) and (\ref{eq: PLlemma: empirical 1}) follow.

By Taylor expansion of $l_0^{(s)}(\Tilde\vartheta)[\mathbf{I}_Z-\mathbf{h}^*]$ at $\vartheta_0$ and Lemma \ref{PLlemma: derivative}, we have
\begin{equation}
\label{eq: PLlemma: empirical 4}
\begin{split}
&  l^{(s)}_0(\Tilde\vartheta)[\mathbf{I}_Z-\mathbf{h^*}]-l^{(s)}_0(\vartheta_0)[\mathbf{I}_Z-\mathbf{h^*}]\\
= &  \mathbb{E}\left[\Delta_1 \sum_{j=1}^s {w}_j(T_1,0)\rho(T_1,V_j,\vartheta_0)[\mathbf{I}_Z-\mathbf{h^*}]\rho(T_1,V_j,\vartheta_0)[\mathbf{I}_Z]\right](\Tilde{\theta}-\theta_0) \\
& + \mathbb{E}\left[\Delta_1 \sum_{j=1}^s \left\{{w}_j(T_1,0)\rho(T_1,V_j,\vartheta_0)[\mathbf{I}_Z-\mathbf{h^*}]\rho(T_1,V_j,\vartheta_0)[\Tilde{f}-f_0]\right \}\right] \\
& + O(d^2(\Tilde\vartheta,\vartheta_0)).
\end{split}
\end{equation}

For the third term, because $d(\Tilde\vartheta,\vartheta_0) = O_p(\Upsilon_n\log^2n)$ and assumption $\sqrt{n}\Upsilon_n^2\to 0$ as $n\to\infty$, we have $d^2(\Tilde\vartheta,\vartheta_0) = o_p(1/\sqrt{n})$. Moreover, by property of $\mathbf{h^*}$ in (\ref{eq: orthogonal}), the second term is zero and
\begin{equation}
\label{eq: PLlemma: empirical 5}
\begin{split}
& \mathbb{E}\left[\Delta_1 \sum_{j=1}^s {w}_j(T_1,0)\rho(T_1,V_j,\vartheta_0)[\mathbf{I}_Z-\mathbf{h^*}]\rho(T_1,V_j,\vartheta_0)[\mathbf{I}_Z]\right] \\
& = \mathbb{E}\left[\Delta_1 \sum_{j=1}^s {w}_j(T_1,0)\left\{\rho(T_1,V_j,\vartheta_0)[\mathbf{I}_Z-\mathbf{h^*}]\right\}^{\otimes 2}\right].    
\end{split}    
\end{equation}
This, together with (\ref{eq: PLlemma: empirical 1}) and (\ref{eq: PLlemma: empirical 4}), completes the proof.

\end{proof}

\begin{lemma}
\label{PLlemma: least favorable direction}
$$\frac{1}{\binom{n}{s}}\displaystyle\sum_{D(s)\subset D(n)}\nabla_f L_{Cox}^{(s)}(\Tilde\vartheta^{SB(s)}_n)[\mathbf{h}^*] := \frac{1}{\binom{n}{s}}\displaystyle\sum_{D(s)\subset D(n)}l^{(s)}(\Tilde\vartheta^{SB(s)}_n)[\mathbf{h}^*] = o_p(1/\sqrt{n}).$$
\end{lemma}

\begin{proof}
It suffices to show that $\frac{1}{\binom{n}{s}}\displaystyle\sum_{D(s)\subset D(n)}l^{(s)}(\Tilde\vartheta^{SB(s)}_n)[\mathbf{h}^*_k] = o_p(1/\sqrt{n})$ for all $k\in\{1,\dots,p\}$. For simplicity, we denote $\mathbf{h}^*_k$ by $h^*$.

With assumption that $h^*$ is an element of $\mathcal{H}_0$, according to the proof of Theorem 1 in \cite{schmidt2020nonparametric}, there exists $h^*_n\in \mathcal{F}$ such that $\lVert h^*_n-h^*\rVert_{L^2}=O(\Upsilon_n\log^2n)$.

By (\ref{eq:SB PLNN minimizer}), we have $\frac{1}{\binom{n}{s}}\displaystyle\sum_{D(s)\subset D(n)}l^{(s)}(\Tilde\vartheta^{SB(s)}_n)[h^*_n] = 0$ so that
\begin{align*}
& \frac{1}{\binom{n}{s}}\displaystyle\sum_{D(s)\subset D(n)}l^{(s)}(\Tilde\vartheta^{SB(s)}_n)[h^*] \\
&= \frac{1}{\binom{n}{s}}\displaystyle\sum_{D(s)\subset D(n)}l^{(s)}(\Tilde\vartheta^{SB(s)}_n)[h^*]-\frac{1}{\binom{n}{s}}\displaystyle\sum_{D(s)\subset D(n)}l^{(s)}(\Tilde\vartheta^{SB(s)}_n)[h^*_n]\\
&=\underbrace{\frac{1}{\binom{n}{s}}\displaystyle\sum_{D(s)\subset D(n)}l^{(s)}(\Tilde\vartheta^{SB(s)}_n)[h^*-h^*_n] - l^{(s)}_0(\Tilde\vartheta^{SB(s)}_n)[h^*-h^*_n]}_\text{$I_{1n}$}  + \underbrace{l^{(s)}_0(\Tilde\vartheta^{SB(s)}_n)[h^*-h^*_n]}_\text{$I_{2n}$}.
\end{align*}
By a similar proof of Lemma \ref{lemma: convergence rate} and entropy calculation, it follows that $I_{1n} = o_p(1/\sqrt{n})$. For term $I_{2n}$, Similarly, by (\ref{ap-eq:compensator}), we have 
\begin{align*}
    \mathbb{E}\left[-\frac{1}{s}\sum_{k=1}^s\Delta_k[h^*-h^*_n](V_k)\right] & = \mathbb{E}\left[-\frac{1}{s}\sum_{k=1}^s\Delta_k\frac{\sum_{j=1}^sY_j(T_k)\exp(\zeta_0(V_j))[h^*-h^*_n](V_j)}{\sum_{j=1}^sY_j(T_k)\exp(\zeta_0(V_j))}\right],
\end{align*}
so that $I_{2n} =$
\begin{align*}
    & = \mathbb{E}\left[-\frac{1}{s}\sum_{k=1}^s\Delta_k\left\{\frac{\sum_{j=1}^sY_j(T_k)\exp(\zeta_0(V_j))[h^*-h^*_n](V_j)}{\sum_{j=1}^sY_j(T_k)\exp(\zeta_0(V_j))}-\frac{\sum_{j=1}^sY_j(T_k)\exp(\zeta_{\Tilde\vartheta^{SB(s)}_n}(V_j))[h^*-h^*_n](V_j)}{\sum_{j=1}^sY_j(T_k)\exp(\zeta_{\Tilde\vartheta^{SB(s)}_n}(V_j))}\right\}\right]\\
    & = -\mathbb{E}\left[\Delta_1\left\{\frac{\sum_{j=1}^sY_j(T_1)\exp(\zeta_0(V_j))[h^*-h^*_n](V_j)}{\sum_{j=1}^sY_j(T_1)\exp(\zeta_0(V_j))}-\frac{\sum_{j=1}^sY_j(T_1)\exp(\zeta_{\Tilde\vartheta^{SB(s)}_n}(V_j))[h^*-h^*_n](V_j)}{\sum_{j=1}^sY_j(T_1)\exp(\zeta_{\Tilde\vartheta^{SB(s)}_n}(V_j))}\right\}\right].
\end{align*}
By Lemma \ref{PLlemma: derivative}, it follows that
$$I_{2n}\lesssim \lVert h^*_n-h^*\rVert_{L^2}\lVert \Tilde\vartheta^{SB(s)}_n-\vartheta_0\rVert_{L^2} = O_p(\Upsilon_n^2\log^4n).$$
By assumption $\sqrt{n}\Upsilon_n^2\to 0$ as $n\to\infty$, we have $I_{2n} = o_p(1/\sqrt{n})$. This proves the lemma.
\end{proof}

\begin{lemma}\label{PLlemma: AVar}
For ${H}_s$ and $\Sigma_s$ defined in Theorem \ref{thm: PLNN},
 $${H}_s = s{\Sigma}_s, \text{ and }{H}_{2s} \succeq {H}_{s}.$$
\end{lemma}

\begin{proof}
The proof strategy is the same as the proof for Theorem \ref{thm: batch size and Hessian}. Note that
\begin{align*}
& {H}_s = \mathbb{E}\left[\Delta_1 \sum_{j=1}^s {w}_j(T_1,0)\left\{\rho(T_1,V_j,\vartheta_0)[\mathbf{I}_Z-\mathbf{h^*}]\right\}^{\otimes 2}\right]\\
= &  \mathbb{E}\left[\frac{1}{s}\sum_{k=1}^s\Delta_k \sum_{j=1}^s {w}_j(T_k,0)\left\{\rho(T_k,V_j,\vartheta_0)[\mathbf{I}_Z-\mathbf{h^*}]\right\}^{\otimes 2}\right] \\
= & \frac{1}{s}\sum_{k=1}^s \left\{\int_0^\tau\mathbb{E}\left[\sum_{j=1}^s {w}_j(u,0)\left\{\rho(u,V_j,\vartheta_0)[\mathbf{I}_Z-\mathbf{h^*}]\right\}^{\otimes 2}Y_k(u)\exp(\zeta_0(V_k))\right]\lambda_0(u)\mathrm{d}u\right\}. 
\end{align*}
where the last equation is due to (\ref{ap-eq:compensator}). By definition of ${w}_j(u,0)$, we have
$$ {w}_j(u,0)\sum_{k=1}^sY_k(u)\exp(\zeta_0(V_k)) = Y_j(u)\exp(\zeta_0(V_j)),$$
so that
\begin{equation}
\label{eq: PLlemma Hs}
{H}_s =\frac{1}{s}\int_0^\tau\mathbb{E}\left[\sum_{j=1}^s \left\{\rho(u,V_j,\vartheta_0)[\mathbf{I}_Z-\mathbf{h^*}]\right\}^{\otimes 2}Y_j(u)\exp(\zeta_0(V_j))\right]\lambda_0(u)\mathrm{d}u.    
\end{equation}
Similarly, by (\ref{ap-eq:compensator}), we have
$$  \mathbb{E}\left[l^{(s)}(\vartheta_0)[\mathbf{I}_Z-\mathbf{h^*}]\right] = \frac{1}{s}\int_0^\tau\mathbb{E}\left[\sum_{j=1}^s \rho(u,V_j,\vartheta_0)[\mathbf{I}_Z-\mathbf{h^*}]Y_j(u)\exp(\zeta_0(V_j))\right]\lambda_0(u)\mathrm{d}u =0. $$
This implies
\begin{equation}
\label{eq: PLlemma Sigmas}
\begin{split}
    {\Sigma}_s & = \mathbb{E}\left[l^{(s)}(\vartheta_0)[\mathbf{I}_Z-\mathbf{h^*}]^{\otimes 2}\right] \\
& = \mathbb{V}\left[l^{(s)}(\vartheta_0)[\mathbf{I}_Z-\mathbf{h^*}]\right]\\
& = \mathbb{V}\left[\int_0^\tau \frac{1}{s}\sum_{j=1}^s \rho(u,V_j,\vartheta_0)[\mathbf{I}_Z-\mathbf{h^*}] \mathrm{d}N_i(u)\right] \\
(1)& =   \frac{1}{s^2}\int_0^\tau\sum_{j=1}^s \mathbb{E}\left[\rho(u,V_j,\vartheta_0)[\mathbf{I}_Z-\mathbf{h^*}]^{\otimes 2} Y_j(u)\exp(\zeta_0(V_j))\right]\lambda_0(u)\mathrm{d}u,
\end{split}
\end{equation}
where (1) follows the same logic as (\ref{eq:var of martingale}). Comparing equations (\ref{eq: PLlemma Hs}) and (\ref{eq: PLlemma Sigmas}) leads to the conclusion $H_s= s\Sigma_s$.

Next we show $H_{2s}\succeq H_s$. Define $[\mathbf{I}_Z-\mathbf{h^*}](V_j):=Z_j-\mathbf{h^*}(X_j)$, then
\begin{equation*}
\begin{split}
{H}_s & =\frac{1}{s}\int_0^\tau\mathbb{E}\left[\sum_{j=1}^s \left\{\rho(u,V_j,\vartheta_0)[\mathbf{I}_Z-\mathbf{h^*}]\right\}^{\otimes 2}Y_j(u)\exp(\zeta_0(V_j))\right]\lambda_0(u)\mathrm{d}u\\
& = \frac{1}{s}\int_0^\tau\mathbb{E}\left[\sum_{j=1}^s \left\{[\mathbf{I}_Z-\mathbf{h^*}](V_j)-\sum_{k=1}^s w_k(u,0)[\mathbf{I}_Z-\mathbf{h^*}](V_k)\right\}^{\otimes 2}Y_j(u)\exp(\zeta_0(V_j))\right]\lambda_0(u)\mathrm{d}u \\
& := I_1^{(s)}+I_2^{(s)}+I_3^{(s)}+I_4^{(s)},
\end{split}
\end{equation*}
where
\begin{align*}
I_1^{(s)} & = \frac{1}{s}\int_0^\tau\mathbb{E}\left[\sum_{j=1}^s [\mathbf{I}_Z-\mathbf{h^*}](V_j)^{\otimes 2}Y_j(u)\exp(\zeta_0(V_j))\right]\lambda_0(u)\mathrm{d}u \\
& = \int_0^\tau\mathbb{E}\left[[\mathbf{I}_Z-\mathbf{h^*}](V)^{\otimes 2}Y(u)\exp(\zeta_0(V))\right]\lambda_0(u)\mathrm{d}u,
\end{align*}
\begin{align*}
I_4^{(s)} & = \frac{1}{s}\int_0^\tau\mathbb{E}\left[\sum_{j=1}^s \left\{\sum_{k=1}^s w_k(u,0)[\mathbf{I}_Z-\mathbf{h^*}](V_k)\right\}^{\otimes 2}Y_j(u)\exp(\zeta_0(V_j))\right]\lambda_0(u)\mathrm{d}u, \\
& = \frac{1}{s}\int_0^\tau\mathbb{E}\left[\left(\sum_{l=1}^sY_j(u)\exp(\zeta_0(V_l))\right)\left\{\sum_{k=1}^s w_k(u,0)[\mathbf{I}_Z-\mathbf{h^*}](V_k)\right\}^{\otimes 2}\right]\lambda_0(u)\mathrm{d}u,
\end{align*}
\begin{align*}
& I_2^{(s)} \\
& = -\frac{1}{s}\int_0^\tau\mathbb{E}\left[\sum_{j=1}^s \left\{\sum_{k=1}^s w_k(u,0)[\mathbf{I}_Z-\mathbf{h^*}](V_j)\otimes[\mathbf{I}_Z-\mathbf{h^*}](V_k)\right\}Y_j(u)\exp(\zeta_0(V_j))\right]\lambda_0(u)\mathrm{d}u \\
& = -\frac{1}{s}\int_0^\tau\mathbb{E}\left[\sum_{k=1}^s w_k(u,0)\sum_{j=1}^s \left\{[\mathbf{I}_Z-\mathbf{h^*}](V_j)Y_j(u)\exp(\zeta_0(V_j))\right\}\otimes[\mathbf{I}_Z-\mathbf{h^*}](V_k)\right]\lambda_0(u)\mathrm{d}u \\
& = -\frac{1}{s}\int_0^\tau\mathbb{E}\left[\left(\sum_{l=1}^sY_j(u)\exp(\zeta_0(V_l))\right)\sum_{k=1}^s w_k(u,0)\sum_{j=1}^s \left\{w_j(u,0)[\mathbf{I}_Z-\mathbf{h^*}](V_j))\right\}\otimes[\mathbf{I}_Z-\mathbf{h^*}](V_k)\right]\lambda_0(u)\mathrm{d}u \\
& = -\frac{1}{s}\int_0^\tau\mathbb{E}\left[\left(\sum_{l=1}^sY_j(u)\exp(\zeta_0(V_l))\right)\left\{\sum_{k=1}^s w_k(u,0)[\mathbf{I}_Z-\mathbf{h^*}](V_k)\right\}^{\otimes 2}\right]\lambda_0(u)\mathrm{d}u= -I_4^{(s)},
\end{align*}
and in a similar manner
\begin{align*}
I_3^{(s)} &= -\frac{1}{s}\int_0^\tau\mathbb{E}\left[\sum_{j=1}^s \left\{\sum_{k=1}^s w_k(u,0)[\mathbf{I}_Z-\mathbf{h^*}](V_k)\otimes[\mathbf{I}_Z-\mathbf{h^*}](V_j)\right\}Y_j(u)\exp(\zeta_0(V_j))\right]\lambda_0(u)\mathrm{d}u\\
&= -\frac{1}{s}\int_0^\tau\mathbb{E}\left[\left(\sum_{l=1}^sY_j(u)\exp(\zeta_0(V_l))\right)\left\{\sum_{k=1}^s w_k(u,0)[\mathbf{I}_Z-\mathbf{h^*}](V_k)\right\}^{\otimes 2}\right]\lambda_0(u)\mathrm{d}u = -I_4^{(s)}.
\end{align*}
Therefore, $H_s = I_1 + I_2^{(s)} +I_3^{(s)}+I_4^{(s)} = I_1 - I_4^{(s)}$, where $I_1$ does not depend on $s$.

To show $H_{2s}\succeq H_s$, it is sufficient to show $I_4^{(2s)}-I_4^{(s)}\preceq 0$, which holds by following the proof for (\ref{eq: H2s-Hs}) in Theorem \ref{thm: batch size and Hessian}.
\end{proof}

\section{Additional Discussion on SGD for Cox Regression}
\subsection{The Convergence Rate of Projected SGD}
Theorem \ref{thm:non-asymp-bound} shows that, when $\alpha\in (0,1)$, the convergence is at rate $O(\frac{1}{t^{\alpha}})$. When $\alpha=1$, the convergence rate is $O(\frac{1}{t})$ if $\mu C>1$, the convergence rate is $O(\frac{\log t}{t})$ if $\mu C=1$, and is $O(\frac{1}{t^{\mu C}})$ if $\mu C<1$. Therefore, when $\alpha=1$, the choice of $C$ is critical. If $C$ is too small, the convergence is at an arbitrarily small rate $O(\frac{1}{t^{\mu C}})$. The optimal convergence rate can be achieved regardless of $C$ by averaging. \citet{lacoste2012simpler} showed that, for projected SGD and if the target loss is strongly convex on $\mathbb{R}^p_B$, then $\lVert \check{\theta}_t-\theta_0 \rVert^2_2=O_p(\frac{1}{t})$ with learning rate $\gamma_t = \frac{C}{t+1}$ for any $C>0$, where $\check{\theta}_t:=\frac{2}{(t+1)(t+2)}\sum_{i=0}^t(i+1)\hat{\theta}_i$.

\subsection{High-probability bound on the Projected SGD}
In our original analysis, we provided the convergence bound in expectation, as this greatly simplifies recursion. By the linearity of expectation, one can replace the stochastic gradient with its expected gradient, which eliminates the noise at each step. This approach has been used in many foundational SGD analyses; for example, see the proof of Theorem 1 in  \citet{moulines2011non}. In contrast, deriving the high-probability bounds is more challenging since one must control the total noise that accumulates from each noisy gradient step. Recent advances, such as \cite{harvey2019tight}, provide high-probability bounds for SGD under certain assumptions. We can verify the general assumptions in \cite{harvey2019tight} to establish a high-probability bound for the projected SGD estimator in the Cox model.

Specifically, to find the minimizer $\theta^*=\arg\min_{\theta\in \Theta}f(\theta)$, the projected SGD updates the iteration through
$$\theta_{t+1} = \Pi_\Theta\left(\theta_t-\gamma_t g(\theta_t)\right)\text{, with } \mathbb{E}[g(\theta_t)] = \nabla f(\theta_t),$$
where $\Pi_\Theta$ is the projection operator on $\Theta$.
By Theorem F.1 in \cite{harvey2019tight}, if \textbf{
\begin{itemize}
    \item[(a)] $f$ is $\mu$-strongly convex on $\Theta$;
    \item[(b)] $f$ is $L$-Lipschitz on $\Theta$;
    \item[(c)] $\nabla f(\theta_t) - g(\theta_t)$, the noise of the stochastic gradient oracle, has norm at most $L$ almost surely;
\end{itemize}}
\noindent then running the projected SGD for $T$ iterations with the step size (learning rate) $\gamma_t = \frac{1}{\alpha t}$, we have, with probability at least $1-\delta$,
$$f(\theta_{T+1})-f(\theta^*)\leq O\left(\frac{L^2\log(T)\log(1/\delta)}{\alpha\mu T}\right).$$
Further, \textbf{if Condition (d) $\nabla f(\theta^*) = 0$ holds}, by the definition of strong convexity, this translates to a bound on the distance to the minimizer:
\begin{equation}
\label{eq: high-p bound}
    \lVert \theta^*-\theta_{T+1}\rVert^2 \leq \frac{2}{\mu}\{f(\theta_{T+1})-f(\theta^*)\}\leq O\left(\frac{L^2\log(T)\log(1/\delta)}{\alpha\mu^2 T}\right),
\end{equation}
with probability at least $1-\delta$. 

For Cox regression, the objective becomes the mini-batch-based loss function $f(\theta)=\mathbb{E}[L^{(s)}_{Cox}(\theta)]$. Under our assumptions (A1)-(A3),(R1)-(R2), and if $\theta_0\in \mathbb{R}^p_B$, we can verify that the above general conditions (a)-(d) hold for the Cox regression: 
\begin{itemize}
    \item[(a)] $\mathbb{E}[L^{(s)}_{Cox}(\theta)]$ is (locally) strongly convex on $\mathbb{R}^p_B$ (by Lemma 2)
    \item[(b)] $\mathbb{E}[L^{(s)}_{Cox}(\theta)]$ is Lipschitz  on $\mathbb{R}^p_B$ (differentiable with bounded gradient)
    \item[(c)] The noise of stochastic gradient, i.e., $\nabla_\theta\mathbb{E}[L^{(s)}_{Cox}(\theta)]-\nabla_\theta L^{(s)}_{Cox}(\theta)$, is bounded due to the boundedness of the covariate $X$ by Assumption (A3) and $\theta_t \in \mathbb{R}^p_B$.
    \item[(d)] the gradient of the loss function at $\theta_0$ is zero ($\nabla_\theta\mathbb{E}[L^{(s)}_{Cox}(\theta)]|_{\theta=\theta_0} = 0$ by Lemma 1)
\end{itemize}
Therefore, we obtain the high-probability bound for the projected SGD estimator in the Cox model, as shown in Equation (\ref{eq: high-p bound}), and the formal statement is presented below. We have added this discussion to the Appendix.

\begin{theorem}\label{thm:hugh-p-bound}
Consider the SGD procedure \begin{equation*}
\hat\theta_{t+1} = \Pi_{\mathbb{R}^p_B}[ \hat\theta_t - \gamma_t \nabla_\theta L_{Cox}^{(s)}(\hat\theta_t)],    
\end{equation*} with the learning rate $\gamma_t = \frac{C}{t}$ and the constant $C>0$. Under the Cox regression, assume assumptions (A1)-(A3), (R1)-(R2), and assume $\lVert \theta_0\rVert \leq B$, then for any integer $s\geq 2$, we have with probability at least $1-\epsilon$,
\begin{equation}
    \lVert \hat\theta^{(s)}_t-\theta_0\rVert ^2 = O\left(\frac{CD^2\log(t)\log(1/\epsilon)}{\mu^2 t}\right),
\end{equation}
where $D=\displaystyle\max_{\theta\in \mathbb{R}^p_B,D(s)} \lVert \nabla_\theta L_{Cox}^{(s)}(\theta) \rVert$, and $\mu$ is the strong-convexity constant in Lemma 2. 
\end{theorem}

\subsection{Averaged SGD (ASGD) and its Inference}
Uncertainty quantification of the SGD is important in practice. For inference purposes, we consider the averaged SGD (ASGD), which enjoys the optimal convergence rate, and its asymptotic behaviour has been well established. Specifically, consider the standard SGD iteration (without the projection)
\begin{equation}
\hat\theta_{t+1} = \hat\theta_t - \gamma_t \nabla_\theta L_{Cox}^{(s)}(\hat\theta_t).    \label{eq:ASGD updates for Cox}
\end{equation}
Instead of reporting the estimator from the last iteration, ASGD reports the running average estimator $\Bar{\hat{\theta}}_t = \frac{1}{t}\sum_{k=1}^t\hat\theta_{k}$. By reducing the variance through the averaging, this estimator is more stable and is more robust to the choice of learning rate $\gamma_t$.

We establish the asymptotic normality of the ASGD estimator for Cox regression through Lemma 1 in \citet{fang2018online}, which restates the arguments in \citet{ruppert1988efficient,polyak1992acceleration}. Because the loss function $\mathbb{E}[L^{(s)}_{Cox}(\theta)]$ is convex and thrice continuously differentiable over $\mathbb{R}^p$ with
    $$\nabla_\theta \mathbb{E}[L^{(s)}_{Cox}(\theta)] = \mathbb{E}\left[\frac{1}{s}\sum_{i=1}^s \Delta_i \left\{-X_i+\frac{S^{(1)}_s (T_i,\theta)}{S^{(0)}_s (T_i,\theta)} \right\}\right],$$
and 
    $$\nabla^2_\theta \mathbb{E}[L^{(s)}_{Cox}(\theta)] = \mathbb{E}\left[\frac{1}{s}\sum_{i=1}^s \Delta_i \left\{\frac{S^{(2)}_s (T_i,\theta)}{S^{(0)}_s (T_i,\theta)}-\frac{S^{(1)}_s (T_i,\theta)^{\otimes 2}}{S^{(0)}_s (T_i,\theta)^{\otimes 2}} \right\}\right],$$
     where $S^{(k)}_s (u,\theta) := \sum_{j=1}^s Y_j(u)\exp(X_j^T \theta)X_j^{\otimes k}$. With this, Assumptions (A1)-(A4) in \citet{fang2018online} can be easily verified under our Assumption (A1)-(A3), (R1)-(R2).
Then the asymptotic distribution of $\Bar{\hat{\theta}}_t$ follows Lemma 1 in \citet{fang2018online}. Specifically, we have
\begin{equation}
    \label{eq: sgd asym norm}
    \sqrt{t}(\Bar{\hat{\theta}}_t-\theta_0)\to^d N(0,A^{-1}_sB_s(A_s^{-1})^T)
\end{equation}
for the learning rate $\gamma_t = \frac{C}{t^\alpha}$ with $\alpha\in(0.5,1)$ and any $C>0$, where $A_s = \mathbb{E}[\nabla^2_\theta L_{Cox}^{(s)}(\theta)]|_{\theta=\theta_0}$ and $B_s = \mathbb{V}[\nabla_\theta L_{Cox}^{(s)}(\theta)]|_{\theta=\theta_0}$. The asymptotic variance of $\Bar{\hat{\theta}}_t$ is closely related to the asymptotic variance of $\Tilde{\theta}^{FB(s)}_{n}$ (the offline estimator with FB sampling strategy). Because $s$ i.i.d. samples are drawn from the population in each iteration step, after $t$th iterations, the total number of input samples is $n = t\times s$. Since $s$ is a fixed constant, the asymptotic variance of $\Bar{\hat{\theta}}_t$ in terms of the total sample size $n$ is $\displaystyle\lim_{n\to\infty}n\mathbb{V}[\Bar{\hat{\theta}}_t] = s\displaystyle\lim_{t\to\infty}t\mathbb{V}[\Bar{\hat{\theta}}_t]  = sA^{-1}_sB_s(A_s^{-1})^T$. Then the asymptotic distribution of $\Bar{\hat{\theta}}_t$ in terms of the total sample size $n=t \times s$ is
$$\sqrt{n}(\Bar{\hat{\theta}}_t-\theta_0)\to^d N(0,sA^{-1}_sB_s(A_s^{-1})^T),$$
which coincides the asymptotic distribution of $\Tilde{\theta}^{FB(s)}_{n}$ given in Theorem 3. The offline estimator $\Tilde{\theta}^{FB(s)}_{n}$ is obtained by sampling $D(s)$ from $D(n|s)$ (split the whole data set into $\frac{n}{s}$ non-overlap batches). Notice that as the sample size $n\to\infty$, sampling $D(s)$ from $D(n|s)$ is equivalent to sampling from the population. This explains the identical asymptotic variance of the two estimators when the sample size $n\to\infty$.

The asymptotic normality (\ref{eq: sgd asym norm}) is useful for quantifying the uncertainty of the ASGD estimator. The asymptotic variance can be estimated by plugging in empirical estimates of each component using the full dataset (with all data stored). For an online approach that estimates the asymptotic variance without storing the entire dataset, we refer readers to \citet{zhu2023online}.











\bibliographystyle{agsm}
\bibliography{reference}